\documentclass{agujournal2019}
\usepackage{lineno}
\usepackage[finalnew]{trackchanges}
\usepackage{soul}
\usepackage{microtype}

\usepackage{multicol,multirow}
\usepackage{amsmath,amssymb,amsfonts}
\usepackage{mathrsfs}
\usepackage{amsthm}
\usepackage{rotating}
\usepackage{booktabs}

\usepackage{appendix}
\usepackage{ifpdf}
\usepackage{xcolor}

\usepackage{subcaption}
\usepackage{lscape}
\usepackage{enumitem}
\usepackage{makecell}
\usepackage{microtype}
\usepackage{floatrow}
\usepackage{verbatim}
\usepackage{algorithm}
\usepackage{algpseudocode}

\usepackage[implicit=false, hidelinks]{hyperref}

\let\oldhref\href
\renewcommand{\href}[2]{\oldhref{#1}{\texttt{#2}}}

\journalname{Journal of Advances in Modeling Earth Systems (JAMES)}

\begin{document}


\title{{Replacing Tunable Parameters in Weather and Climate Models with State-Dependent Functions using Reinforcement Learning}}

%
%




\authors{Pritthijit Nath\affil{1}, Sebastian Schemm\affil{1}, Henry Moss\affil{1,2}, Peter Haynes\affil{1}, \\ Emily Shuckburgh\affil{3}, and Mark J. Webb\affil{4}}

\affiliation{1}{Department of Applied Mathematics and Theoretical Physics, University of Cambridge, UK}
\affiliation{2}{School of Mathematical Sciences, Lancaster University, UK}
\affiliation{3}{Department of Computer Science and Technology, University of Cambridge, UK}
\affiliation{4}{Met Office Hadley Centre, UK}

\correspondingauthor{Pritthijit Nath}{pn341@cam.ac.uk}




\begin{keypoints}
\item We investigate using RL to learn policies that set tunable model parameters as functions of model state during training.
\item Results show RL-assisted schemes reduce errors across various idealised climate models relative to static baselines.
\item Multiple RL agents that learn in parallel and share weights accelerate convergence and improve skill across latitude bands.
\end{keypoints}


%
%

%
%


\begin{abstract}
Weather and climate models rely on parametrisations to represent unresolved sub-grid processes. Traditional schemes rely on fixed coefficients that are weakly constrained and tuned offline, contributing to persistent biases that limit their ability to adapt to underlying physics. This study presents a framework that learns components of parametrisation schemes online as a function of the evolving model state using reinforcement learning (RL) and evaluates \change{RL-driven}{policy-driven} parameter updates across idealised testbeds spanning a simple climate bias correction (SCBC), a radiative-convective equilibrium (RCE), and a zonal mean energy balance model (EBM) with single-agent and federated multi-agent settings. Across nine RL algorithms, Truncated Quantile Critics (TQC), Deep Deterministic Policy Gradient (DDPG), and Twin Delayed DDPG (TD3) achieved the highest skill and stable convergence, with performance assessed against a static baseline using area-weighted RMSE, temperature and pressure-level diagnostics. For the EBM, single-agent RL outperformed static parameter tuning with the strongest gains in tropical and mid-latitude bands, while federated RL on multi-agent setups enabled specialised control and faster convergence, with a six-agent DDPG configuration using frequent aggregation yielding the lowest area-weighted RMSE across the tropics and mid-latitudes. The learnt corrections were also physically meaningful as agents modulated EBM radiative parameters to reduce meridional biases, adjusted RCE lapse rates to match vertical temperature errors, and stabilised heating increments to limit drift. {Overall, results show that RL can learn skilful state-dependent parametrisation components in idealised settings, offering a scalable pathway for online learning within numerical models and a starting point for evaluation in weather and climate models.}
\end{abstract}

\section*{Plain Language Summary}
Weather and climate models cannot fully resolve small-scale processes such as cloud formation, radiation, and turbulence, so these effects are represented using simplified rules known as parametrisations. These rules are usually tweaked during model development and then held fixed during simulations, which can contribute to persistent biases when models are compared with observations. In this study, we explore whether reinforcement learning (RL), used as an online trial-and-error training method, can be used to learn machine-learning (ML) components that set existing tunable parameters as a function of the model state, while reducing biases against temporally sparse observations. We test this approach across a hierarchy of simplified climate models, ranging from bias correction to single-column convection and zonal energy balance models. Results show that RL can reduce errors relative to traditional methods while adjusting parameters in physically meaningful ways. When multiple RL agents share information, learning is faster and more responsive to local conditions. Overall, these findings suggest that RL could support the development of state-dependent parametrisation components that are trained online, then frozen for operational use, with modest computational cost compared to large contemporary ML models.

%
%


\section{Introduction}
\label{sec:introduction}
Weather and climate models solve the governing equations of fluid dynamics, thermodynamics, continuity, and moisture transport on discretised grids, but processes occurring at scales smaller than the grid resolution cannot be explicitly represented~\cite{kalnay_atmospheric_2002, hartmann_climate_2016}. To account for these unresolved processes, such as moist convection, cloud microphysics, boundary-layer turbulence, and gravity wave drag, models rely on parametrisations that approximate their aggregate influence on the resolved flow~\cite{stensrud_parameterization_2007, randall_climate_2007}. Although indispensable for long-term simulations, parametrisations are a major source of structural error since they depend on empirical closures and limited observations~\cite{randall_climate_2007, vial_interpretation_2013, webb_cloud_2017}. 
This reliance on simplified formulations often leads to persistent biases in climate sensitivity, precipitation, and circulation patterns~\cite{soden_assessment_2006, flato_evaluation_2014}. For instance, in weather forecasts, errors in parametrising mesoscale convection over the continental United States can increase ensemble spread downstream over the North Atlantic~\cite{lojko_remote_2022, baumgart_potential_2018}, while in climate simulations, parametrisation errors can result in an uncertain response to warming. Additionally, many schemes are resolution-dependent and degrade when grid spacing approaches the scales of the represented processes, constraining their applicability in kilometre-scale models~\cite{hallberg_using_2013, stevens_dyamond_2019, schneider_opinion_2024}. Despite decades of refinement, parametrisations remain the dominant source of uncertainty in general circulation model (GCM) projections~\cite{ceppi_cloud_2017}, motivating the development of scale-adaptive machine learning (ML)-based alternatives which, although still in their early stages, may help reduce biases, improve robustness, and improve confidence in future weather and climate projections~\cite{morcrette_scale-aware_2025}.

Historically, much tuning of numerical models has relied on manual, expert-driven adjustments targeting a small set of global or regional diagnostics such as the top-of-atmosphere (TOA) energy balance or global precipitation patterns~\cite{hourdin_art_2017}. These manual procedures are time-consuming, subjective, and often can yield parameter choices that overcompensate for structural errors elsewhere in the model rather than reducing process-level error~\cite{mauritsen_tuning_2012}. Automated tuning workflows help mitigate some of the associated cost and reproducibility issues, yet they remain constrained by the choice of summary metrics and emulator accuracy, which can bias the explored parameter subspace~\cite{bonnet_tuning_2025, lguensat_semi-automatic_2023}. Offline ML approaches, trained to map resolved states to subgrid tendencies using high-resolution simulations or observations, show promise but frequently fail to transfer reliably when coupled into a full GCM, since offline loss functions do not guarantee online stability or realistic long-term climatology~\cite{chen_stable_2025, brenowitz_machine_2020}. This offline–online gap arises because ML parametrisations often learn spurious correlations and omit coupled-system feedbacks, allowing small errors to accumulate into large biases and producing instabilities or degraded circulation and precipitation statistics when integrated forward~\cite{bertoli_revisiting_2025, lin_navigating_2025}. Furthermore, both traditional tuning and many offline ML studies under-quantify uncertainty, requiring large ensembles or Bayesian calibration to assess robustness, which adds significant computational cost and complicates operational adoption~\cite{howland_parameter_2022, dunbar_calibration_2021}. Finally, when both approaches are hardware-accelerated, offline-trained ML parametrisations are not always necessarily faster than well-tuned conventional parametrisations, reducing the case for ML unless it demonstrably improves predictive skill~\cite{bertoli_revisiting_2025}.

Reinforcement learning (RL)~\cite{sutton_reinforcement_1998, silver_welcome_2025}, one of the key driving forces behind the reasoning and alignment of large language models (LLMs)~\cite{guo_deepseek-r1_2025, ouyang_training_2022}, gained prominence from mid-2010s when it was shown to demonstrate superhuman performance in games such as Atari~\cite{mnih_playing_2013}, Go~\cite{silver_mastering_2016} and general gameplay~\cite{schrittwieser_mastering_2020}, as well as recently in physical domains such as robot table tennis~\cite{su_hitter_2025}. In the weather and climate context, RL offers a state-aware alternative to static, hand-tuned parametrisations, where policies (action selection rules) set parameters as a function of the model state at each timestep and the learning task is framed as a sequence of decisions optimised for long-horizon performance in the coupled model during training. Unlike offline surrogates trained on short-term simulations or reduced diagnostics, RL learns in a closed loop with the numerical model (without requiring it to be end-to-end differentiable), potentially allowing rewards to directly enforce physical constraints~\cite{novati_automating_2021, kurz_deep_2023} and numerical stability while balancing short-term forecast skill against long-term climatology. Continuous-control algorithms such as deep deterministic policy gradients (DDPG)~\cite{silver_deterministic_2014, lillicrap_continuous_2019} and truncated quantile critics (TQC)~\cite{kuznetsov_controlling_2020} provide practical mechanisms for stable updates in high-dimensional action spaces, which map naturally onto the multivariate parameter sets of climate models. Evidence from other domains of complex physics control demonstrates that deep RL can maintain stability under strict constraints, for instance in magnetic shape control of tokamaks~\cite{degrave_magnetic_2022}, closed-loop separation control in turbulent flows~\cite{bae_scientific_2022, font_deep_2025}, and optimisation of wind farm power generation~\cite{mole_reinforcement_2025}. Moreover, approaches such as federated and distributed RL approaches align naturally with the domain decomposition strategies of general circulation models, where local agents can be trained on subdomains and periodically aggregated into a coherent global policy~\cite{jin_federated_2022}. Collectively, these features highlight RL as a promising pathway towards state-dependent parametrisations that adapt with evolving climate states. 

Related approaches include online ensemble Kalman inversion (EKI)~\cite{christopoulos_online_2024, kovachki_ensemble_2019}, which estimates parameters in-situ by iteratively updating closures to match target diagnostics and has shown strong results in single-column settings. {Methods such as calibrate-emulate-sample (CES)}~\cite{cleary_calibrate_2021, dunbar_calibration_2021} {and related emulator-based calibration strategies provide a powerful route for estimating well-calibrated static parameter values or parameter distributions offline, together with uncertainty quantification, and are therefore an important complementary class of approaches. In contrast, the main motivation for RL in the present work is the natural casting of parametrisations as a sequential control problem, allowing tunable parameters to be adjusted online as a function of the evolving model state. This makes RL particularly attractive when the desired parametrisation is state-dependent rather than fixed, although a full like-for-like comparison with CES is beyond the scope of the present exploratory study.} 

{Recent hybrid physics--AI studies have shown that subgrid closures can be learnt and coupled online within host climate models while retaining physical structure through embedded machine-learned subgrid variability schemes}~\cite{giles_embedding_2024},~{made scale-aware for cloud fraction and condensate parametrisation by learning from coarse-grained kilometre-scale simulations}~\cite{morcrette_scale-aware_2025},~{stabilised over long online integrations in comprehensive atmospheric models using subgrid emulators trained on embedded convection-permitting simulations}~\cite{hu_stable_2025},~{and combined with physically consistent equation discovery and automatic tuning to reduce persistent cloud-cover errors in hybrid AI-climate pipelines}~\cite{bertoli_revisiting_2025, grundner_reduced_2025}.

Building on the promise of RL, this study aims to design, implement, and critically evaluate RL-assisted parametrisation schemes that learn from observations while preserving the governing physical principles encoded in the host model, including conservation laws. Rather than replacing existing physics, \change{the RL-components replace static tunable parameters with state-dependent functions}{static tunable parameters are replaced with state-dependent policy functions} that learn to control these parameters in a way that remains consistent with the model’s energy conservation and dynamics. Since running directly on full GCMs is computationally expensive and often difficult to interpret, this study instead focuses on experiments across a hierarchy of idealised climate testbeds allowing {the use of simple reward functions to assess the effectiveness of  generic reward designs, alongside} rapid prototyping and efficient screening of algorithms before moving to full GCM experiments in the future. {These testbeds were chosen deliberately as controlled proof-of-concept environments in which the control objective is well defined, the relevant tunable parameters are known, and policy behaviour can be interpreted clearly against reference climatologies or calibrated targets. While they do not reproduce the full difficulty of chaotic multi-scale systems, they provide a useful first stage for assessing whether RL can learn meaningful state-dependent adjustments at all before moving to substantially more complex atmospheric models.} The progression begins with single-agent formulations: first, a simple temperature bias correction environment is formulated where an RL agent learns {a policy that outputs} state-dependent corrections to a heating rate parameter for bias-correcting the temperature value. Next, a radiative–convective equilibrium (RCE) setup is designed where the agent adjusts two tunable parameters, evaluated in terms of long-horizon stability and temperature profile bias against reanalysis mean. Finally, building on these results, progressively more complex experiments are carried out with a multi-parameter Budyko–Sellers zonal mean energy balance model (EBM)~\cite{budyko_effect_1969, sellers_global_1969, north_theory_1975}, together with federated coupling strategies that extend from robust single-agent continuous-control setups to multi-agent federated designs that mirror the spatial decomposition of GCMs. A central aim of this transition is to quantify how learning dynamics change under spatial decomposition and whether agents achieve improved skill through parallel learning in local regions. {As the RL agent modifies tunable parameters within the existing model structure, key conservation properties of the underlying physical model remain governed by the host model formulation.} Together, these idealised experiments provide practical prototypes for integrating RL into operational parametrisation development, along with code for diagnostics and reproducible training recipes.

The remainder of this paper is structured as follows. Section~\ref{sec:methods} introduces the basics of RL, outlines the federated learning setup and coupling strategies used to integrate RL into climate simulations, and then describes the hierarchy of climateRL testbeds along with the training workflow used. Section~\ref{sec:results_and_discussion} reports results from single- and multi-agent experiments, with discussions on  stability, skill, physical meaningfulness and braoder implications on future parametrisation development. Finally, Section~\ref{sec:conclusion} summarises the key findings, highlights avenues for future research, and points to openly available code and experiment data for transparency and reproducibility.

\clearpage

\section{Methods}
\label{sec:methods}
\subsection{Reinforcement Learning (RL)}

RL addresses sequential decision-making problems under uncertainty by allowing an agent to interact with an environment and optimise its behaviour through cumulative rewards~\cite{silver_mastering_2016, kiran_deep_2021, font_deep_2025}. Unlike supervised learning which depends on labelled input–output pairs, RL relies on trial-and-error interaction with the environment where policies improve as the agent receives feedback from its own actions. In contrast to other online learning strategies such as bandits, which optimise only immediate rewards without state information~\cite{lattimore_bandit_2020}, RL utilises the formalism of a Markov Decision Process (MDP) to use state information to model and solve sequential decision making problems. {For our application, this means that the evolving numerical model state is treated as the state of the decision process, parameter updates are treated as actions, and the reward quantifies how desirable the resulting model behaviour is over time.}

\subsubsection{{MDP Formalism}}

\begin{figure}[!h]
\centering
\includegraphics[width=0.8\linewidth]{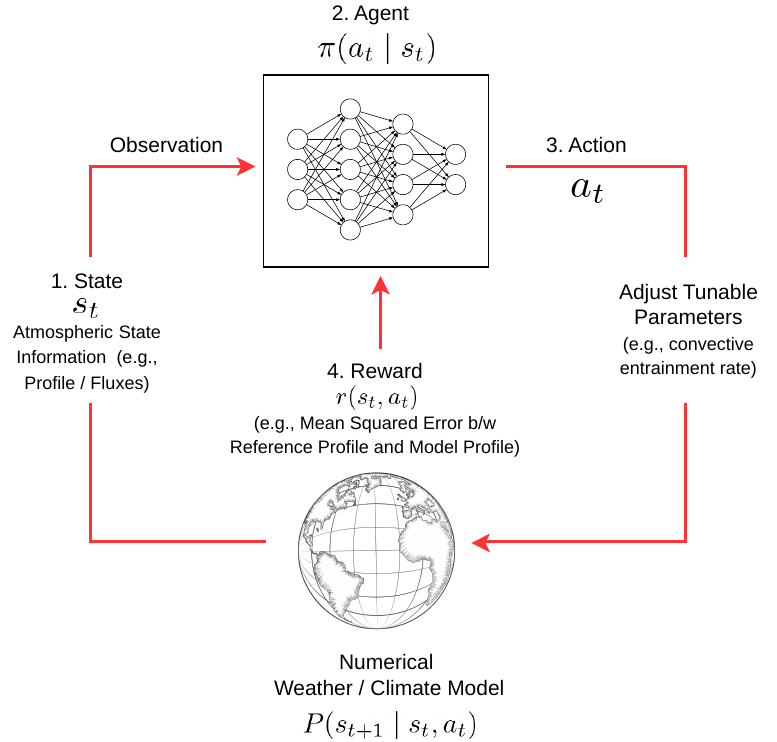}
\caption{{MDP formulation of the RL-based parameter-control framework. The numerical weather or climate model provides the state $s_t$ to the agent, which outputs an action $a_t$ that adjusts tunable parameters. The model then evolves to the next state $s_{t+1}$, and a reward $r(s_t,a_t)$, such as a negative mean-squared error against a reference profile, is used to improve the policy.}}
\label{fig:methods-mdp-flow}
\end{figure}

At the core of the formalism (shown in Figure~\ref{fig:methods-mdp-flow}), is an ``agent'' that interacts with an ``environment'' over discrete timesteps by observing a state \( s_t \), selecting an action \( a_t \) according to its policy \( \pi(a_t \mid s_t) \) (which may be stochastic or deterministic), receiving reward \( r_t \), and transitioning to a new state \( s_{t+1} \) under some unknown dynamics \( P(s_{t+1} \mid s_t, a_t) \). The fundamental ingredients are: (i) the state space \( \mathcal{S} \), (ii) the action space \( \mathcal{A} \), (iii) the reward function \( r(s_t,a_t) \), (iv) the transition dynamics \( P \), (v) the action policy \( \pi \), and (vi) the cumulative discounted reward (known as return) \( R_t = \sum_{k=0}^\infty \gamma^k r_{t+k} \), with discount factor \( \gamma \in [0,1) \) over an episode (a finite sequence of timesteps from an initial state until a specified terminal time or stopping criterion). Together, these elements help define {the climate-model environments (introduced later in this text) as} a MDP, where the aim {in contrast to identify a single optimal constant parameter set}, is to learn a policy function \( \pi(a_t \mid s_t) \) {mapping model state to adjustments in the tunable parameters}, that maximises the expected return \( R_t \) while balancing exploration and exploitation (selectively preferring actions that appear to maximise \( R_t \)). 

Here, the MDP formalism describes the climate-model environment, while RL serves as the realisation that uses the MDP formalism to learn the parameters of the policy function. \add{An ``agent'' here denotes the learning entity, consisting of the policy and the RL algorithm used to update the weights of the policy network. As, the policy is the learnt state-to-action function inside the agent, the two are functionally interchangeable when discussing action selection. For comparisons, agents are labelled by the RL algorithm used. In the literature,} RL algorithms are commonly divided into two classes: model-free approaches, which learn directly from sampled system trajectories (e.g., Q-learning~\cite{watkins_q-learning_1992}, REINFORCE~\cite{williams_simple_1992}), and model-based approaches, which additionally learns a surrogate model \( P_\theta \) to support planning and improve data efficiency. In this study, the focus is on model-free methods, where RL agents interact directly with the numerical model \(P\) at each timestep by selecting an action \(a_t\) that sets the tunable parameters, after which the model integrates the state from \(s_t\) to \(s_{t+1}\), eliminating the need for a separate surrogate dynamics model.

\subsubsection{TD Update}
The state-value function \( V^\pi(s) \) and the action-value function \( Q^\pi(s,a) \) quantify the expected return from a state \( s \), and from a state–action pair \( (s,a) \) respectively, when following a policy \( \pi \). Both functions satisfy the ``Bellman equations", which express each value as the immediate reward plus the discounted estimate of the subsequent state (or state–action pair). In realistic environments, solving these equations exactly is infeasible, and approximation methods are required. Temporal-difference (TD) learning provides one such approach by updating estimates directly from already sampled transitions and ``bootstrapping'' from existing predictions:
\begin{align}
V(s_t) &\leftarrow V(s_t) + \alpha \big[ r_t + \gamma V(s_{t+1}) - V(s_t) \big], \\
Q(s_t,a_t) &\leftarrow Q(s_t,a_t) + \alpha \big[ r_t + \gamma Q(s_{t+1},a_{t+1}) - Q(s_t,a_t) \big],
\end{align}
where the bracketed terms denote the TD errors for \( V \) and \( Q \), $\alpha$ represents the learning rate and $\gamma$ denotes the discount factor as described before. TD methods underpin many value-based RL algorithms such as Q-learning.  By updating incrementally at each timestep without requiring full episodic returns, TD methods provide the foundation of many scalable model-free RL algorithms.

\subsubsection{On-Policy vs Off-Policy}

In on-policy methods, the agent improves the same policy \( \pi \) that is used while generating its training data (in our case, produced by the numerical model \( P \)). Examples include REINFORCE and Proximal Policy Optimisation (PPO)~\cite{schulman_proximal_2017}, which require fresh trajectories for each update. Although this increases sample complexity, such approaches often can produce faster learning. Off-policy methods, by contrast, allow the learning of a target policy \( \pi_{\mathrm{target}} \) from experience collected under a similar (but different) policy (e.g., a time-lagged version of the current policy). This distinction enables reuse of past data and greatly improves sample efficiency. Algorithms such as Deep Q-learning~\cite{mnih_playing_2013} and Deep Deterministic Policy Gradient (DDPG)~\cite{lillicrap_continuous_2019} achieve stability through several mechanisms: bootstrapping, where value estimates are updated using other learnt estimates; target networks, where slowly updated copies of networks reduce instability arising from feedback loops; and replay buffers, which break correlations in trajectories by storing and resampling transitions. These techniques underpin the stability of many modern off-policy algorithms such as Twin Delayed DDPG (TD3)~\cite{fujimoto_addressing_2018}, Soft Actor–Critic (SAC)~\cite{haarnoja_soft_2018}, and Truncated Quantile Critics (TQC)~\cite{kuznetsov_controlling_2020}.

\subsubsection{{Policy Gradients}}

In policy-based RL, a parametrised stochastic policy \( \pi_\theta(a \mid s) \) is optimised to maximise the expected return under a set of state-action trajectories. Given a trajectory \( \tau = (s_0, a_0, s_1, a_1, \ldots) \) generated by \( \pi_\theta \) with discount factor \( \gamma \in [0,1) \), the performance objective can be written as:
\begin{equation}
J(\theta) = \int_\tau \pi_\theta(\tau) R(\tau) d\tau  \ = \mathbb{E}_{\tau \sim \pi_\theta} \!\left[ R(\tau) \right] 
= \mathbb{E}_{\tau \sim \pi_\theta} \!\left[ \sum_{t=0}^\infty \gamma^t r(s_t, a_t) \right].
\end{equation}
To compute the gradient of this objective, the \textcolor{black}{log-derivative trick} is applied and expanded across timesteps:
\begin{align}
\nabla_\theta J(\theta) 
&= \nabla_\theta \int_\tau \pi_\theta(\tau) R(\tau) \, d\tau 
= \int_\tau \textcolor{black}{\nabla_\theta \pi_\theta(\tau)} R(\tau) \, d\tau 
= \int_\tau \textcolor{black}{\pi_\theta(\tau) \nabla_\theta \log \pi_\theta(\tau)} R(\tau) \, d\tau \\
&= \mathbb{E}_{\tau \sim \pi_\theta} \!\left[ \nabla_\theta \log \pi_\theta(\tau) R(\tau) \right] 
= \mathbb{E}_{\tau \sim \pi_\theta} \!\left[ \sum_{t=0}^\infty \nabla_\theta \log \pi_\theta(a_t \mid s_t) \, R_t \right],
\end{align}
where \( R_t = \sum_{k=t}^\infty \gamma^{k-t} r(s_k, a_k) \) is the cumulative discounted reward from time \( t \). This formulation yields an unbiased gradient estimator and forms the basis of the REINFORCE algorithm upon which numerous modern RL algorithms are based, although in practice, variance-reduction techniques such as baselines or advantage functions are often introduced in addition to improve stability and efficiency of learning.

\subsubsection{Actor–Critic Methods}
\label{sec:methods-rl-actor-critic-methods}

Actor–critic algorithms combine the strengths of value-based learning (via TD update) and policy-based learning (via policy gradients) by maintaining two sets of learnable function approximators. The actor, parametrised by \( \theta \), defines the policy \( \pi_\theta(a \mid s) \), while the critic, parametrised by \( w \), estimates returns through a value function \( V_w(s) \) or an action-value function \( Q_w(s,a) \). The actor is updated via gradient \underline{a}scent on the policy objective:
\[
\nabla_\theta J(\theta) = \mathbb{E}_{\pi_\theta} \!\left[ \nabla_\theta \log \pi_\theta(a_t \mid s_t) \, A^\pi(s_t,a_t) \right],
\]
where the advantage function \( A^\pi(s_t,a_t) = Q^\pi(s_t,a_t) - V^\pi(s_t) \) reduces variance by comparing action values against a state-dependent baseline.  

For deterministic policies \( \mu_\theta(s) \), as used in algorithms such as DPG, DDPG and TD3, the update, after mathematical simplifications, takes the form
\[
\nabla_\theta J(\theta) = \mathbb{E} \!\left[ \nabla_a Q_w(s,a) \big|_{a=\mu_\theta(s)} \, \nabla_\theta \mu_\theta(s) \right].
\]

Meanwhile, the critic updates its parameters \( w \) by minimising the temporal-difference (TD) error,
\[
\delta = r + \gamma Q_w(s_{t+1}, \pi_\theta(s_{t+1})) - Q_w(s,a), 
\quad \mathcal{L}(w) = \tfrac{1}{2} \delta^2,
\]
using gradient \underline{de}scent. This squared-error loss \( \mathcal{L}(w) \) encourages the critic’s estimates to align with the one-step bootstrapped target \( r + \gamma Q_w(s_{t+1},\pi_\theta(s_{t+1})) \), ensuring consistency with the Bellman equation~\cite{sutton_reinforcement_1998}.  

Through this interplay, the actor refines its policy based on the critic’s feedback, while the critic improves its value estimates from trajectories sampled through interactions with the environment. Collectively, this dual update mechanism yields a stable and sample-efficient framework that underpins various current continuous-control RL algorithms used in practice.

\subsubsection{Challenges of Deep RL and Mitigation Strategies}

The use of deep neural networks as function approximators (for actor and critic components) makes RL powerful but also prone to instability. Several sources of difficulty arise in practice:

\begin{itemize}
  \item \textbf{Non-stationarity:} As the policy improves, the distribution of observed states changes, creating a moving target for policy learning. Replay buffers mitigate this by storing past transitions and sampling them as approximately independent and identically distributed (i.i.d.) batches, which helps to stabilise updates.  

  \item \textbf{Bootstrapping Error:} Temporal-difference targets depend on their own estimates, which can amplify errors and cause divergence (known as ``overestimation bias``).. Stability is improved by using slowly updated target networks, where parameters are updated according to \( \theta^{\text{target}} \leftarrow \tau \theta + (1 - \tau)\theta^{\text{target}} \) with \( \tau \ll 1 \).  

  \item \textbf{Correlated Samples:} Consecutive transitions are highly correlated, violating the i.i.d.~assumption required for stochastic gradient descent (SGD). Replay buffers reduce this problem by randomising the sampling of experience.  

  \item \textbf{High Gradient Variance:} Policy gradients can exhibit large variance, slowing convergence. Techniques such as baseline subtraction (e.g., using \( V^\pi(s) \), as discussed in Section~\ref{sec:methods-rl-actor-critic-methods}) and entropy regularisation (used in methods such as SAC) help reduce variance and promote more consistent exploration.  
\end{itemize}

Together, these strategies: replay buffers, target networks, variance-reducing baselines, and entropy regularisation, {are not treated here as separate engineering modifications, but as standard algorithmic components of modern deep RL methods,} that form the basis of stable and scalable deep RL methods (discussed later in the text and in Appendix~\ref{app:methods-rl-algorithm-summaries}), enabling their application to high-dimensional continuous-control problems.

\subsection{Federated Learning}
\subsubsection{FedRL Horizontal Domain Decomposition}

Operational GCMs, such as the Met Office Unified Model (UM)~\cite{brown_unified_2012}, use domain decomposition in which the global domain $\Omega$ is partitioned into subdomains $\{\Omega_i\}_{i=1}^N$, each mapped to a Message Passing Interface (MPI) process (known as rank) $P_i$. This architecture lends itself naturally to decentralised RL: a local agent $\mathcal{A}_i$ is assigned to each rank $P_i$, where it observes a state $s_i^t \in \mathbb{R}^d$ from prognostic (state variables) and selects actions $a_i^t$ (e.g., adjusting physical parameters such as the convective entrainment rate) according to a policy distribution $\pi_{\theta_i}(a \mid s_i^t)$. In this formulation, each subdomain can develop regime-specific control strategies (policies), reflecting the fact that different parts of the domain experience distinct climates. At every timestep, the $i$-th agent receives a local reward, \(r_i^t = -\| \phi_i^{\text{model}}(t) - \phi_i^{\text{target}}(t) \|^2\), where $\phi_i$ is a diagnostic (observed or reference quantity) of interest (e.g., temperature profile or radiative flux), thus encouraging improved accuracy relative to observations or high-resolution reference simulations at the regional scale. This decentralised design ensures scalability via parallel learning while requiring only minimal modifications to the structure of the physics in the dynamical core. {Although the regions are not assumed to be independent and remain dynamically coupled through the host numerical model, spatial decomposition is still considered useful as it mirrors the existing GCM structure and allows multiple local policies to be explored in parallel, with periodic aggregation enabling information sharing.}

To stabilise training and enforce global consistency, this decentralised setup is extended using a federated reinforcement learning (FedRL) framework~\cite{jin_federated_2022}. In this approach, each agent $\mathcal{A}_i$ on rank $P_i$ interacts with its local environment $\mathcal{E}_i$, collects trajectories, and updates its parameters $\theta_i$ using policy gradients. After $K$ episodes, the agents synchronise through a global averaging step, \(
\theta^{\text{global}} = \tfrac{1}{N} \sum_{i=1}^N \theta_i \), and the aggregated parameters are broadcast back so that $\theta_i \leftarrow \theta^{\text{global}}$ for all $i$. This procedure, shortened as the {FLAG} cycle (short for {\underline{F}ine-tune \underline{L}ocal, \underline{A}ggregate \underline{G}lobal}), repeats every $K$ episodes: agents perform local fine-tuning updates for $K-1$ episodes, followed by global synchronisation on the $K^{\text{th}}$ episode. In this way, the framework balances the benefits of regional specialisation and global coherence. For climate modelling, regime-aware policies that remain globally coherent are generally preferable, since GCMs apply the same physics everywhere and must generalise under climate change (such as differing responses in warm and cool regimes). For global NWP, however, varying degrees of regional specialisation can be useful, for e.g., differentiating between tropics and high latitudes where grid-cell sizes and dominant processes differ.

This federated aggregation step accelerates convergence by allowing agents to benefit indirectly from information learnt in other regions, while still preserving data locality since each agent updates solely from its own trajectories without requiring a full-state exchange. This design closely parallels the structure of coupled GCMs, in which the dynamical core advances the state evolution while the physics package, representing sub-grid processes, interacts with it non-linearly~\cite{gross_bridging_2016}. By maintaining this separation of concerns, the FLAG cycle improves sample efficiency and preserves global consistency, making the approach scalable and well-suited for HPC environments, thus potentially enabling practical integration of RL-assisted parametrisations into production-grade numerical weather prediction (NWP) and climate systems.

\subsection{climateRL Environments}

\subsubsection{Simple Climate Bias Correction Model (SCBC)}
\label{sec:methods-rlenvs-scbc-model}

The Simple Climate Bias Correction (SCBC) environment defines a scalar temperature adjustment model in which the agent applies a heating control term \( u(t) \) to steer the model temperature \( T(t) \in \mathbb{R} \) towards a prescribed {constant} observational reference \( T_{\text{observed}} = 321.75~\text{K} \). The system evolves over discrete timesteps \( t \in \{0, 1, \dots, T_{\text{final}}\} \), with the temperature updated in three steps.  

In the first step, the current temperature is combined with a relaxation term that pulls the state towards a given {constant} background physical temperature \( T_{\text{physics}} \):
\begin{equation}
T^{\ast}(t+1) = T(t) + \varepsilon_1 \cdot \left( \frac{T_{\text{physics}} - T_{\text{current}}(t)}{T_{\text{physics}} - T_{\text{observed}}} \right),
\end{equation}
where \( T_{\text{current}}(t) = T(t) \) and \( \varepsilon_1 = 0.2 \) regulates the strength of this physical relaxation.

A second step introduces a bias-correction relaxation toward the target observation, yielding:
\begin{equation}
T^{\ast\ast}(t+1) = T^{\ast}(t+1) + \varepsilon_2 \cdot \left( \frac{T_{\text{observed}} - T^{\ast}(t+1)}{T_{\text{physics}} - T_{\text{observed}}} \right),
\label{eq:methods-rlenvs-scbc-bias-correction}
\end{equation}
with \( \varepsilon_2 = 0.1 \). This term acts as a nudging mechanism that applies a correction proportional to the residual discrepancy between the intermediate and observed states.  

A third step adds a control mechanism using the heating control term \( u(t) \) for the RL agent to perform corrections towards the observed model temperature \( T_{\text{observed}} \) and create the final update:

\begin{equation}
T(t+1) = T_{\text{new}}(t) = T^{\ast\ast}(t+1) + u(t),
\end{equation}

Both the first and the second steps, utilise the normalisation factor \( (T_{\text{physics}} - T_{\text{observed}})^{-1} \), ensuring that the updates scales consistently with the gap between the physical model and observations.

To improve numerical stability and prevent large temperature magnitudes from dominating the learning signal, all temperature values are normalised by first subtracting the freezing point of water (\( 273.15 \,\text{K} \)) and then scaling by a factor of \( \tfrac{1}{100} \):
\begin{equation}
T_{\text{observed}} \leftarrow \frac{T_{\text{observed}} - 273.15}{100}, \quad
T_{\text{physics}} \leftarrow \frac{T_{\text{physics}} - 273.15}{100}.
\end{equation}
Since all temperature variables are now scaled by a factor of 100, the agent’s control action \( u \) (representing an additive heating term) is restricted to the interval \( [-1, 1] \). This constraint ensures that perturbations remain physically realistic relative to the rescaled temperature state.

\textbf{Single-agent Environments}\\

\noindent The SCBC (\texttt{scbc}) environments (simulation interfaces) are implemented as a Gymnasium (Gym)~\cite{brockman_openai_2016, towers_gymnasium_2024} environment, providing a standardised RL interface for testing an agent’s ability to correct systematic bias in a simplified programmable setup. The model simulates the evolution of temperature over 200 timesteps, with updates governed by the three-step formulation described above. Three variants are designed to progressively challenge the agent under different reward formulations, as described below:

\texttt{scbc-v0}:~In this baseline variant, the reward at each timestep is defined in terms of the squared value of the bias-correction introduced:
\begin{equation}
r(t) = - \left[ \left( \frac{T_{\text{observed}} - T_{\text{new}}(t)}{T_{\text{physics}} - T_{\text{observed}}} \right) \cdot \varepsilon_2 \right]^2.
\end{equation}
This reward explicitly reflects the model’s nudging step and incentivises the agent to minimise the bias-correction required to track the observation.

\texttt{scbc-v1}:~Here, the bias-correction step in Eq.~\eqref{eq:methods-rlenvs-scbc-bias-correction} is disabled (\( \varepsilon_2 = 0 \)), and the reward is defined simply as the negative squared error between the current state and the observational target:
\begin{equation}
r(t) = - (T_{\text{observed}} - T_{\text{current}}(t))^2.
\end{equation}
This design removes dependence on the model’s internal bias term, yielding a more intuitive and stable training signal while requiring the agent to implicitly discover the bias-correction step.

\texttt{scbc-v2}:~The third variant extends \texttt{v1} by introducing temporal sparsity. The reward signal is provided only every five timesteps, with a constant penalty assigned otherwise:
\begin{equation}
r(t)= 
\begin{cases}
- (T_{\text{observed}} - T_{\text{current}}(t))^2, & \text{if } t \bmod 5 = 0, \\
-1, & \text{otherwise}.
\end{cases}
\end{equation}
This sparse feedback setup tests the agent’s ability to handle delayed rewards and encourages the agent to learn strategies that remain robust over extended horizons.

\subsubsection{Radiative-Convective Equilibrium (RCE)}

Radiative–Convective Equilibrium (RCE)(such as the one shown in ~\cite{manabe_thermal_1967}) represents the balance between radiative heating and convective transport in a vertical atmospheric column under horizontally homogeneous conditions. The prognostic variable is the temperature profile \( T(z,t) \) over height \( z \in [0,H] \), governed by the thermodynamic energy equation:
\begin{equation}
C_p \rho(z) \frac{\partial T}{\partial t}(z,t)
= - \frac{\partial F_{\text{rad}}}{\partial z}(z,t)
+ Q_{\text{conv}}(z,t),
\label{eq:methods-rlenvs-rce-energy-balance}
\end{equation}
where \( C_p \) is the specific heat capacity at constant pressure, \( \rho(z) \) the density profile, \( F_{\text{rad}} \) the net vertical radiative flux, and \( Q_{\text{conv}} \) the convective heating rate.  

At equilibrium (\(\frac{\partial T}{\partial t}(z,t) = 0\)), radiative flux divergence is balanced by convective adjustment, producing a stable vertical temperature profile. The lower troposphere is maintained near a moist-adiabatic lapse rate through convective adjustment, while the stratosphere remains radiatively controlled. This simplified framework provides an ideal testbed for evaluating parametrisations of vertical energy transport.

\textbf{Single-agent Environments}\\

\noindent The RCE environment (\texttt{rce}) is implemented using \texttt{climlab}~\cite{rose_climlab_2018} on Gym to model a global-mean vertical temperature profile \( T(z) \). Radiative transfer is computed with the Rapid Radiative Transfer Model for GCMs (RRTMG) scheme~\cite{clough_atmospheric_2005, iacono_radiative_2008}, while convective adjustment (implemented as a separate module in \texttt{climlab}) enforces neutral stability by constraining the lapse rate to remain below a critical moist-adiabatic threshold. Within this setup, we introduce an RL agent which modifies a small set of physically interpretable parameters, such as the effective surface emissivity (inside RRTMG) and the critical lapse rate used in the convective adjustment process. The resulting profile is compared against reanalysis climatology, and the agent is rewarded for reducing errors. The reward at time \( t \) is defined as the negative mean-squared error between the simulated temperature profile and the {observed climatology (average of the NCEP/NCAR}~\cite{kalnay_ncepncar_1996}{ monthly mean reanalysis over time, latitude, and longitude)}, \(r(t) = - \frac{1}{N} \sum_{i=1}^{N} \left[ T_{\text{simulated}}(t,z_i) - T_{\text{reanalysis}}(z_i) \right]^2\), where \( N = 17 \) denotes the number of vertical levels. This formulation provides a continuous and physically interpretable signal that encourages the agent to produce profiles consistent with the climatology.

Three variants are tested in the RCE environment, each increasing in complexity by expanding the set of parameters controlled by the agent, as detailed below:

\texttt{rce-v0}:~The baseline environment, where {the agent learns a state-dependent policy to adjust two global scalar parameters}: the effective longwave surface emissivity and the critical lapse rate for convective adjustment. This setup tests whether RL can learn bulk thermodynamic controls on the vertical profile.

\texttt{rce17-v0}:~An extension of the baseline in which the agent learns a {state-dependent policy function to control a} 17-dimensional vector of critical lapse rate values, one for each model layer. This formulation introduces additional vertical structure into the convective adjustment scheme, enabling the agent to represent non-uniform heating profiles.

\texttt{rce17-v1}:~A further extension where the agent {learns to jointly control} the vertical profile of specific humidity along with the critical lapse rate values. This coupling links radiative transfer to water-vapour feedback, creating a more challenging problem that requires the agent to balance moist thermodynamics with radiative equilibrium.

To ensure physical realism and maintain numerical stability, the parameters controlled by the agent are restricted to bounded ranges. The critical lapse rate for convective adjustment is confined to \( \Gamma_{\text{crit}} \in [5.5, 9.8] \ \text{\textdegree C km}^{-1} \), spanning typical moist to dry adiabatic conditions. The effective surface emissivity is limited to \( \epsilon \in [0, 1] \), representing the range from a perfect reflector to a black body. In the most complex variant, the specific humidity profile is bounded within \( q \in [0, 0.005] \ \text{kg kg}^{-1} \), ensuring realistic vertical moisture content. These constraints preserve thermodynamic consistency while preventing the RL agents from exploring non-physical regions of parameter space.

\subsubsection{Energy Balance Model (EBM)}

The Budyko--Sellers EBM~\cite{budyko_effect_1969, sellers_global_1969, north_theory_1975} (\texttt{ebm}) is an idealised, latitude-resolved model designed to represent the zonal-mean surface temperature \( T_s(\phi) \) as a function of latitude \( \phi \). The model captures the balance between absorbed shortwave radiation, outgoing longwave radiation (OLR), and meridional heat transport (described using a downgradient diffusion assumption) governed by a zonal-mean energy balance equation:
\begin{equation}
C(\phi)\frac{\partial T_s}{\partial t} =
\underbrace{(1 - \alpha(\phi))Q(\phi)}_{\text{absorbed shortwave}}
- \underbrace{(A + B T_s)}_{\text{longwave cooling}}
+ \underbrace{\frac{D}{\cos\phi} \frac{\partial}{\partial \phi}\!\left( \cos\phi \frac{\partial T_s}{\partial \phi} \right)}_{\text{diffusive transport}}.
\label{eq:methods-ebm-equation}
\end{equation}
Here, \( C(\phi) \) denotes the effective heat capacity, \( \alpha(\phi) \) the albedo, \( Q(\phi) \) the insolation, \( A \) and \( B \) the OLR coefficients, and \( D \) the meridional heat transport parameter. These parameters are usually prescribed as constants tuned to match observations. The EBM is discretised into $N = 96$ latitude bands for numerical integration of the temperature field. Within the FedRL framework, \( A \) and \( B \) are treated as adaptive control variables and are learnt as {state-dependent policy outputs} of the RL agents. The reward is defined as the negative mean squared error (MSE): ~\(r(t) = - \tfrac{1}{N} \sum_{i=1}^{N}\Big[ T_{\text{simulated}}(t,\phi_i) - T_{\text{observed}}(\phi_i) \Big]^2\), between simulated and observed climatology {(average of the NCEP/NCAR monthly mean reanalysis}~\cite{kalnay_ncepncar_1996}{  over time and longitude)} to encourage learning control strategies that reduce regional biases. 

\textbf{Single-agent and multi-agent Environments}\\

\noindent Four environments (with progressive complexities) are developed from the Budyko--Sellers EBM (schematics in Figures~\ref{fig:app-methods-ebm-v01-dataflow}-\ref{fig:app-methods-ebm-v3-dataflow}, constraints in Table~\ref{tbl:rce-ebm-parameter-ranges}). Each environment introduces additional spatial decomposition or coupling complexity, providing a hierarchy of testbeds that gradually resemble the the distributed nature of full GCMs, described below:

\texttt{ebm-v0}:~Implemented using \texttt{climlab} and Gym, this baseline configuration uses a single agent that observes the full temperature profile and learns {a state-dependent policy function to control} five global scalar parameters: \(A\) and \(B\) (OLR coefficients), \(\alpha_0\) and \(\alpha_2\) (albedo terms), and \(D\) (diffusive heat transport) applied uniformly across all latitudes. 

\texttt{ebm-v1}:~This variant extends \texttt{ebm-v0} by allowing \(A\) and \(B\) to vary with latitude. The agent now learns {a state-dependent policy function to control} a 96-dimensional vector for each parameter,  observing the full temperature profile, enabling spatially localised adjustments to radiative properties. 

\texttt{ebm-v2}:~In this FedRL configuration, the latitudinal domain is divided into non-overlapping regions, each governed by a distinct RL agent operating on its own \texttt{climlab} EBM instance. Each agent observes the full temperature profile but learns {to control} region-specific values of \(A\) and \(B\). Local rewards are computed via MSE, and agents synchronise after every \(K\) episodes using FedRL (implemented in \texttt{flwr} with orchestration through SmartSim). This setup enables learning specific to conditions in local regimes while maintaining coordinated updates across regions. 

\clearpage

\begin{table}[!h]
\centering
\caption{Parameter ranges used in the \texttt{rce-v0/rce17-v0/v1} (top-3) and \texttt{ebm-v0/1/2/3} (bottom-5) experiments}
\begin{tabular}{l l c c}
\toprule
\textbf{Parameter} & \textbf{Description} & \textbf{Range} & \textbf{Canonical Value} \\
\midrule
$\Gamma_{\text{crit}}$ & Critical lapse rate      & $[5.5, \, 9.8] \  \text{\textdegree C km}^{-1}$ & 6.5 \ $\text{\textdegree C km}^{-1}$ \\
$\epsilon$ & Surface emissivity      & $[0, \, 1]$ & 1 \\
$q$ & Specific humidity      & $[0, \, 0.005] \ \text{kg kg}^{-1}$ & - \\
\midrule
$A$        & OLR intercept        & $[140, \, 420] \ \mathrm{W\,m^{-2}}$  & 210 $\mathrm{W\,m^{-2}}$ \\
$B$        & OLR slope            & $[1.95, \, 2.05] \ \mathrm{W\,m^{-2}\,^\circ C^{-1}}$ & 2 $\mathrm{W\,m^{-2}\,^\circ C^{-1}}$ \\
$\alpha_0$ & Albedo baseline      & $[0.3, \, 0.4]$ & 0.354 \\
$\alpha_2$ & Albedo amplitude     & $[0.2, \, 0.3]$ & 0.25 \\
$D$ & Diffusive transport     & $[0.55, \, 0.65]$ & 0.6 \\
\bottomrule
\end{tabular}
\label{tbl:rce-ebm-parameter-ranges}
\end{table}

\texttt{ebm-v3}:~Designed to more closely reflect operational GCMs, this variant introduces a central \texttt{climlab} parent process that integrates the global EBM state. Subdomain RL agents act on latitudinal subregions and transmit their updated \(A\) and \(B\) OLR parameters to the parent, which advances the full simulation and broadcasts the updated temperature profile back to each agent. Albedo parameters  \(\alpha_0\), \(\alpha_2\) and the diffusive transport parameter \(D\) are transmitted too but spatially averaged by the parent due to limitations in the \texttt{climlab} implementation. Communication is handled by SmartSim with SmartRedis, while synchronisation of the policy network weights is performed via FedAvg after every \(K\) episodes. Together with a decentralised architecture, this setup simulates RL–numerical model coupling with realistic spatial hierarchy and efficient inter-process communication.

\clearpage

\section{Results and Discussion}
\label{sec:results_and_discussion}
\subsection{Experimental Outline}

\begin{figure}[!h]
    \centering
    \includegraphics[width=0.9\textwidth]{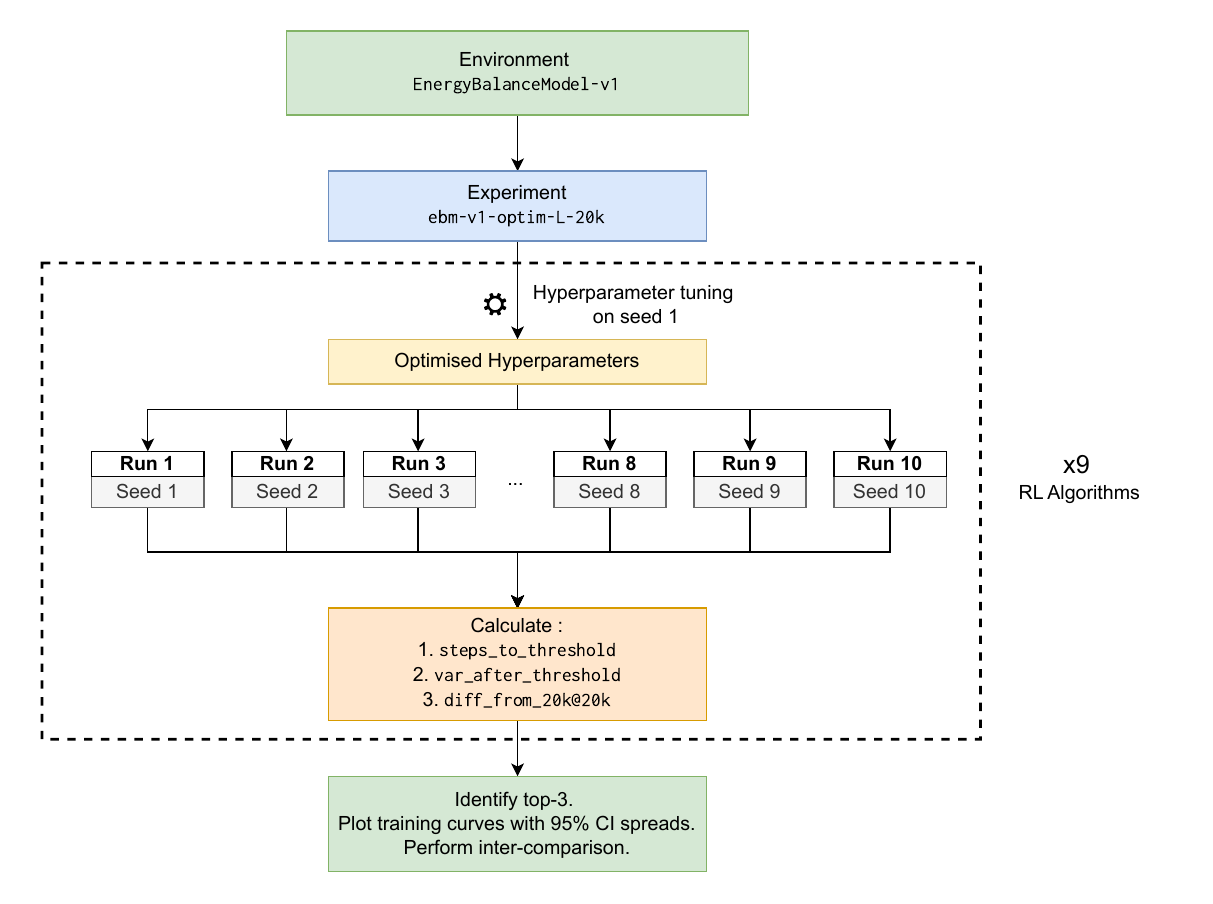}
    \caption{Schematic of the experimental workflow for the single agent climateRL experiments (e.g., \texttt{ebm-v0/1}). The process begins with hyperparameter tuning on seed~1, followed by evaluation across 10 random seeds and evaluation metric computation (steps-to-threshold, variance-after-threshold, and final return difference) for all nine algorithms. Top-3 algorithms are then selected and training curves analysed with 95\% confidence intervals.}
    \label{fig:results-ebm-v01-experiment-flow}
\end{figure}

\begin{figure}[!h]
    \centering
    \includegraphics[width=\textwidth]{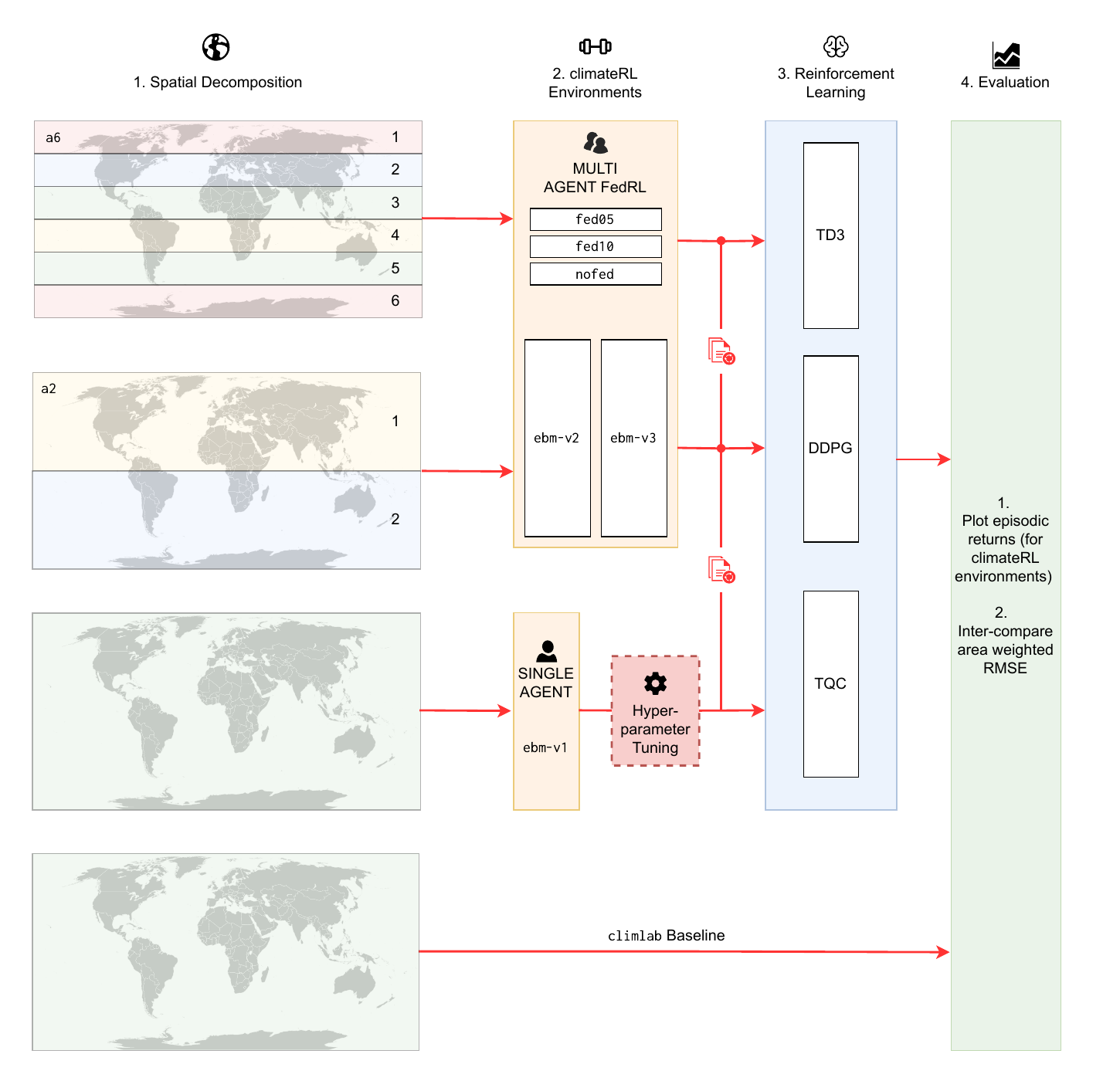}
    \caption{Pipeline for the \texttt{ebm-v1/2/3} experiments. The process begins with configuring the Budyko–Sellers EBM in either single-agent (\texttt{ebm-v1}) or spatially decomposed multi-agent forms (\texttt{ebm-v2}, \texttt{ebm-v3}) using two (\texttt{a2}) or six (\texttt{a6}) regions. Agents are trained with one of three RL algorithms (DDPG, TD3, TQC) under FedRL coordination schemes \texttt{fed05}, \texttt{fed10}, or \texttt{nofed}. In multi-agent settings, policies are periodically aggregated via FedRL every \(K\) episodes. Hyperparameters tuned for \texttt{ebm-v1} are transferred over to \texttt{ebm-v2/v3}. Finally trained models are assessed on their training curves and benchmarked against a static \texttt{climlab} baseline, using a skill measure such as areaWRMSE across 30° latitude groups.}
    \label{fig:results-ebm-v23-experiment-flow}
\end{figure}

A structured four-step protocol is followed to ensure robust and reproducible evaluation of RL-assisted parametrisation schemes. The process begins with hyperparameter tuning, followed by multi-seed training, post-training skill assessment, and finally FedRL evaluation. An overview of this workflow is provided in Figures~\ref{fig:results-ebm-v01-experiment-flow} and \ref{fig:results-ebm-v23-experiment-flow}.

\textbf{1. Hyperparameter Optimisation:}~Hyperparameters for each RL algorithm are tuned scalbly using \texttt{Optuna} with 100 trials on seed 1. The default \texttt{TPESampler}, which implements the Tree-structured Parzen Estimator (TPE) algorithm~\cite{watanabe_tree-structured_2023}, is used. TPE uses kernel density estimation (KDE) to model the distributions of promising and less promising configurations separately, directing the hyperparameter search towards regions likely to yield better performance.{Each trial is run for a fixed number of timesteps }(Table ~\ref{tbl:app-methods-exp-setup-optuna-timesteps}), {which provides a useful measure of computational budget than raw wall-clock time in these idealised and computationally inexpensive environments. In addition, hyperparameter tuning runs are capped at a maximum wall-clock budget of three hours per tuning experiment, with an objective pre-defined as the maximum episodic return in the final episode.} {The best set of hyperparameters is then selected and reused for all subsequent runs in that experimental configuration. Since the actor and critic networks are chosen to be compact, the per-timestep forward and backward pass overheads remain inexpensive.}

\textbf{2. Multi-seed Evaluation:}~To capture robustness and variability, the tuned hyperparameters are then tested across 10 random seeds (1–10). For each seed, the RL agent is retrained from scratch and executed for the full experiment duration, with results recorded independently. Aggregated training curves from these runs are used to rank algorithms, with the evaluation strategies described earlier identifying the top three performers.

\textbf{3. Skill Assessment and Action Interpretability:}~The top-3 algorithms are subsequently evaluated in inference mode, where the trained policy is executed for one episode without updating weights, mimicking deployment in an operational setting. Performance is measured using previously defined metrics, with results reported as mean and standard deviation across 10 seeds for both RCE and EBM environments. In addition, inference trajectories of agent actions are analysed to provide insight into policy behaviour.

\textbf{4. Federated RL and Global Policy Evaluation:}~For multi-agent experiments (\texttt{ebm-v2} and \texttt{ebm-v3}), performance metrics such as areaWRMSE are applied not only to fine-tuned local policies but also to the aggregated global policy obtained at intermediate steps via FedRL. This enables direct comparison (even though having different reward structures), highlighting whether non-local policies yield improved or degraded skill relative to decentralised local policies.

{\textbf{5. Static Baseline:}~The static baseline is chosen as a strong non-adaptive reference point. For \texttt{rce} and \texttt{ebm}, the reference model profile using parameters with the canonical values considered in the Climate Laboratory Book}~\cite{rose_climate_2026}{(also discussed in Table}~\ref{tbl:rce-ebm-parameter-ranges}{) are used. To ensure robustness, the canoncial values were independently verified using \texttt{Optuna} (although not discussed), ensuring that comparisons are made against parameters which are well-tuned instead of arbitrary static choices. The purpose of this static baseline is hence to assess whether state-dependent RL control can improve upon the best fixed calibration available under the same environment and objective.} {This also distinguishes the present RL setting from calibration-focused approaches such as CES, whose primary goal is to infer fixed parameter values or distributions offline rather than to learn a state-dependent control policy that updates parameters sequentially during model rollout.}

\subsection{Single-agent RL}

\subsubsection{SCBC Environment}

\begin{figure}[p]
    \centering
    \includegraphics[scale=0.65]{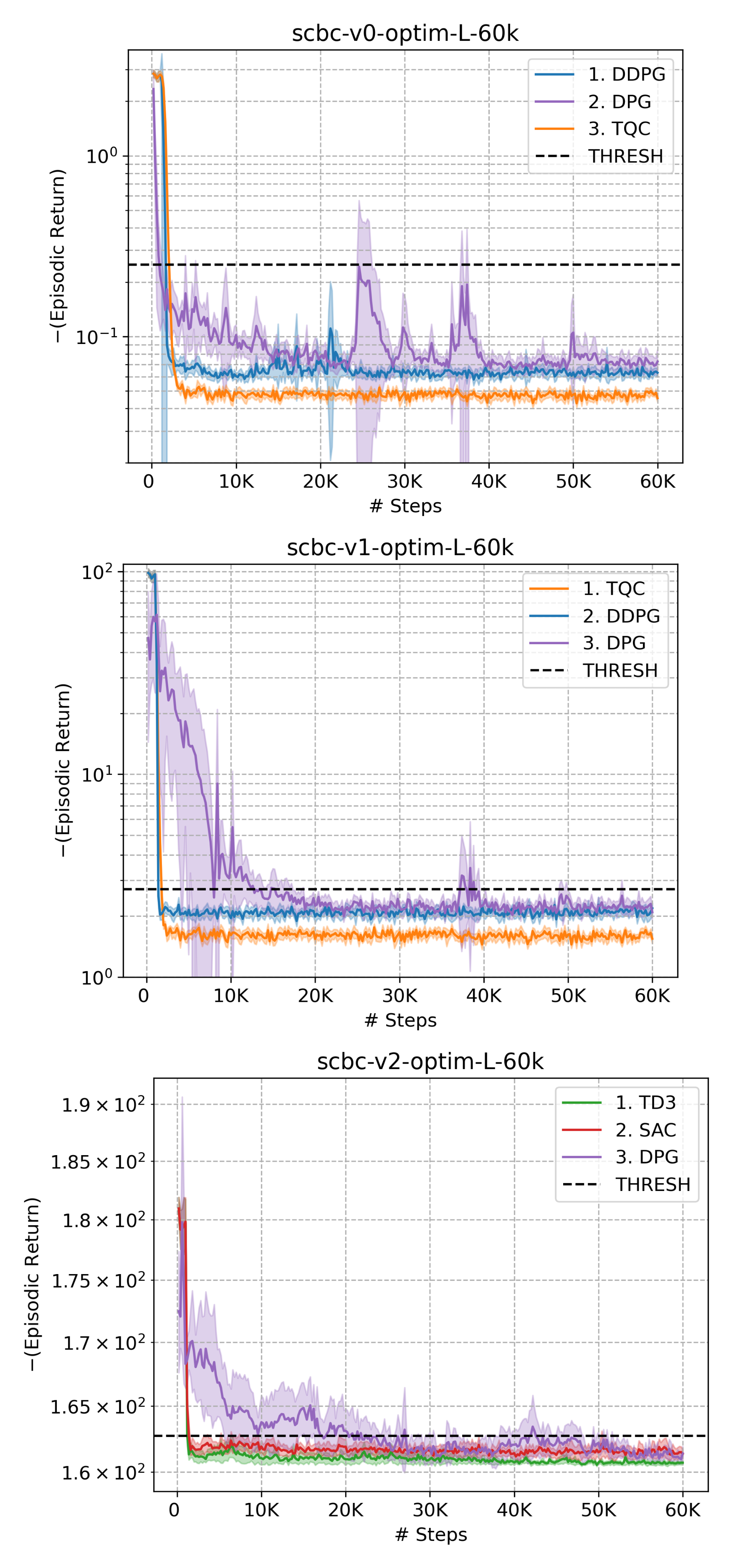}
    \caption{Training curves across 10 seeds for SCBC environments (\texttt{v0} - with bias-correction, \texttt{v1} - w/o bias-correction, \texttt{v2} - with sparse rewards) under the \texttt{optim-L-60k} configuration. Episodic returns (each episode = 200 steps) are plotted on a log scale. Shaded regions denote $\pm$1.96 standard deviation (95\% confidence intervals). The top-3 RL algorithms (DPG, DDPG, TD3, TQC and SAC) in \texttt{scbc-v0/1/2-optim-L-60k} environments are shown. Threshold values are mentioned in Table~\ref{tbl:app-methods-exp-setup-eval-thresholds}. Training curves for other configurations are in Appendix~\ref{app:results-scbc-training}.}
    \label{fig:results-scbc-training-curves}
\end{figure}

\textbf{Training Dynamics}               

Figure~\ref{fig:results-scbc-training-curves} and Appendix~\ref{app:results-scbc-training} presents the training dynamics for all SCBC variants (\texttt{scbc-v0/v1/v2}) under four configurations: \texttt{optim-L}, \texttt{optim-L-60k}, \texttt{homo-64L}, and \texttt{homo-64L-60k}. Rewards generally converge within 10k steps, and longer budgets (60k) do not consistently improve stability. In \texttt{scbc-v2-optim-(L/L-60k)}, TD3 is the only algorithm that reliably ranks among the top-3 across seeds, with other methods exhibiting greater variability. Among different tuning configurations, the \texttt{optim-L} and \texttt{homo-64L} setups converge more quickly and with lower variance.

\begin{table}[!h]
\centering
\caption{Top-3 appearance frequency for each RL algorithm across all SCBC runs (10 seeds)}
\label{tbl:results-scbc-top3-frequency}
\ttfamily
\begin{tabular}{ccc}
\toprule
\textbf{\textrm{Rank}} & \textbf{\textrm{Algorithm}} & \textbf{\textrm{Frequency}} \\
\midrule
1 & \textrm{TD3}  & 10 \\
2 & \textrm{TQC}  & 9  \\
3 & \textrm{DDPG} & 9  \\
4 & \textrm{DPG}  & 6  \\
5 & \textrm{SAC}  & 2  \\
\bottomrule
\end{tabular}
\end{table}

Among the top-performing algorithms, TD3, DDPG, and TQC show rapid initial learning followed by stabilisation. TQC occasionally achieves slightly lower episodic returns in certain configurations, while DDPG and TD3 often track each other closely. By contrast, DPG exhibits large uncertainties at isolated timesteps (suggestive of catastrophic forgetting~\cite{ven_continual_2025}) which undermines its overall reliability. Across the three environment variants, \texttt{scbc-v1} emerges as the most stable. In the \texttt{optim-L}, \texttt{homo-64L}, and \texttt{homo-64L-60k} setups, TD3 and DDPG yield nearly overlapping reward trajectories across all seeds. Overall, TD3 stands out as the most robust and consistently reliable algorithm under the evaluation framework of Appendix~\ref{sec:app-methods-exp-setup-eval-single-agent-RL}, appearing in the top-3 across every seed. This trend is also confirmed in Table~\ref{tbl:results-scbc-top3-frequency}, where TQC and DDPG also feature prominently in the top-3 rankings.

\textbf{Temperature Corrections}

\begin{figure}[p]
    \centering
    \includegraphics[scale=0.65]{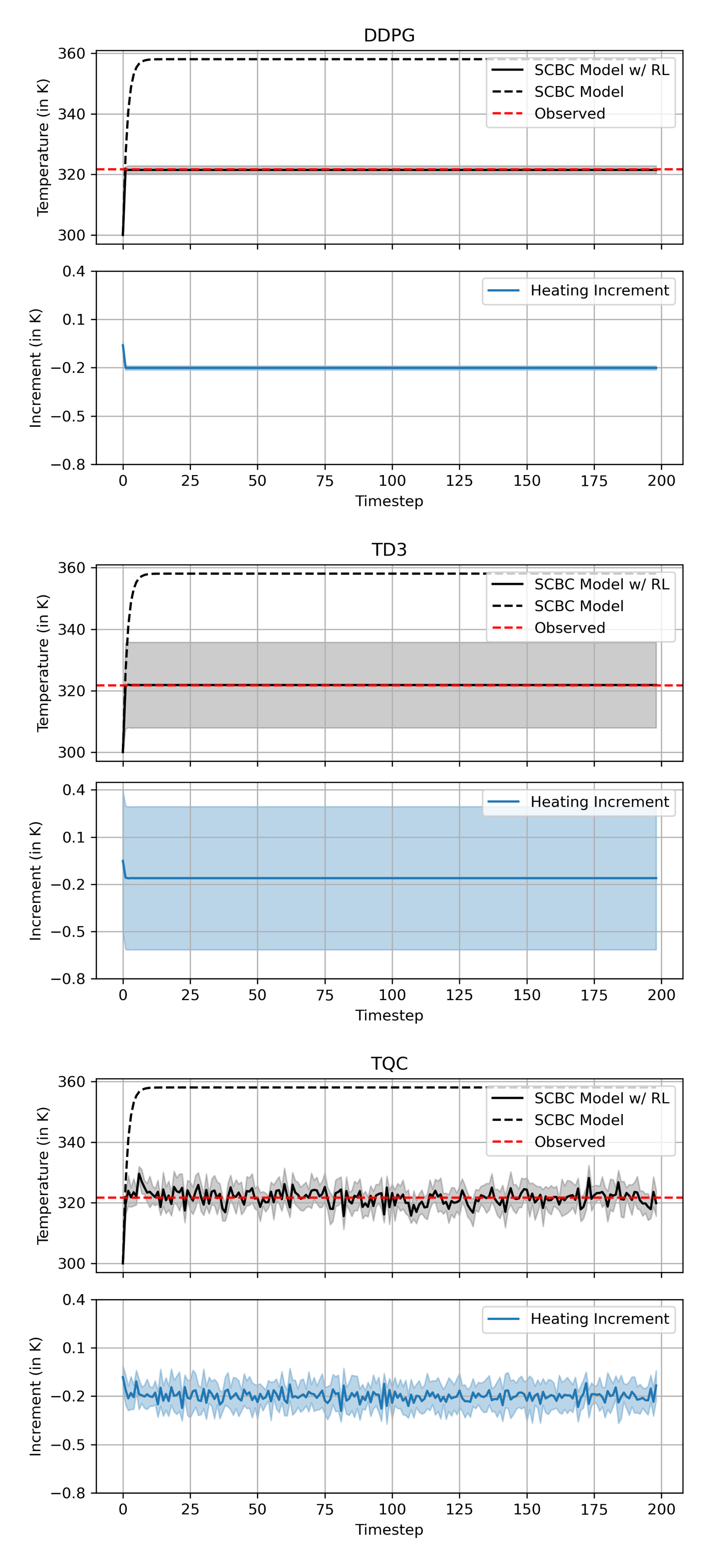}
    \caption{Heating increment dynamics for \texttt{scbc-v0-optim-L-60k}. Top panels: temperature evolution compared to the observed temperature (321.75 K). Bottom panels: normalised heating increments applied by the RL agent. Shaded regions denote $\pm$1.96 standard deviation (95\%). Black dashed line indicates the SCBC model w/ bias-correction as reference. The reference SCBC model stabilised around 358 K driven by the cancellation between the relaxation term and the bias-correction components.}
    \label{fig:results-scbc-action-interpret}
\end{figure}

Figure~\ref{fig:results-scbc-action-interpret} shows the offline skill evaluation of the top-3 RL algorithms: TD3, TQC, and DDPG, trained under the \texttt{scbc-v0-optim-L-60k} configuration. The red dashed line marks the observed target temperature (321.75 K). The control SCBC model without RL and with relaxed physics consistently overshoots the observed target, stabilising at 380 K. Two key observations follow: (1) once the target temperature is reached, maintaining it requires a steady heating increment of roughly –0.2, as can be inferred mathematically from the SCBC dynamics in Section~\ref{sec:methods-rlenvs-scbc-model}, and (2) only TQC and DDPG (not TD3) appear among the top-3 algorithms for this experiment.  

DDPG learns to maintain a stable heating profile centred around -0.2, locking the model output precisely to the target with no variability across seeds. TD3 produces a similar mean profile but with larger, persistent variance in the heating increments. This broader uncertainty band indicates unstable learning, leading some seeds to overshoot the target, a likely reason why TD3 fails to appear in the top-3 in this experiment. TQC, in contrast, adapts its heating actions dynamically while maintaining the observed temperature throughout the episode. The RL-assisted SCBC model closely follows the target temperature with low variance across seeds, indicating that TQC learns a robust, state-aware control policy capable of correcting the bias in the baseline parametrisation. 

These results show that DDPG and TQC demonstrate strong skill in bias correction and heating control. The contrasting behaviour of TD3, despite strong training performance under the \texttt{scbc-v0-optim-L} experiment, highlights the sensitivity of RL algorithms to hyperparameter choices and environment configuration.

\subsubsection{RCE Environment}

\textbf{Training Dynamics}

\begin{figure}[p]
    \centering
    \includegraphics[scale=1.05]{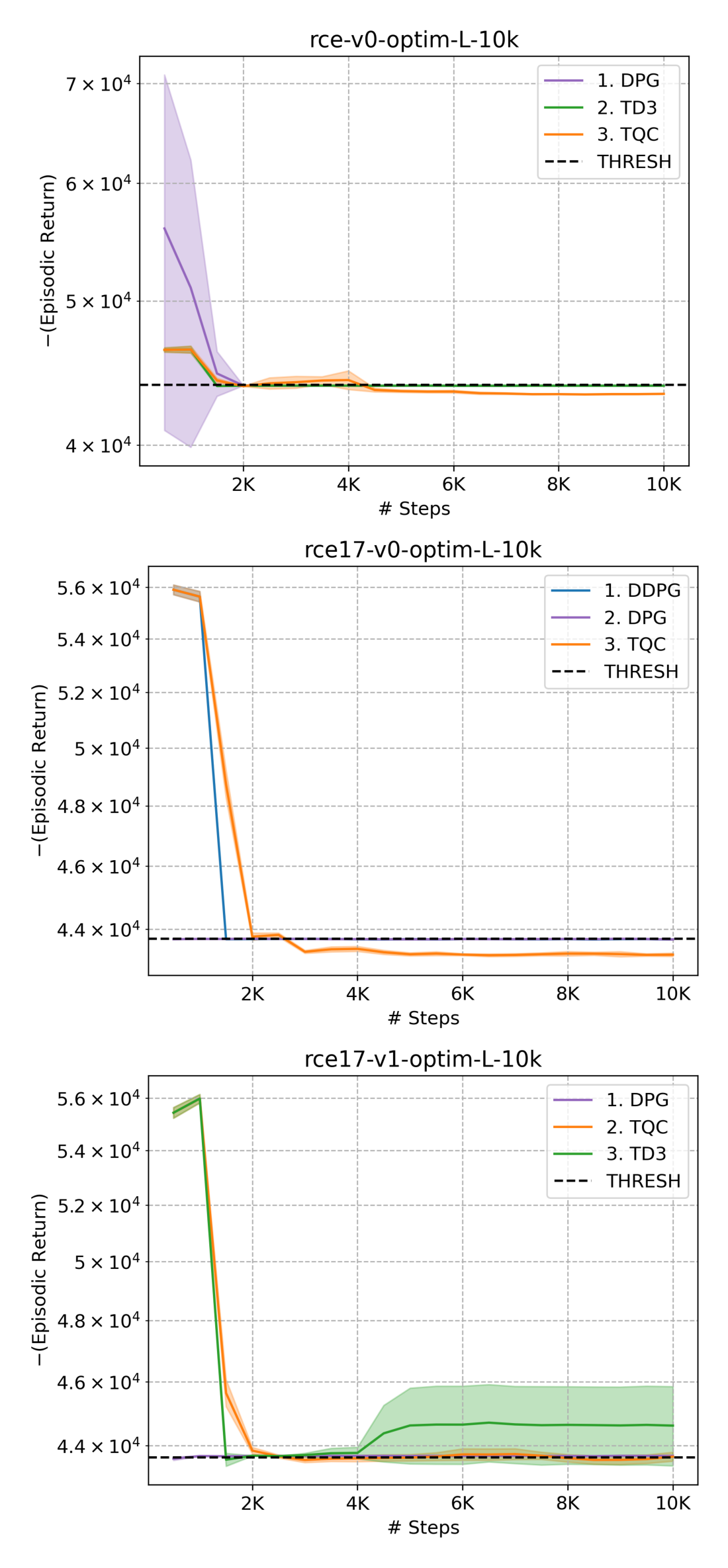}
    \caption{Training curves for RCE environments (\texttt{rce-v0}, \texttt{rce17-v0}, \texttt{rce17-v1}) under the \texttt{optim-L-10k} configuration. Episodic returns (each episode = 500 steps) are plotted on a log scale. Shaded regions denote $\pm$1.96 standard deviation (95\% confidence intervals). Threshold values are mentioned in Table~\ref{tbl:app-methods-exp-setup-eval-thresholds}. DPG, TD3 (\texttt{rce-v0}) and DDPG, DPG (\texttt{rce17-v0}) converge close to the threshold value. Training curves for other configurations are in Appendix~\ref{app:results-rce-training}.}
    \label{fig:results-rce-training-curves}
\end{figure}

Figure~\ref{fig:results-rce-training-curves} and Appendix~\ref{app:results-rce-training} shows the training dynamics across the RCE environments: \texttt{rce-v0}, \texttt{rce17-v0}, and \texttt{rce17-v1}, under four configurations: \texttt{optim-L}, \texttt{optim-L-10k}, \texttt{homo-64L}, and \texttt{homo-64L-10k}. Across all cases, convergence is rapid, typically within 2k–4k timesteps (4-8 episodes). The \texttt{rce-v0} variant displays the most stable post-convergence behaviour, with reduced spread and tight clustering across algorithms. While DPG shows notable early variance in the \texttt{optim-L} setup compared to PPO and TRPO, this instability diminishes with extended tuning. In general, convergence is sharp and variance small across all RCE setups, though some wider spreads occur for DPG in \texttt{rce-v0}, DDPG in \texttt{rce17-v0-optim-L}, and TD3 in \texttt{rce17-v1-optim-L-10k}.  

According to the ranking framework in Section~\ref{sec:app-methods-exp-setup-eval-single-agent-RL}, DPG emerges as the most reliable algorithm, consistently appearing in the top-3 across all seeds and configurations (Table~\ref{tbl:results-rce-top3-frequency}). TQC and DDPG also perform strongly, particularly in the \texttt{rce17} environments, with TQC occasionally edging ahead in episodic return, consistent with its stability and sample efficiency. TD3, which performed well in SCBC, is less reliable in RCE and did not feature in the top-3.  

\begin{table}[!h]
\centering
\caption{Top-3 appearance frequency for each RL algorithm across RCE runs (10 seeds)}
\label{tbl:results-rce-top3-frequency}
\ttfamily
\begin{tabular}{ccc}
\toprule
\textbf{\textrm{Rank}} & \textbf{\textrm{Algorithm}} & \textbf{\textrm{Frequency}} \\
\midrule
1 & \textrm{DPG}  & 10 \\
2 & \textrm{TQC}  & 8  \\
3 & \textrm{DDPG} & 7  \\
4 & \textrm{TD3}  & 5  \\
5 & \textrm{PPO}  & 3  \\
6 & \textrm{TRPO} & 2  \\
7 & \textrm{SAC}  & 1  \\
\bottomrule
\end{tabular}
\end{table}

Overall, the RCE environments present a well-conditioned optimisation landscape characterised by fast convergence and low reward oscillation. However, algorithm rankings are less consistent across configurations than in SCBC, as reflected in the wider spread of top-3 frequencies, suggesting greater sensitivity to hyperparameter variation and environmental noise in the RCE setting.

\textbf{Skill Evaluation}

\begin{figure}[p]
    \centering
    \includegraphics[scale=1.05]{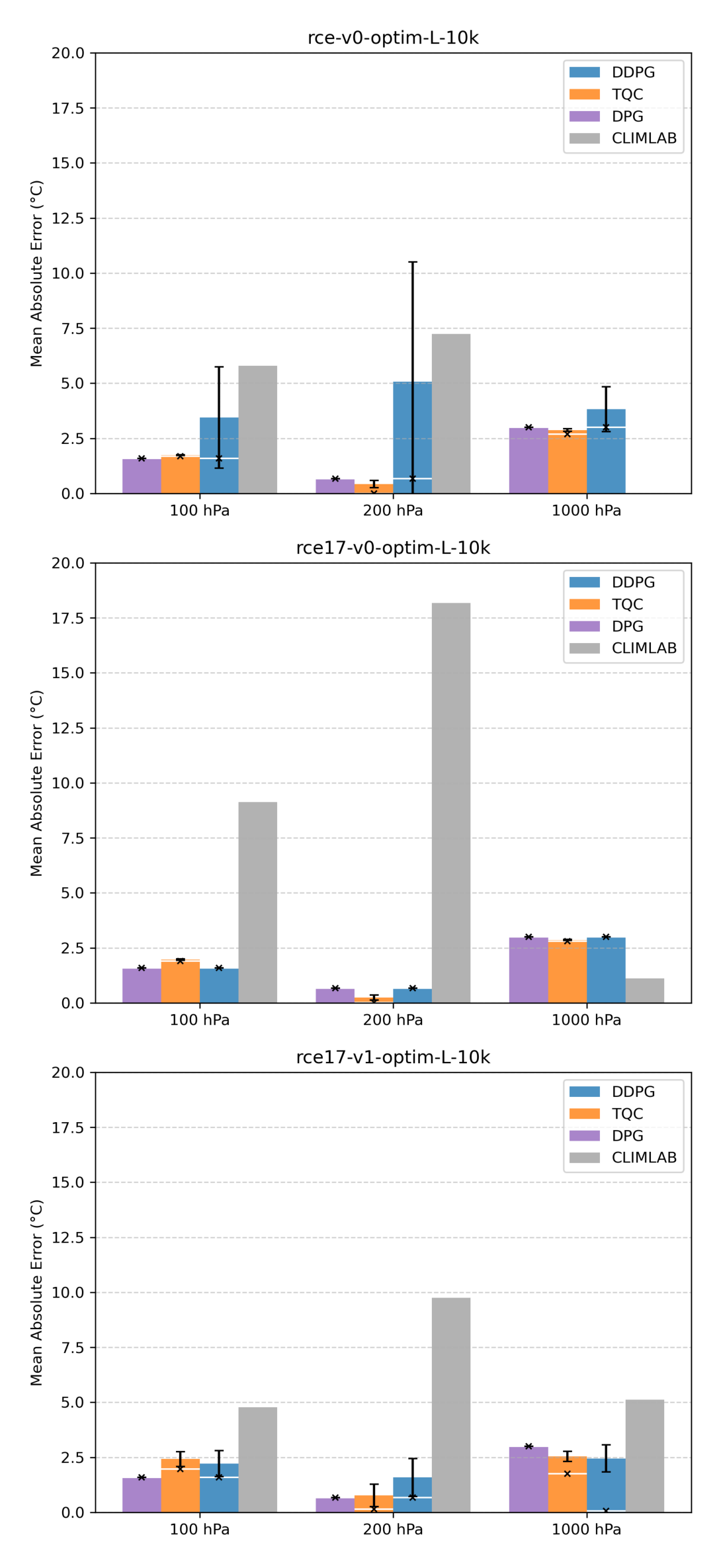}
    \caption{Mean absolute temperature error (°C) at 100 hPa, 200 hPa, and 1000 hPa for the \texttt{optim-L-10k} configuration. White horizontal bars with a cross indicate the best-performing seed for each algorithm. Error bars represent 95\% confidence intervals over 10 seeds. The zero error for the vanilla \texttt{climlab} model at 1000 hPa in \texttt{rce-v0} is with a constant lapse rate set at 6.5 (unlike MALR in \texttt{rce17-v0} and MALR with water vapour coupling in \texttt{rce17-v1}).}
    \label{fig:results-rce-skill-eval}
\end{figure}

Figure~\ref{fig:results-rce-skill-eval} shows the mean absolute error at 100 hPa, 200 hPa, and 1000 hPa for the RCE \texttt{optim-L-10k} configuration. The baseline \texttt{climlab} RCE model exhibits substantial temperature biases, particularly at 200 hPa, where errors exceed 5–17.5°C. By contrast, RL-assisted (DPG, DDPG, and TQC) models consistently show reduced errors across 100 hPa and 200 hPa. At 1000 hPa in \texttt{rce-v0}, the baseline model performs well (relative to the RL-assisted models) with its fixed lapse rate of 6.5°C km\(^{-1}\) unlike in the \texttt{rce17} environments where the moist adiabatic lapse rate (MALR) scheme is applied.

Skill remains robust across different RCE environments and training setups. DPG delivers the most stable performance, with narrow confidence intervals across levels and consistently lowest errors at 100 hPa, outperforming DDPG and TQC, especially where the baseline bias is the second largest. TQC achieves competitive results at 200 hPa, though with slightly wider uncertainty bands. DDPG shows moderate skill, outperforming the baseline in nearly all cases (except at 1000 hPa in \texttt{rce-v0} and \texttt{rce17-v0}), but exhibits greater variance, particularly in the \texttt{rce-v0} and \texttt{rce17-v0} setups. 

All RL algorithms show great skill in reducing errors at 100 hPa and 200 hPa. These levels correspond to the tropopause and mid- to upper troposphere, regions that are highly sensitive to model parametrisations. Improvements at these atmospheric heights suggest that RL agents are capable of learning assistive parametrisation components that reduce biases in predicted variables..

\textbf{Vertical-Level Corrections}

\begin{figure}[p]
    \centering
    \includegraphics[scale=1.05]{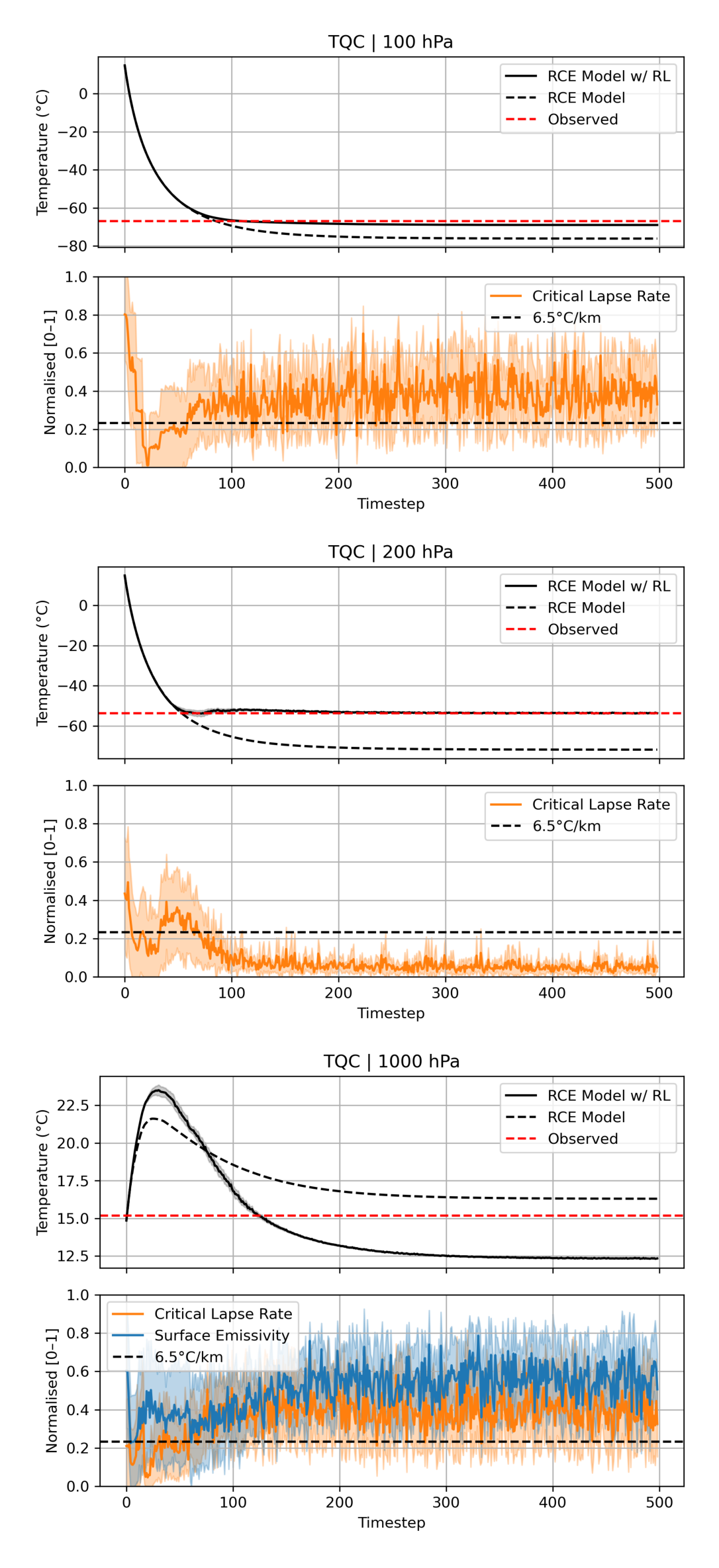}
    \caption{Top: Temperature trajectories at 100 hPa, 200 hPa, and 1000 hPa for TQC in \texttt{rce17-v0-optim-L-10k}. Bottom: Evolution of normalised critical lapse rate and surface emissivity. RL-assisted models closely track observations at 100 and 200 hPa, replacing fixed parameters by values that vary by level and as a function of model state. Shaded regions denote $\pm$1.96 standard deviation (95\%). Canonical value of 6.5~°C km\(^{-1}\) \(\approx 0.23\) in the normalised scale.}
    \label{fig:results-rce17-tel-tqc}
\end{figure}

Figure~\ref{fig:results-rce17-tel-tqc} illustrates the temperature trajectories and the evolution of RL-controlled actions at 100 hPa, 200 hPa, and 1000 hPa for TQC in the \texttt{rce17-v0-optim-L-10k} setup. At 100 hPa, the RL-assisted model diverges (around timestep 100) from the vanilla RCE profile before stabilising near the observed target, accompanied by a steady upward adjustment of the critical lapse rate that reduces vertical mixing in the upper atmosphere. At 200 hPa, typically one of the levels with high bias in the baseline, the agent sustains a low critical lapse rate and maintains a moderate temperature gradient. In both layers, critical lapse rate adjustments mirror the shape of the temperature profile: a sharper curvature at 100 hPa near the tropopause, where the lapse rate values approach the canonical 6.5~°C km\(^{-1}\) (\(\approx 0.23\) in the normalised scale), and a more gradual transition at 200 hPa. At 1000 hPa, however, the agent is less effective in reducing the near-surface bias. Despite jointly adjusting the critical lapse rate and surface emissivity, the model remains offset from observations, with the high critical lapse rate suppressing vertical mixing and the increased surface emissivity (over the timesteps) likely enhancing OLR, altering the surface energy to drive surface cooling. 

This dual-parameter modulation highlights the agent’s attempt to coordinate multiple physical controls, by responding to the changing state of the model rather than converging to a static solution. Yet the persistent surface bias indicates that more sophisticated corrections may be required. In particular, the structural simplicity of the convection scheme may limit its ability to simultaneously match temperatures across all vertical levels, even when parameters are allowed to vary with state. 

\clearpage

\subsubsection{Zonal EBM Environments}

\begin{figure}[p]
    \centering
    \includegraphics[scale=0.9]{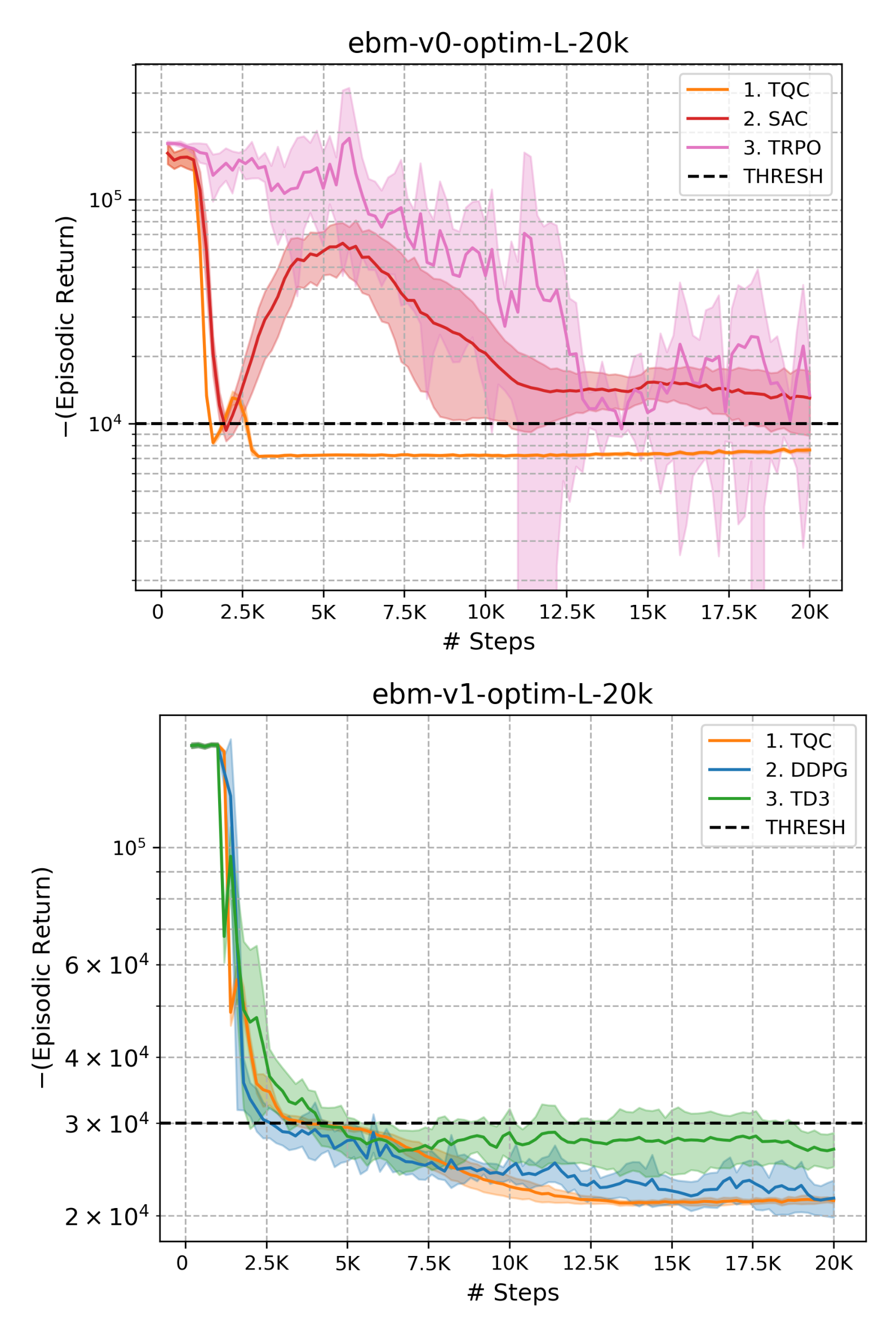}
    \caption{Training curves for single-agent EBM environments (\texttt{ebm-v0}, \texttt{ebm-v1}) under the \texttt{optim-L-20k} tuning regime. Episodic returns (each episode = 200 steps). Shaded regions denote $\pm$1.96 standard deviation (95\% confidence intervals). Threshold values are mentioned in Table~\ref{tbl:app-methods-exp-setup-eval-thresholds}. Training curves for other configurations are in Appendix~\ref{app:results-ebm-v01-training}.}
    \label{fig:results-ebm-v01-training-curves}
\end{figure}

\textbf{Training Dynamics}

Figure~\ref{fig:results-ebm-v01-training-curves} and Appendix~\ref{app:results-ebm-v01-training} presents training dynamics for the EBM environments \texttt{ebm-v0} and \texttt{ebm-v1} under four configurations: \texttt{optim-L}, \texttt{optim-L-20k}, \texttt{homo-64L}, and \texttt{homo-64L-20k}. In contrast to the SCBC and RCE experiments, the EBMs, particularly \texttt{ebm-v0}, display substantially higher variance and less stable convergence, due to their tightly coupled dynamics making small parameter perturbations produce large, temperature changes. Learning generally stabilises only after about 10k steps, with some algorithms (TRPO and SAC in \texttt{ebm-v0}) exhibiting catastrophic forgetting or oscillation during early training. TQC is the only method to maintain a consistently stable profile across all seeds. Increasing the tuning budget from \texttt{optim-L} to \texttt{optim-L-20k} (except in \texttt{ebm-v0}), as discussed in Section~\ref{sec:app-methods-exp-setup-eval-single-agent-RL},  improves both stability and smoothness of convergence, suggesting that the EBM reward landscape demands more extensive exploration than the simpler SCBC and RCE environments.

\begin{table}[!h]
\centering
\caption{Top-3 appearance frequency for each RL algorithm across single-agent EBM runs (10 seeds)}
\label{tbl:results-ebm-single-top3-frequency}
\ttfamily
\begin{tabular}{ccc}
\toprule
\textbf{\textrm{Rank}} & \textbf{\textrm{Algorithm}} & \textbf{\textrm{Frequency}} \\
\midrule
1 & \textrm{TQC}  & 8 \\
2 & \textrm{TD3}  & 6  \\
3 & \textrm{DDPG}  & 4  \\
4 & \textrm{SAC} & 4  \\
5 & \textrm{TRPO}  & 1  \\
\bottomrule
\end{tabular}
\end{table}

\vspace{-0.5cm}

Across all algorithms, TQC emerges as the most robust and consistently reliable, ranking highest in all eight configurations (Table~\ref{tbl:results-ebm-single-top3-frequency}). TD3 achieves moderate performance but exhibits higher variance, while DDPG and SAC struggle to stabilise, particularly under shorter tuning budgets. TRPO, the only on-policy method to feature within the top-3, makes only a marginal contribution, appearing once among the top performers. Performance and stability are strongly influenced by both RL algorithm choice and tuning duration, emphasising the importance of careful tuning for this class of geophysical experiments.

\begin{figure}[p]
    \centering
    \includegraphics[scale=0.9]{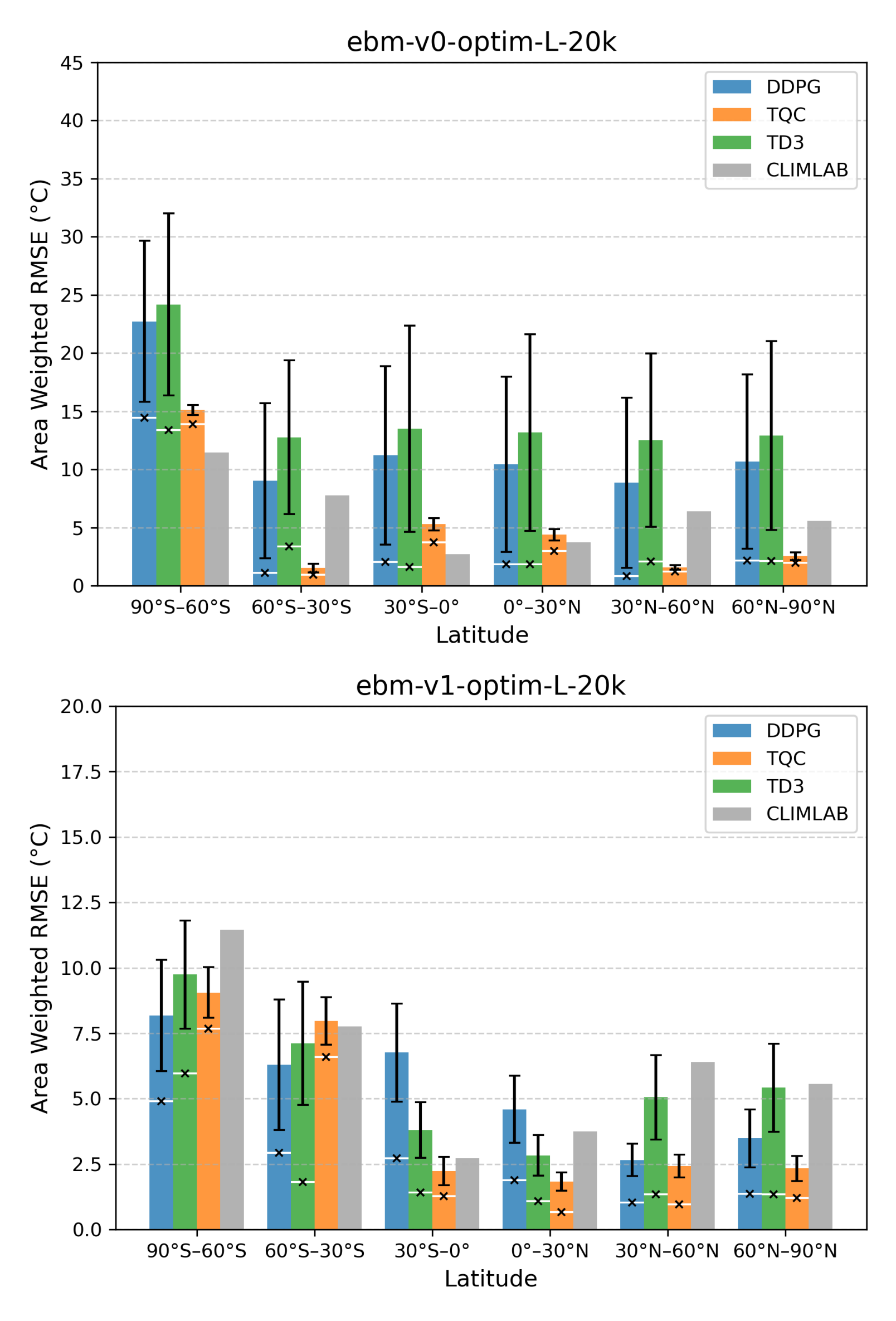}
    \caption{areaWRMSE of zonal mean temperatures across six latitude bands for three experiments under \texttt{ebm-v0} and \texttt{ebm-v1}. Skill is evaluated using areaWRMSE between predicted and reference zonal temperature profiles, averaged with 95\% confidence intervals over 10 seeds. White horizontal bars with a cross indicate the best-performing seed for each scheme.}
    \label{fig:results-ebm-v01-skill-areaWRMSE}
\end{figure}

\textbf{Skill Evaluation}

\begin{figure}[!h]
    \centering
    \includegraphics[scale=0.9]{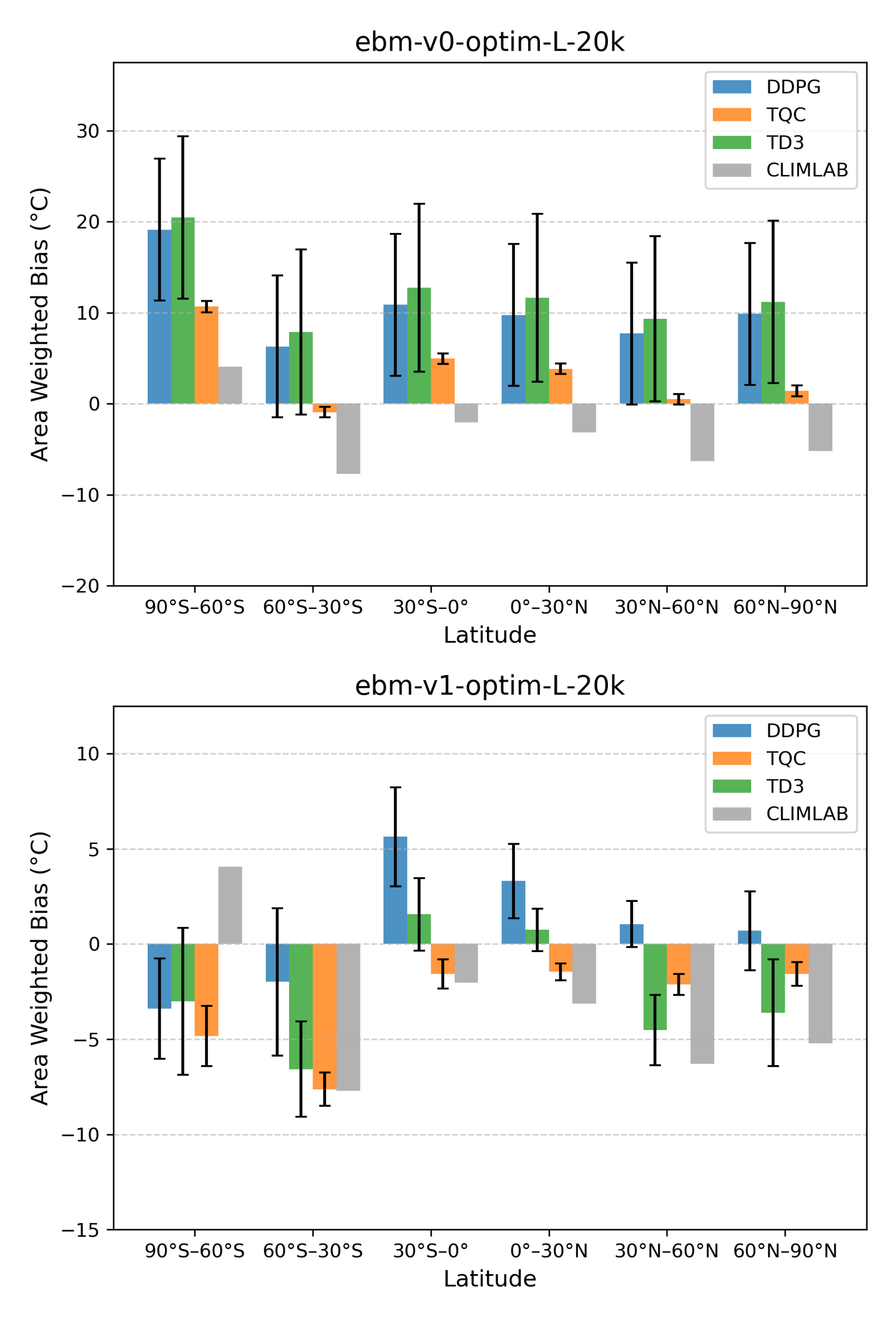}
    \caption{Area-weighted zonal mean temperature bias averaged with 95\% spreads over 10 seeds across six latitude bands for \texttt{ebm-v0} and \texttt{ebm-v1} under top-performing RL algorithms. Negative values indicate underestimation and positive values indicate overestimation of temperature relative to observations.}
    \label{fig:results-ebm-v01-skill-bias}
\end{figure}

Figures~\ref{fig:results-ebm-v01-skill-areaWRMSE} and~\ref{fig:results-ebm-v01-skill-bias} jointly assess EBM skill using areaWRMSE and area weighted zonal mean temperature bias across six latitude bands. The baseline \texttt{climlab} model exhibits particularly large errors around Antartica (90°S–60°S), consistent with its ocean only structure and lack of land representation, generating a biased pattern (due to structural limitations) that is warm over Antarctica and cool elsewhere, with pronounced Southern Ocean warming and widespread cooling biases. RL assisted runs using the best performing seeds reduce areaWRMSE across most latitudes and substantially shrink these biases, with the most consistent gains in \texttt{ebm-v1}, where learning $A$ and $B$ per latitude provides sufficient flexibility to partially compensate for the structural errors over Antarctica. However, the inter-seed spread remains large, emphasising the importance of algorithmic stability and careful hyperparameter tuning in these parameter sensitive settings. Among the top three methods, TQC is the most robust, showing narrow confidence intervals in areaWRMSE and the most uniform bias reductions across zones, whereas TD3 and DDPG typically outperform the baseline but with greater variance, including overshooting in \texttt{ebm-v0} and introducing spurious tropical warming, whilst also showing underestimation tendencies in \texttt{ebm-v1}. Taken together, the areaWRMSE and model bias diagnostics indicate that TQC not only lowers zonal errors but also yields more physically meaningful spatial patterns than either TD3 or DDPG.

\begin{figure}[p]
    \centering
    \includegraphics[scale=0.75]{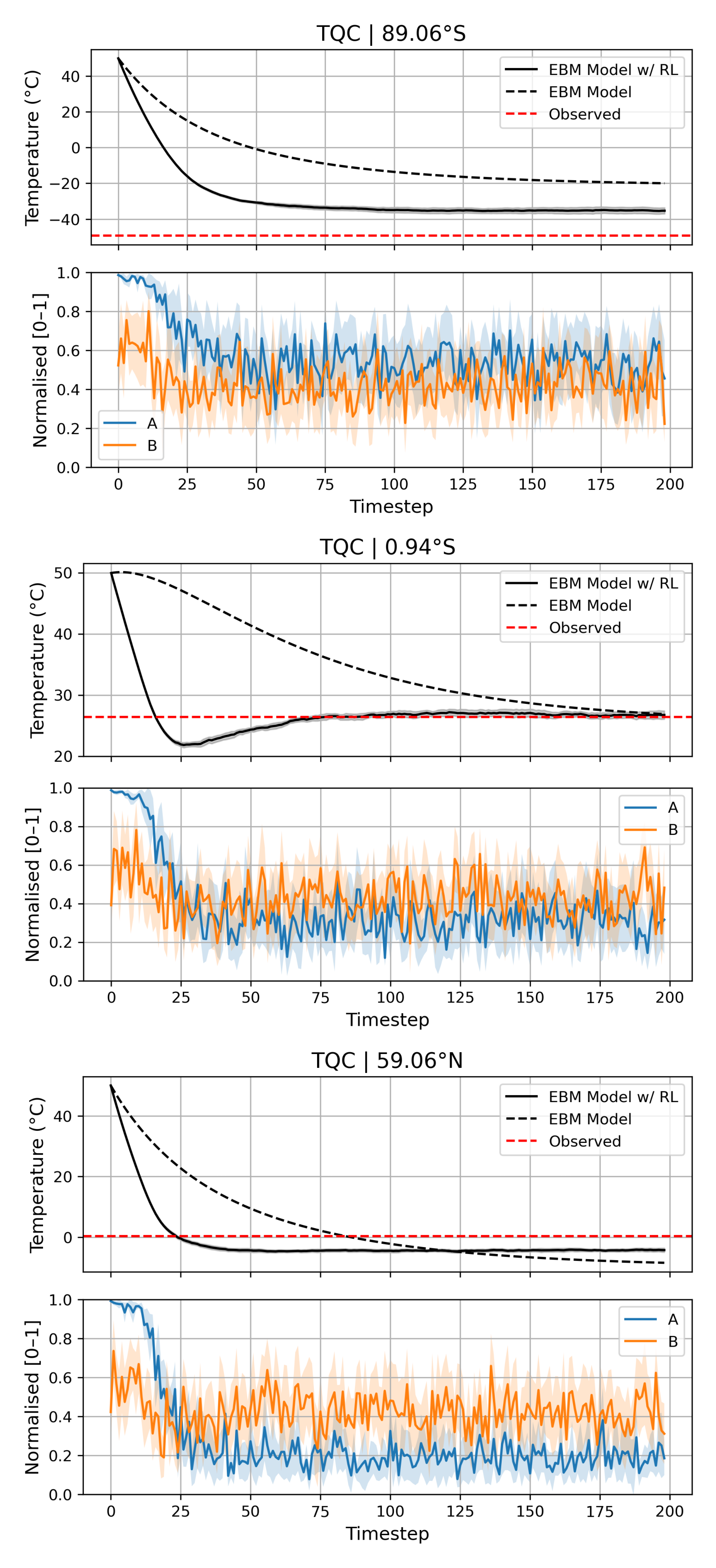}
    \caption{Temperature trajectories (top) and normalised actions (bottom) for parameters \( A \) and \( B \) under TQC at 89.06°S, 0.94°S, and 59.06°N in the \texttt{ebm-v1-optim-L-20k} experiment. Shaded regions denote ±1.96 standard deviation (95\% confidence intervals).}
    \label{fig:results-ebm-latitude-actions}
\end{figure}

\clearpage

\textbf{State-Dependent Variations in $A$ and $B$ at Different Latitudes}

Figure~\ref{fig:results-ebm-latitude-actions} shows the temperature trajectories and the evolution of radiative parameters \( A \) and \( B \) at three representative latitudes: 89.06°S, 0.94°S, and 59.06°N, under the \texttt{ebm-v1-optim-L-20k} configuration with TQC. At the South Pole, the baseline \texttt{climlab} model produces strong warm biases due to structural limitations. The RL agent compensates by stabilising \( A \) and \( B \) between 0.4 and 0.6 (relative to the reference \( A = 0.25 \)), enhancing radiative cooling and reducing bias. At the tropics and mid-latitudes, final biases are smaller: the agent makes a sharp change to \( A \) (in response to the changing global temperature profile) around timestep 25, then gradually refines the parameters, stabilising both \( A \) and \( B \) betweeen 0 and 0.4. Because the model is initialised from a warm isothermal state of 50°C, this early suppression of \( A \) reflects the agent’s attempt to rapidly dissipate excess heat, while \( B \) remains comparatively steady, suggesting that adjustments are driven primarily through \( A \) with minor adjustments of \( B \). These latitude-specific corrections demonstrate the agent’s ability to learn geographically dependent, modulating radiative balance in accordance with regional climate conditions.  

\subsection{Multi-agent RL}

\begin{figure}[p]
    \centering
    \includegraphics[scale=0.70]{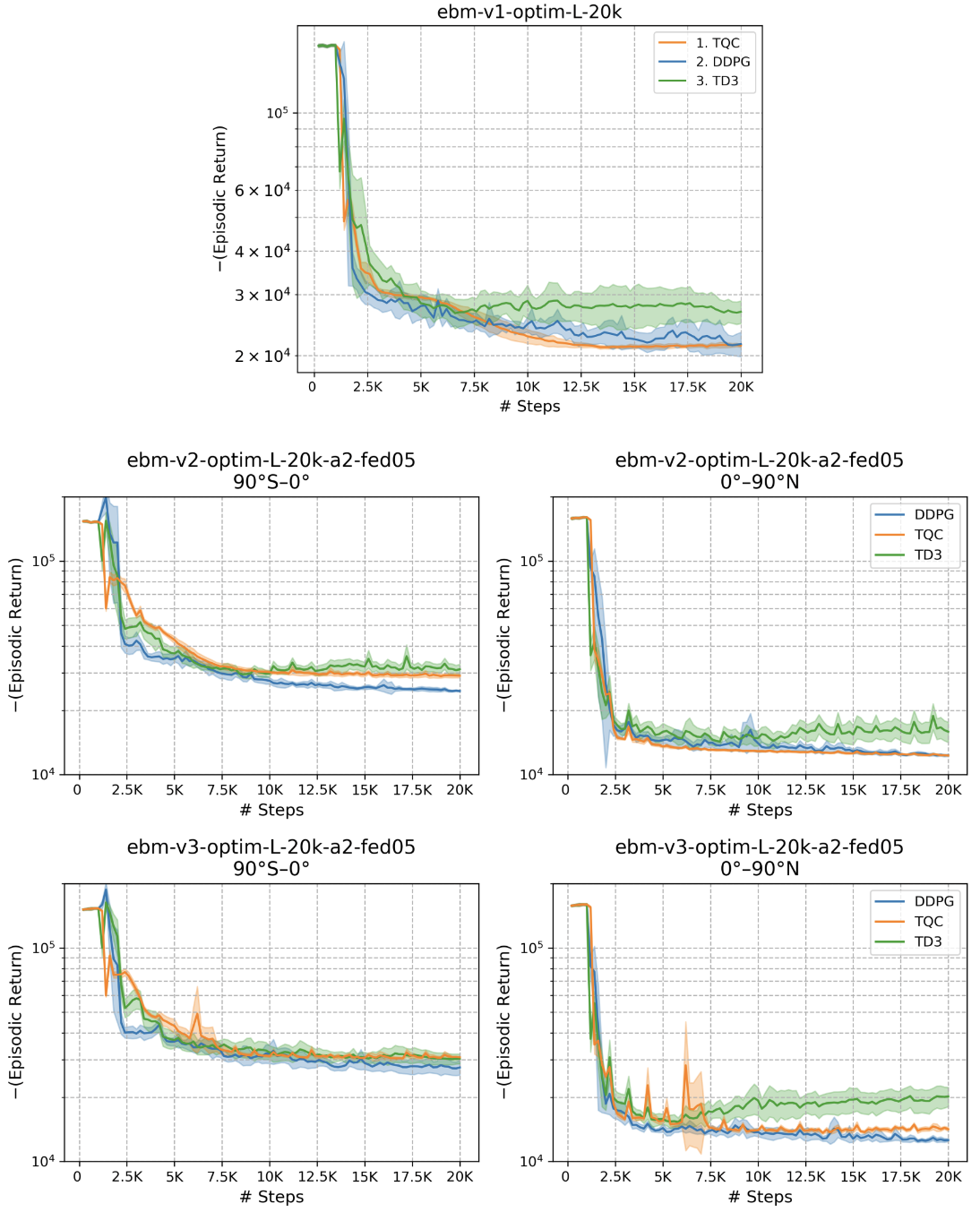}
    \caption{Episodic return curves (log-scaled) with 95\% CI spreads over 10 seeds for three RL algorithms: TQC, DDPG, and TD, across three climateRL environments. Left: \texttt{ebm-v1} (single-agent setup with global input, global reward, and latitude-specific parameters; reproduced from Figure~\ref{fig:results-ebm-v01-training-curves} for comparison). Middle: \texttt{ebm-v2} (multi-agent FedRL configuration with shared global profile input and local rewards). Right: \texttt{ebm-v3} (multi-agent FedRL configuration with partitioned inputs and local rewards, closely resembling GCM-style spatial decomposition). \texttt{ebm-v2/3} training curves are from \texttt{a2} (hemispheric decomposition) with \texttt{fed05} (aggregated every 5 episodes) setting. Threshold not shown for \texttt{ebm-v1}, as top-3 algorithms are already identified. }
    \label{fig:results-ebm-v23-returns}
\end{figure}

\textbf{Training Dynamics}

Figure~\ref{fig:results-ebm-v23-returns} compares training dynamics of multi-agent FedRL in \texttt{ebm-v2/3-optim-L-20k} under \texttt{fed05} with the single-agent baseline \texttt{ebm-v1}. Relative to the global single-agent setup, convergence in the multi-agent \texttt{a2} (hemispheric) configuration is both faster and more stable, with learning stabilising around 5k–7.5k steps compared to slower convergence beyond 10k steps in \texttt{ebm-v1}. In the \texttt{ebm-v2} hemispheric setup, convergence is nearly steady for DDPG, TD3, and TQC, with only minor fluctuations and low variance across seeds. In \texttt{ebm-v3}, DDPG remains the most stable with the least inter-seed variance, while TD3 exhibits moderate fluctuations but trends upward over time. TQC, despite strong performance in single-agent experiments, undergoes catastrophic forgetting mid-training, suggesting greater sensitivity to the increased complexity of multi-agent coordination.

\paragraph{Local Skill Evaluation}

\begin{figure}[p]
\centering
\includegraphics[scale=0.9]{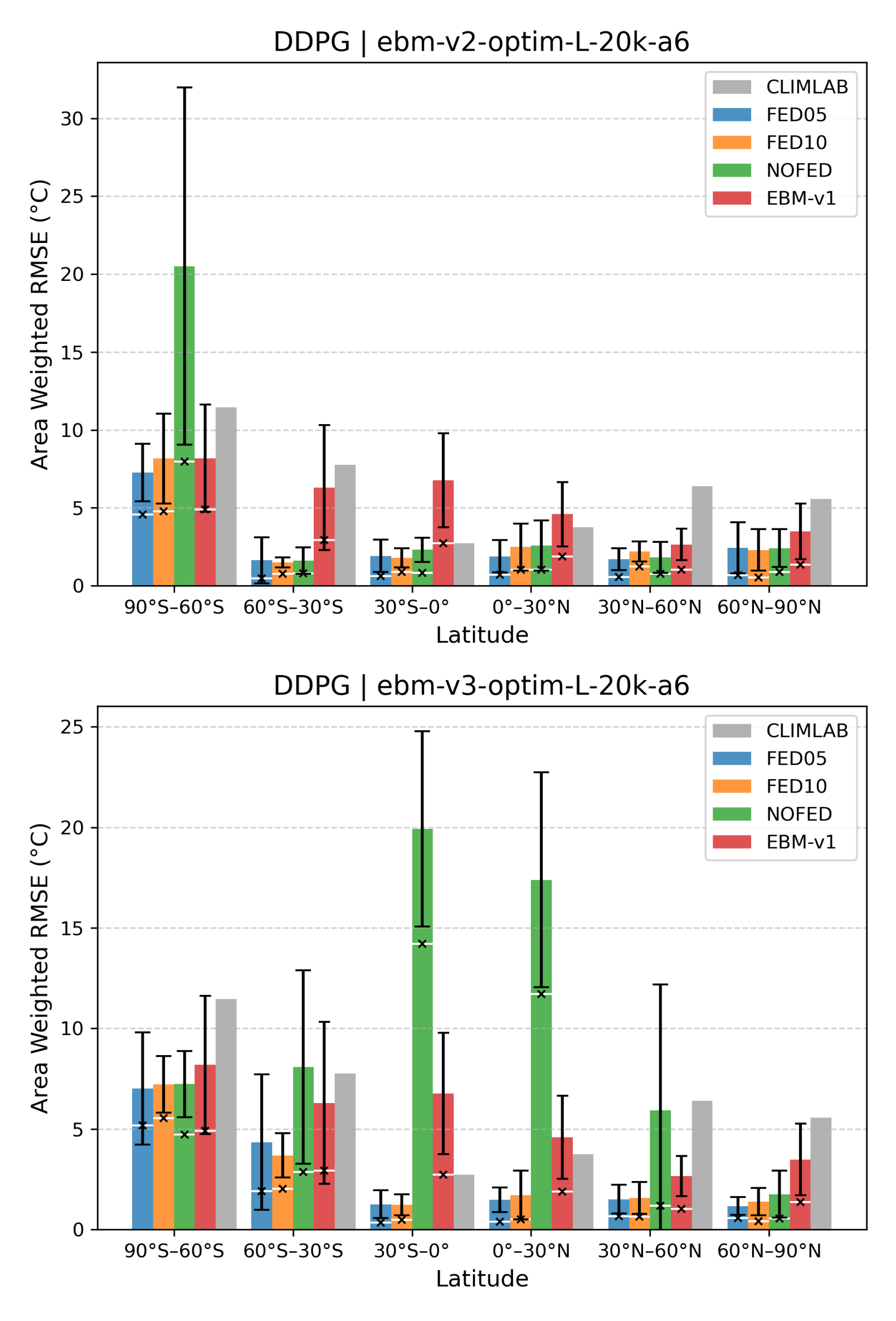}
\caption{Comparison of zonal skill achieved by DDPG under FedRL coordination in \texttt{ebm-v2} and \texttt{ebm-v3}, both using the 6-agent spatial decomposition (\texttt{a6}). Skill is evaluated using areaWRMSE between predicted and reference temperature profiles, averaged with 95\% CI spreads over 10 seeds. Each subplot reports results for three FedRL schemes: \texttt{fed05}, \texttt{fed10}, \texttt{nofed}, along with single-agent \texttt{ebm-v1} and the static \texttt{climlab} baseline. White horizontal bars with a cross indicate the best-performing seed for each scheme. Both setups adopt the same policy network design and hyperparameters as \texttt{ebm-v1}. Results for \texttt{a2} and TD3/TQC in Appendix~\ref{app:results-ebm-v23-ddpg-local-skill}.}
\label{fig:results-ebm-v23-ddpg-local-skill}
\end{figure}

Figure~\ref{fig:results-ebm-v23-ddpg-local-skill} shows that across nearly all latitude bands, \texttt{fed05} outperforms both the static baseline and the non-federated (\texttt{nofed}) setups, with strongest gains in the tropics relative to other best performing alternatives. In both \texttt{ebm-v2} and \texttt{ebm-v3}, areaWRMSE is reduced by more than 50\% in 30°S–0° and 0°–30°N relative to \texttt{ebm-v1}. Improvements are strongest in \texttt{ebm-v3}, with region-specific sliced inputs. By contrast, \texttt{fed10} still improves on \texttt{nofed} but mostly shows higher variance and less consistent benefits than \texttt{fed05} in \texttt{ebm-v2} and \texttt{ebm-v3}, underscoring the importance of frequent aggregation (\texttt{fed05}) for stable coordination. In polar regions, all federated schemes match or surpass \texttt{ebm-v1}, indicating that local specialisation helps to resolve regions like Antarctica better (despite structural limitations). Even under coarse decomposition (\texttt{a2}, Appendix~\ref{app:results-ebm-fedrl-skill-metrics}), DDPG in \texttt{ebm-v2/3} achieves monotonic convergence and low final errors, highlighting its robustness across spatial setups. Overall, these results confirm the benefits of regional specialisation through FedRL and demonstrate DDPG’s stability and efficiency under varying reward structures and input resolutions, making it well-suited to GCM-style architectures. Additional results for TD3 and TQC (Appendix~\ref{app:results-ebm-v23-ddpg-local-skill}) show that while competitive at times, both suffer from higher variance and instability, particularly under frequent aggregation or in equatorial and polar regions.

\textbf{Globally Uniform Policy Skill Evaluation}

\begin{figure}[p]
\centering
\includegraphics[scale=0.9]{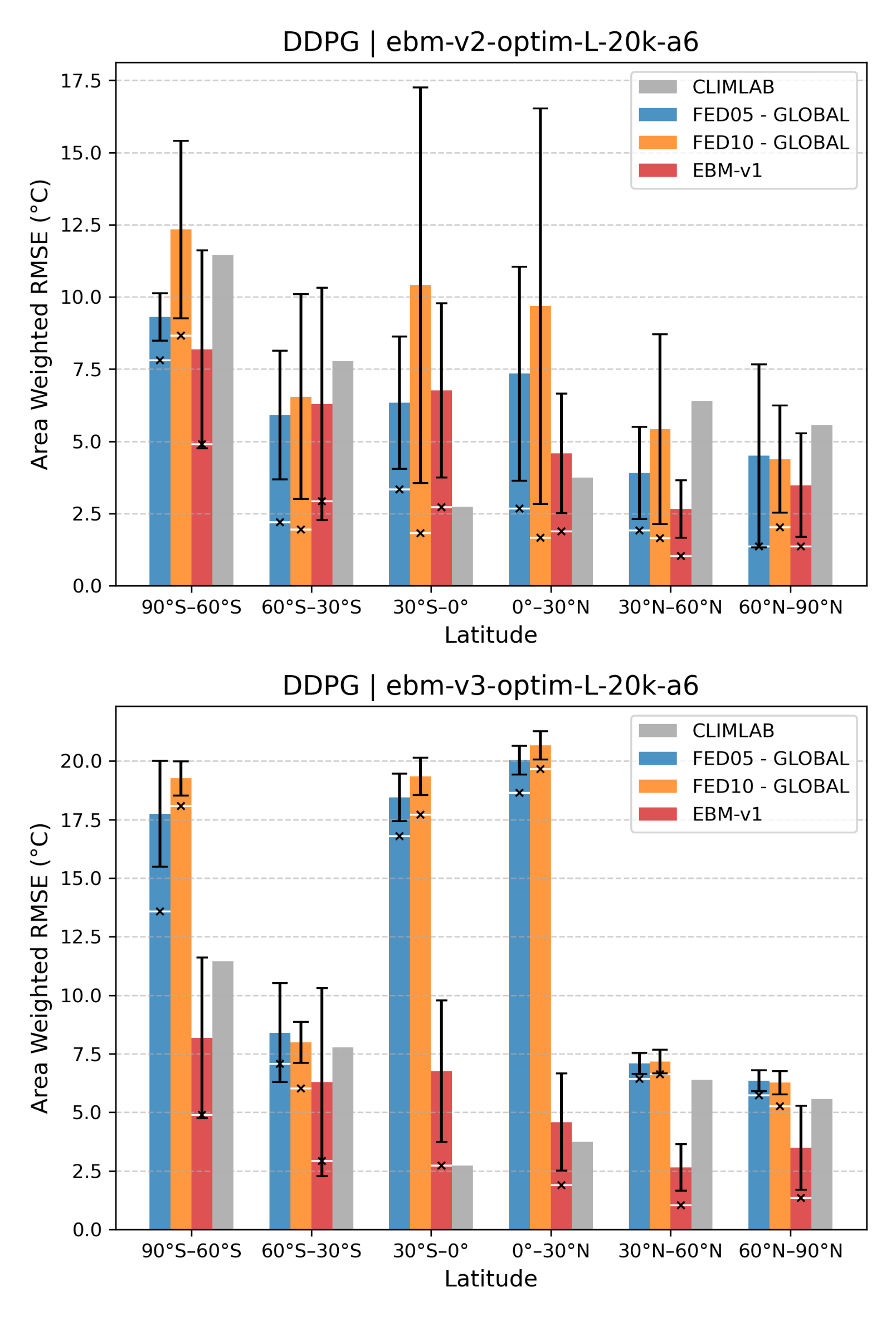}
\caption{Comparison of global policy zonal skill achieved by DDPG under FedRL coordination in \texttt{ebm-v2} and \texttt{ebm-v3}, both using the 6-agent spatial decomposition (\texttt{a6}). Each subplot reports results for three FedRL schemes: \texttt{fed05-GLOBAL}, \texttt{fed10-GLOBAL}, along with single-agent \texttt{ebm-v1} and the static \texttt{climlab} baseline.}
\label{fig:results-ebm-v23-ddpg-global-skill}
\end{figure}

Figure~\ref{fig:results-ebm-v23-ddpg-global-skill} shows inference with globally aggregated non-local policies (\texttt{-GLOBAL}) degrades performance, and collapsing region-specific strategies into a single policy increases both areaWRMSE and inter-seed variance. While DDPG (in \texttt{fed05-GLOBAL}) preserves some regional fine-tuning benefits, these are diminished in the global aggregation, with \texttt{fed10-GLOBAL} performing worse due to infrequent synchronisation. In \texttt{ebm-v2} (see Figure~\ref{fig:app-results-ebm-v23-ddpg-local-skill}), DDPG under the \texttt{a2} setup yields modest gains in the tropics and southern mid-latitudes, but in \texttt{ebm-v3} errors increase further, with robustness declining relative to the \texttt{climlab} baseline. Overall, while FedRL supports effective local specialisation, globally aggregated policy rollouts struggle to reconcile heterogeneous regimes, although more frequent aggregation (\texttt{fed05-GLOBAL}) offers slightly greater stability than \texttt{fed10-GLOBAL}.

\subsection{Discussion}

\subsubsection{Convergence and Stability of RL Algorithms}

Across all experiments, the convergence behaviour of RL agents is strongly environment-dependent. The single-agent SCBC and RCE environments generally yield smoother and more stable training curves owing to their simpler reward landscapes and lack of spatial coupling. By contrast, both \texttt{ebm-v0} and \texttt{ebm-v1} exhibit greater inter-seed variance and require longer training budgets for convergence. This reflects the intrinsic difficulty of EBMs, where each action affects global temperature gradients through diffusive interactions. 

Among the nine RL algorithms evaluated, {TQC} consistently delivers the most stable and robust performance across environments (with the exception of \texttt{a6} in FedRL settings). Its ensemble of quantile critics~\cite{kuznetsov_controlling_2020} provides fine-grained value estimation, enabling stable learning under high-variance returns. This strength is most evident in long-horizon tuning experiments (\texttt{optim-L-20k}) and in federated setups with frequent aggregation (\texttt{fed05} and \texttt{fed10}), where TQC achieves narrow confidence intervals across most latitude bands (with high latitudes being a notable exception). By maintaining a diverse critic ensemble, TQC is better equipped to mitigate noisy gradients and non-stationary policy updates, challenges that are inherent in climate environments with coupled feedbacks.  

{DDPG} also emerges as a strong performer, frequently ranking in the top-3 across SCBC, RCE, and both single- and multi-agent EBM experiments. While it lacks the critic ensemble of TQC, its lightweight actor–critic architecture offers computational efficiency with a smaller memory and compute footprint. DDPG remains competitive under both short and long tuning budgets and shows robustness across spatial decompositions such as \texttt{a2} and \texttt{a6}. Its relative simplicity makes it particularly attractive for large-scale or resource-constrained deployments where training must be distributed across many agents. Unlike TQC, which depends on GPU acceleration to handle ensembles of critics, DDPG offers a more tractable solution for operational climate applications where computational cost and action interpretability are both critical, making it a practical baseline for geophysical RL, especially in federated contexts.  

{DPG} and {TD3} demonstrate promising but less reliable performance. DPG exhibits high inter-seed variance and catastrophic forgetting, especially in early training phases, as seen in the RCE experiments. This behaviour likely reflects its lack of target networks, replay buffers, and deeper critic architectures, which leaves its value estimates sensitive to Q-value errors. Despite these limitations, DPG occasionally achieves top-3 performance, indicating that its simplicity can still yield effective policies under favourable loss landscapes. TD3 improves upon DDPG by introducing twin critics to reduce overestimation bias and clipped noise for stabilising target actions, though at the cost of additional hyperparameters. Nevertheless, TD3 suffers from instability in certain FedRL EBM configurations, particularly in equatorial bands where sharp policy shifts and transient collapses are observed.  

FedRL decomposition further improves stability by reducing the dimensionality and scope of each agent’s learning task. In the \texttt{ebm-v2} and \texttt{ebm-v3} experiments, spatially decomposed multi-agent setups, particularly under the \texttt{a6} configuration, converge much faster than single-agent setups, with most runs stabilising well before 10k steps. This acceleration arises from localised reward signals and simplified policy search spaces when agents operate over narrower latitude bands. Spatial decomposition thus enables more specialised learning through periodic policy aggregation. However, setups with more agents such as \texttt{a6} also introduce higher sensitivity to hyperparameters: both TD3 and TQC show catastrophic forgetting, suggesting that parameters tuned for single-agent global setups may not transfer directly to multi-agent regional FedRL environments with modified inputs and reward landscapes.

\subsubsection{Skill Evaluation Across Latitudes}

Single-agent RL models show heterogeneous skill across latitude bands, shaped by both climatic dynamics and reward structures. In the zonal EBM experiments, tropical and mid-latitude regions consistently exhibit lower bias and narrower confidence intervals, while polar zones (especially 90°S–60°S) show higher variance due to structure limitations in the EBM due to absence of land representations. Algorithms such as TQC maintain robust skill even under these high-variance conditions, whereas TD3 and DDPG display localised instabilities, including overshooting near the tropics and mid-latitudes.  

FedRL enhances zonal skill by allowing agents to specialise in their own regions while synchronising periodically through global aggregation. In \texttt{ebm-v2} and \texttt{ebm-v3}, frequent aggregation (\texttt{fed05}) consistently improves accuracy in tropical and mid-latitude bands, with TQC (in \texttt{ebm-v2}) and DDPG often outperforming their single-agent baselines. Finer decomposition (\texttt{a6}) produces sharper corrections in warmer zones, while coarser setups (\texttt{a2}) yield more stable performance at the poles. In \texttt{ebm-v3}, where input states are restricted relative to \texttt{ebm-v2}, TD3 and TQC show a deterioration in zonal skill compared with \texttt{ebm-v1}, reinforcing the need for environment-specific tuning when reward formulations or inputs differ.  

Deploying a globally aggregated policy during inference introduces a trade-off: while it reduces memory overheads by collapsing policies into a single set of weights, it weakens region-specific adaptations. DDPG emerges as the most stable across both local and global settings, particularly under \texttt{a6}, whereas TQC, though strong in localised fine-tuning, suffers significant performance degradation when globally averaged. These results highlight the tradeoff between specialisation and generalisation in FedRL setups. Overall, federated coordination with well-chosen hyperparameters enables scalable and regime aware learning, outperforming both static baselines and globally trained single-agent RL.

\subsubsection{Physical Interpretability and Alignment}

A key advantage of the proposed RL framework is the physical interpretability of learnt policies, particularly in geophysical environments where actions correspond to parameters within the parametrisation schemes. In the single-agent \texttt{ebm-v1} experiments, agents demonstrated latitude-dependent corrections such as at the poles, where the RL agents increased radiative coefficients \( A \) enhancing OLR-driven cooling and alleviating persistent warm biases. In the tropics and mid-latitudes, parameter values were kept more moderate, consistent with flatter meridional temperature gradients, with warming in mid-latitudes compensating for baseline cooling biases. 

Comparable interpretability is observed in the SCBC and RCE environments. In SCBC, high-performing algorithms such as DDPG and TQC converge to stable heating increments of \(-0.2\), matching the theoretical requirement to maintain the climatological target of 321.75~K. In the RCE setup, agents modulate the critical lapse rate dynamically with altitude, with TQC showing the most consistent adjustments across pressure levels. These behaviours correspond to physically meaningful modifications of vertical mixing and radiative cooling, demonstrating that RL agents, when appropriately constrained, can internalise and act upon thermodynamic balances. These alignments with established physical principles enhances trust in the learnt parametrisations and supports potential integration into operational climate models.

\begin{figure}[p]
\centering
\includegraphics[scale=0.68]{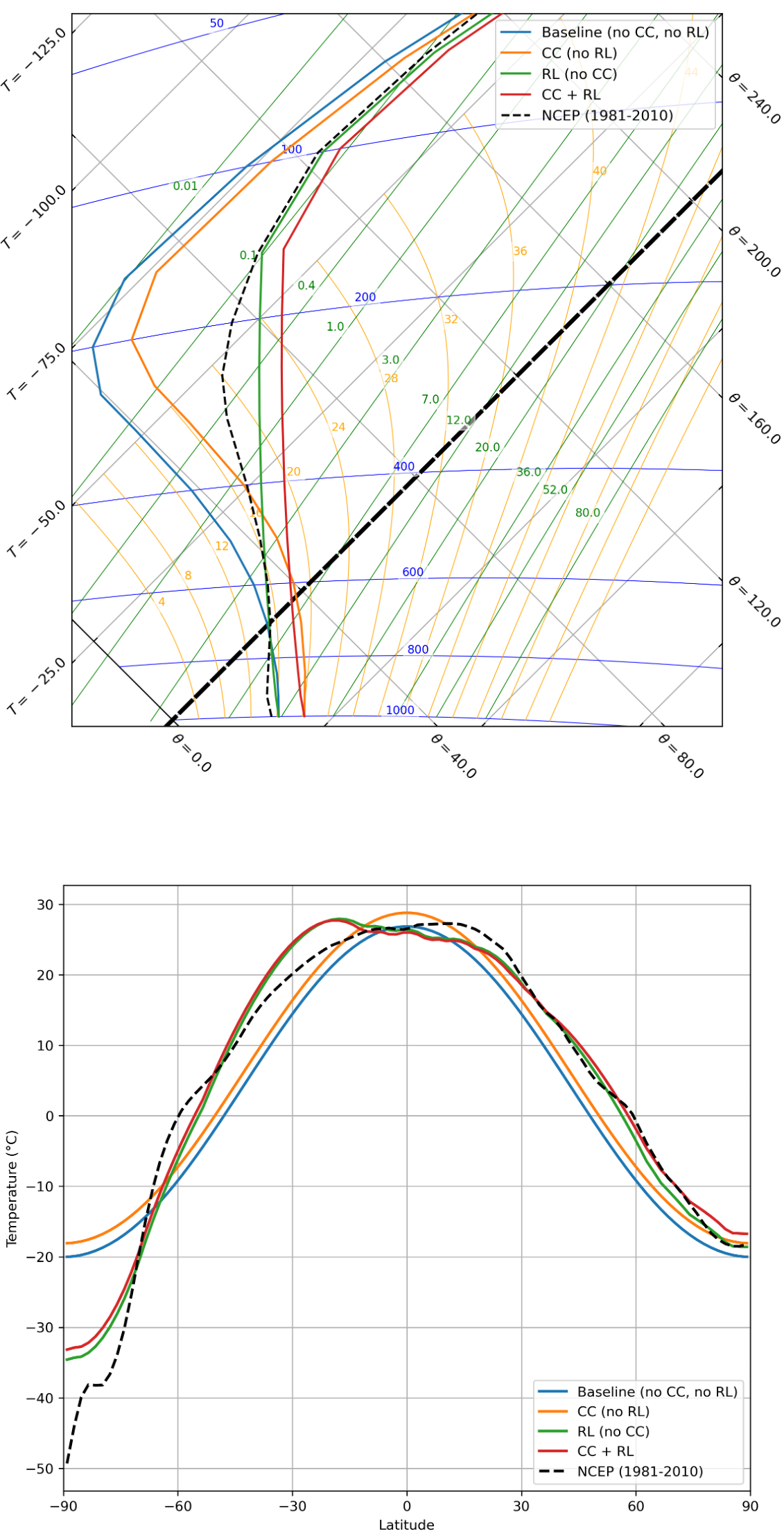}
\caption{{Robustness of learnt RL parametrisation under imposed forcing perturbations. Tephigram (top) for the \texttt{rce17-v0-optim-L-10k} (seed 1) DDPG experiment under a constant +4 K sea-surface-temperature forcing, showing the baseline (\texttt{no CC, no RL}), climate-change only (\texttt{CC, no RL}), RL only (\texttt{RL, no CC}), and combined climate-change plus RL configurations (\texttt{CC + RL}) against the NCEP climatological reference. Zonal-mean surface temperature profiles (bottom) for the \texttt{ebm-v1-optim-L-20k} (seed 3) DDPG experiment under an imposed 4 W m$^{-2}$ forcing, compared across the same four configurations and against the NCEP climatology. \texttt{no CC} denotes the unperturbed control setting, corresponding to 0 K forcing in the RCE and 0 W m$^{-2}$ forcing in the EBM.}}
\label{fig:discussion-cc-rce-ebm}
\end{figure}

\subsubsection{{Robustness under Imposed Forcing Perturbations}}

{As a secondary evaluation, we tested whether the learnt control policies remained physically reasonable under imposed climate-change-like perturbations, using a +4 K sea-surface-temperature perturbation in the RCE and a 4 W m$^{-2}$ forcing in the EBM. These experiments were not included in the main ranking procedure, but were instead designed as a stress test of generalisation under perturbed conditions. The aim was to assess whether policies trained to reduce bias in the control climate still preserve sensible thermodynamic and meridional temperature structures once the forcing is changed. Figure}~\ref{fig:discussion-cc-rce-ebm}~{summarises the response using two complementary diagnostics: the vertical thermodynamic structure on a tephigram for the RCE in the top panel, and the zonal-mean surface temperature response for the EBM in the bottom panel.}

{In the RCE experiment, the perturbed profiles remain within a physically plausible thermodynamic envelope, with no evidence of grossly unrealistic lapse-rate behaviour after warming is imposed. The baseline (\texttt{no CC, no RL}) and climate-change-only (\texttt{CC, no RL}) profiles both depart more clearly from the reference structure through the lower and middle troposphere, whereas the RL-assisted profiles remain closer to the NCEP profile over a substantial part of the column. The combined \texttt{CC + RL} case is particularly notable in that it retains a coherent vertical structure while shifting in the expected warmer direction, suggesting that the learnt control is not simply tuned to one fixed background state but continues to apply dynamically meaningful corrections under perturbation. Although the magnitude of warming is smaller than in the non-RL case, the warming is also more strongly concentrated near the surface than in the upper troposphere. This vertical structure is not usually seen in comprehensive climate models, but may reflect an artefact of the simple model setup rather than a physically realistic climate-change response.} 

In the EBM case, while the non-RL curves show expected warming, the \texttt{CC + RL} and \texttt{RL (no CC)} curves remain especially close to the reference structure in the southern and tropical latitudes, although the northern extratropics still show residual departures. This behaviour suggests that the system response in this idealised setup is less sensitive than would typically be expected in a physical climate model. Since the imposed forcing enters through a perturbation to the OLR intercept term $A$, one possible explanation is that the perturbed values remain within a regime where the controller applies corrections similar to those learnt in the control case, thereby damping the warming response. \add{Another interpretation can be a partial overfit of the policy to the control climatology used in the reward. The learnt control may act like a climatological nudging term, pulling the perturbed simulation back towards the reference state rather than allowing the forced response to emerge.} This ensuing weak response highlights a limitation of the present simple configuration, where the interaction between the external forcing and the learnt control can produce an unrealistically weak climate sensitivity. If similar behaviour were to arise in a more comprehensive climate model, it would require careful scrutiny to differentiate whether the controller was masking a physically meaningful forced response or correcting a genuine model bias. \add{Future reward and policy designs should therefore separate mean-state bias reduction from preservation of physically meaningful climate sensitivity, for example by using anomaly-based rewards, training across multiple forcing regimes, including forced-response targets, or penalising excessive damping of externally imposed signals.}

{Taken together, these experiments provide preliminary evidence that the learnt RL policies can retain physically interpretable behaviour under climate-change-like perturbations rather than collapsing outside the control setting. For the RCE, this is reflected in the maintenance of a realistic vertical thermodynamic structure under warming, while for the EBM it is reflected in the preservation of the broad meridional temperature gradient despite an imposed radiative perturbation to a controlled parameter (although a warming response was desired). Even so, these results are encouraging because they suggest that the learnt policies can remain meaningful under perturbed forcing conditions, an important requirement for future deployment in more realistic weather and climate modelling configurations.}

\subsubsection{\add{Reward Design}}

\add{The reward function is the main mechanism through which scientific objectives enter the MDP formulation. In the present study, rewards are deliberately kept simple because the testbeds are idealised and have clear diagnostic targets: temperature bias in SCBC, vertical temperature-profile error in RCE, and zonal-mean temperature error in the EBM. These relatively ad-hoc rewards are useful at this proof-of-concept stage because they allow us to isolate whether state-dependent policies can learn meaningful parameter updates before introducing more complex multi-objective rewards.}

\add{As this approach is extended to more comprehensive weather and climate models, reward design should move beyond ad-hoc error-based objectives and be co-developed with domain scientists. Such rewards may include terms for forecast skill, top-of-atmosphere energy balance, precipitation statistics, climate drift, conservation residuals, parameter smoothness, and penalties for numerically unstable or physically implausible states. In this progression, simple mean-squared-error rewards provide a useful baseline for controlled experimentation, while more physically informed and multi-objective rewards will be needed for operational applications.}

\subsubsection{Implications for Weather and Climate Model Parametrisations}

The results presented here suggest that RL has significant potential to deliver promising improvements for weather and climate model parametrisations in the future, though confirming its practical value will require substantially more demanding tests in full GCM settings and under altered forcing \change{regimes..}{regimes.} Compared to traditional schemes which depend on static coefficients tuned offline through costly experiments, RL, particularly under federated and spatially decomposed regimes, offers a practical dynamic alternative in which agents adapt parameter values online as a function of the evolving model state. {This should be viewed as complementary to, rather than a replacement for, established calibration approaches such as CES. Methods like CES are especially valuable when the primary aim is offline calibration and uncertainty quantification over static parameter values. By contrast, the main advantage of RL in the present context is that it directly supports online, state-dependent parameter adjustment through sequential interaction with the model. This flexibility, however comes with trade-offs, including a less direct treatment of posterior uncertainty in the current implementation, although providing a natural framework when the parametrisation itself is intended to adapt dynamically to the evolving simulation state.} {These results should therefore be viewed as an early-stage demonstration in controlled idealised environments, providing a tractable hierarchy for testing and screening state-dependent control strategies before moving to more realistic atmospheric models.}

The emergence of geographically differentiated strategies, such as increasing the OLR parameters \( A \) and \( B \) in polar regions to counter persistent warm biases or gradually raising the critical lapse rate near the tropopause (high enough to prevent convective adjustment), demonstrates that physically interpretable parameter control can arise directly from reward-driven optimisation and federated learning which integrates naturally with MPI-based parallelisation used in many models operational at most modelling centres worldwide. The findings also highlight the advantages of federated coordination, where frequent aggregations across distributed learning agents enable robust and scalable learning while retaining local specialisation. This is especially important for resolving spatial heterogeneity in processes such as meridional heat transport and regional radiative balances. Substantial reductions in aWRMSE and improved skill across latitude bands, most notably in the tropics and mid-latitudes, indicate that multi-agent RL can potentially not only match but surpass traditional approaches in accuracy.  

Crucially, the RL-assisted parametrisations do not apply black-box corrections. Instead, they learn neural-network policy functions that set tunable parameters within existing physical parametrisations as a function of the model state, thus improving model performance while remaining embedded within the underlying physics. These policy networks are simple multilayer perceptrons ($\leq$ 200,000 parameters each), making them manageable for future examination with explainable AI techniques to understand how parameter choices vary across regimes. {This sequential, gradient-based policy learning may also become increasingly attractive as the dimensionality of the control policy grows, since learning a neural state-dependent policy through offline ensemble calibration would be expected to require substantially more model evaluations.} As GCMs advance toward higher resolutions and greater process complexity, the integration of interpretable, regime-aware, and online-learnt parametrisations may become a cornerstone of next-generation modelling strategies.

\clearpage

\section{Conclusion}
\label{sec:conclusion}
Climate models are indispensable for simulating the Earth system and making long-term climate projections. Their predictive accuracy, however, is constrained by the inability to resolve small-scale processes such as convection, radiation, and turbulent mixing which in turn contributes to the systematic biases commonly seen when climate model output is evaluated against observations. These processes are represented through parametrisations, simplified formulations with tunable parameters, which are traditionally tuned offline and remain static throughout simulations. This study explored reinforcement learning (RL) as a framework {for learning policy functions that adjust tunable parameters dynamically as a function of the model state, enabling state-dependent control across spatial regimes.} {In this sense, the main modelling shift is from static parameter tuning to learning state-dependent control policies whose outputs replace fixed parameter values during model integration.}

A hierarchy of idealised testbeds, spanning the heating increment simple climate bias correction model (SCBC), radiative convective equilibrium (RCE), and the zonal energy balance model (EBM), was developed to systematically evaluate RL algorithms across increasing levels of physical and spatial complexity. {These environments were intended as interpretable proof-of-concept testbeds for state-dependent tuning, rather than as complete surrogates for chaotic closure-learning systems.} Experiments encompassed both single-agent setups (\texttt{scbc-v0/v1/v2}, \texttt{rce-v0/rce17-v0/v1}, \texttt{ebm-v0/v1}) and multi-agent federated setups (\texttt{ebm-v2/v3}), in which agents operated over regions and synchronised through global aggregation. Nine RL algorithms were benchmarked, with Truncated Quantile Critics (TQC)~\cite{kuznetsov_controlling_2020}, Deep Deterministic Policy Gradient
 (DDPG)~\cite{lillicrap_continuous_2019}, and Twin-Delayed DDPG (TD3)~\cite{fujimoto_addressing_2018} consistently achieving the highest skill and most stable convergence across configurations. Performance was assessed using area-weighted root mean squared error (areaWRMSE), model bias, temperature errors and pressure-level diagnostics (as appropriate), with results compared against a static \texttt{climlab} baseline with parameters fitted via linear regression.

In single-agent setups, RL agents demonstrated superior performance over static parametrisations, particularly in the tropics and mid-latitudes (for the EBM environments). TQC consistently achieved robust skill with narrow confidence intervals across SCBC, RCE, and EBM environments. In RCE, where vertical sensitivity is key, RL agents were able to vary critical lapse rates to reduce biases in temperature profile, while in SCBC, stable heating increments and temperature evolution indicated successful correction of structural drift. DDPG performed competitively across all environments, benefiting from computational simplicity and minimal tuning overhead. TD3, though effective in several experiments, showed local instability in high-variance zones such as equatorial EBM bands.

Federated reinforcement learning (FedRL) enabled regional regime-aware control by decomposing global space into zonal agents with periodic policy aggregation. In \texttt{ebm-v2/3} with DDPG, the 6-agent configuration with frequent updates (\texttt{fed05}) achieved low zonal RMSE in tropical and mid-latitudes, surpassing both \texttt{ebm-v1} and \texttt{climlab}. However, in inference mode (for \texttt{ebm-v2/v3}), the globally aggregated set of weights (used in intermediate FedRL steps) introduced trade-offs, where being deployed as a single non-local policy, it diluted regime-awareness, particularly for TQC in high-gradient regions. DDPG emerged as the most robust algorithm balancing stability and generalisability. RL agents were also found to enact physically meaningful corrections, modulating radiative parameters \( A \) and \( B \) in the zonal EBM to correct for meridional biases, and correcting critical lapse rates in the RCE model to align with the climatological structure, demonstrating their capacity to learn appropriate control strategies. {We note, however, that this evidence for regime-awareness is presently limited to differentiated behaviour across idealised zonal subdomains and does not yet constitute a full test of robustness or generalisation across regimes. In particular, aggregated global policies used at inference can weaken localisation. Evaluating transfer across distinct forcing condition and regional configurations is therefore a natural and important direction for future experiments with higher-fidelity models.}

The results presented here point towards an exciting future paradigm for weather and climate model parametrisations, one that is learnable and aligned with both observational and physical constraints. {By focussing} on idealised heating-increment models, convection parametrised testbeds, and zonal energy balance models, {a key purpose  was to identify which RL algorithms provide the most reliable starting points for progression up the model hierarchy, thereby reducing the search space before moving to substantially more complex and expensive systems. In this regard, DDPG emerged as one of the most consistently strong performers across the test cases and therefore provides a sensible starting point for follow-up experiments where} the core methodology is readily extensible to more complex systems, such as the current Unified Model (UM) and the forthcoming Momentum system at the UK Met Office. Extending the approach to these more complex systems is a natural next step and is already under active development for a follow-up study. Looking ahead, integrating RL-assisted parametrisations into operational models, {with climate invariant states}~\cite{beucler_climate-invariant_2024}, potentially via hybrid interfaces using SmartSim~\cite{partee_using_2022}, FTorch~\cite{atkinson_ftorch_2025} and TorchClim~\cite{fuchs_torchclim_2024}, {with appropriate physics-aware rewards} offers a promising direction for future work. 

In summary, this work shows that RL, when deployed in scalable and federated forms, {can support the design of numerical-model parametrisation components that respond to local state and physical constraints in controlled idealised settings, with future work needed to assess robustness, uncertainty, and behaviour under altered forcing regimes relevant to weather and climate applications.}~As global modelling frameworks transition towards CMIP8, with greater emphasis on high-resolution dynamics and regional-global coupling, the demand for self-learning, flexible parametrisation schemes is set to grow. This study presents a promising prototype for such systems, where RL agents dynamically calibrate model behaviour while implicitly learning ML-based components of existing parametrisation schemes without compromising their physical integrity.

%
%

\section*{Code and Data Availability}
The code for this project and its documentation are available in our GitHub repositories:~\cite{nath_p3jitnathclimate-rl_2024} (\href{https://github.com/p3jitnath/climate-rl}{https://github.com/p3jitnath/climate-rl}) and~\cite{nath_p3jitnathclimate-rl-fedrl_2025} (\href{https://github.com/p3jitnath/climate-rl-fedRL}{https://github.com/p3jitnath/climate-rl-fedRL}). A lightweight easy-to-use API with regressions tests for community use is developed and made available at ~\cite{nath_p3jitnathclimate-rl-fedrain-api_2026} (\href{https://github.com/p3jitnath/climate-rl-fedrain-api}{https://github.com/p3jitnath/climate-rl-fedrain-api}). The climate change sensitivity tests (based on the API) can be found at ~\cite{nath_p3jitnathclimate-rl-cc_2026} (\href{https://github.com/p3jitnath/climate-rl-cc}{https://github.com/p3jitnath/climate-rl-cc}). Data for all experiment runs are available at our Zenodo repository ~\cite{nath_replacing_2025} (\href{https://doi.org/10.5281/zenodo.17116349}{https://doi.org/10.5281/zenodo.17116349}).

The software for this project was developed using Python~\cite{van_rossum_python_2007} on VSCode~(\href{https://code.visualstudio.com}{https://code.visualstudio.com}) and Jupyter Notebooks~\cite{kluyver_jupyter_2016} (\href{https://jupyter.org}{https://jupyter.org}). A number of Python packages have been used including:
\begin{itemize}
\itemsep0em 
\item climlab~\cite{rose_climlab_2018}~(\href{https://climlab.readthedocs.io/en/latest}{https://climlab.readthedocs.io/en/latest}) 
\item gymnasium~\cite{towers_gymnasium_2024}~(\href{https://gymnasium.farama.org/index.html}{https://gymnasium.farama.org/index.html}) 
\item matplotlib~\cite{hunter_matplotlib_2007}~(\href{https://matplotlib.org}{https://matplotlib.org}) 
\item numpy~\cite{oliphant_guide_2006}~(\href{https://numpy.org}{https://numpy.org}) 
\item optuna~\cite{akiba_optuna_2019}~(\href{https://optuna.org}{https://optuna.org})
\item pandas~\cite{reback_pandas-devpandas_2020}~(\href{https://pandas.pydata.org}{https://pandas.pydata.org})
\item ray~\cite{moritz_ray_2018} (\href{https://docs.ray.io/en/latest/index.html}{https://docs.ray.io/en/latest/index.html})
\item ray tune~\cite{liaw_tune_2018}~(\href{https://docs.ray.io/en/latest/tune/index.html}{https://docs.ray.io/en/latest/tune/index.html})
\item stable\_baselines3~\cite{raffin_stable-baselines3_2021}~(\href{https://pypi.org/project/stable-baselines3}{https://pypi.org/project/stable-baselines3}) 
\item tephi~\cite{little_tephi_2014}~(\href{https://tephi.readthedocs.io/en/latest/index.html}{https://tephi.readthedocs.io/en/latest/index.html}) 
\item torch~\cite{paszke_pytorch_2019}~(\href{https://pytorch.org}{https://pytorch.org}) 
\item tyro~\cite{yi_brentyityro_2026}~(\href{https://brentyi.github.io/tyro}{https://brentyi.github.io/tyro}) 
\item xarray~\cite{hoyer_xarray_2017}~(\href{https://docs.xarray.dev/en/stable}{https://docs.xarray.dev/en/stable}) 
\item smartsim~\cite{partee_using_2022}~(\href{https://www.craylabs.org/docs/overview.html}{https://www.craylabs.org/docs/overview.html}) 
\item flower~\cite{beutel_flower_2022}~(\href{https://flower.ai}{https://flower.ai})
\end{itemize} 

Code for RL algorithms (DDPG, TD3, PPO, SAC, TQC) were adapted from the cleanRL~\cite{huang_cleanrl_2022} (\href{https://github.com/vwxyzjn/cleanrl/tree/master/cleanrl}{https://github.com/vwxyzjn/cleanrl/tree/master/cleanrl}) project repository. TRPO was adapted in the cleanRL style by 
Yuhua Jiang~(\href{https://github.com/Jackory}{https://github.com/Jackory}). DPG, REINFORCE and AVG were adapted in the cleanRL style by Pritthijit Nath. This manuscript was prepared using \LaTeX~(\href{https://www.latex-project.org}{https://www.latex-project.org}) on Overleaf~(\href{https://www.overleaf.com}{https://www.overleaf.com}).

\noindent \textbf{LLM Usage Disclosure} \\
The authors acknowledge the use of AI language models, specifically ChatGPT~(GPT-5.2 and GPT-4o \href{https://chatgpt.com}{https://chatgpt.com}), during the preparation of this work. These tools were used to polish language usage and improve the overall clarity of the manuscript, as well as to assist with designing plotting code for the graphs. All AI-generated content was reviewed, verified, and edited by the authors to ensure accuracy and appropriateness.

\acknowledgments
P. Nath was supported by the {UKRI Centre for Doctoral Training in Application of Artificial Intelligence to the study of Environmental Risks} [EP/S022961/1] (\href{https://ai4er-cdt.esc.cam.ac.uk}{https://ai4er-cdt.esc.cam.ac.uk}). Mark Webb was supported by the Met Office Hadley Centre Climate Programme funded by DSIT. This work used JASMIN, the UK’s collaborative data analysis environment (\href{https://www.jasmin.ac.uk}{https://www.jasmin.ac.uk}) managed by UKRI NERC and STFC. We thank Andrew Shao and Alessandro Rigazzi (HPE) for their valuable assistance with setting up SmartSim on JASMIN. {We are also grateful to Tapio Schneider and to three anonymous reviewers whose comments on the manuscript led to substantial improvements.}

\noindent \textbf{Conflict of Interest Statement} \\
The authors declare that they have no conflict of interest.

%
%

\bibliography{bibliography_zotero}

\clearpage

\begin{appendix}

\label{app:additional_methods}
\section{Additional Methods}

\subsection{Schematic Diagrams}

\textbf{Energy Balance Model (EBM) Environments}

\begin{figure}[!h]
\centering
\includegraphics[width=\linewidth]{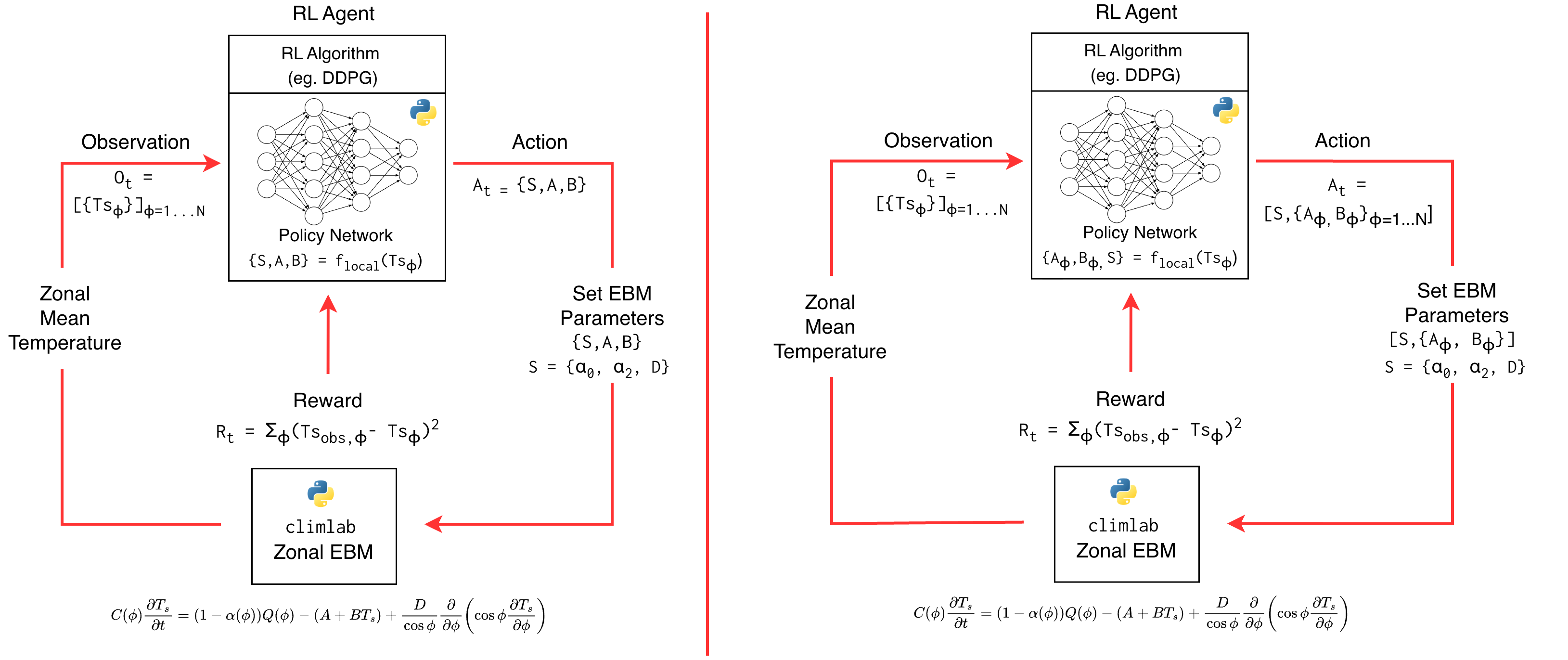}
\caption{\texttt{ebm-v0} (left) and \texttt{ebm-v1} (right) single-agent setup. In \texttt{ebm-v1}, the global agent observes the full zonal-mean temperature profile and outputs latitude-dependent OLR parameters \( \{A_\phi, B_\phi\} \)  as well as independent ones \( \{\alpha_0, \alpha_2, D\} \). Loss from observations are computed over all 96 latitudes.}
\label{fig:app-methods-ebm-v01-dataflow}
\end{figure}

\begin{figure}[!h]
\centering
\includegraphics[width=\linewidth]{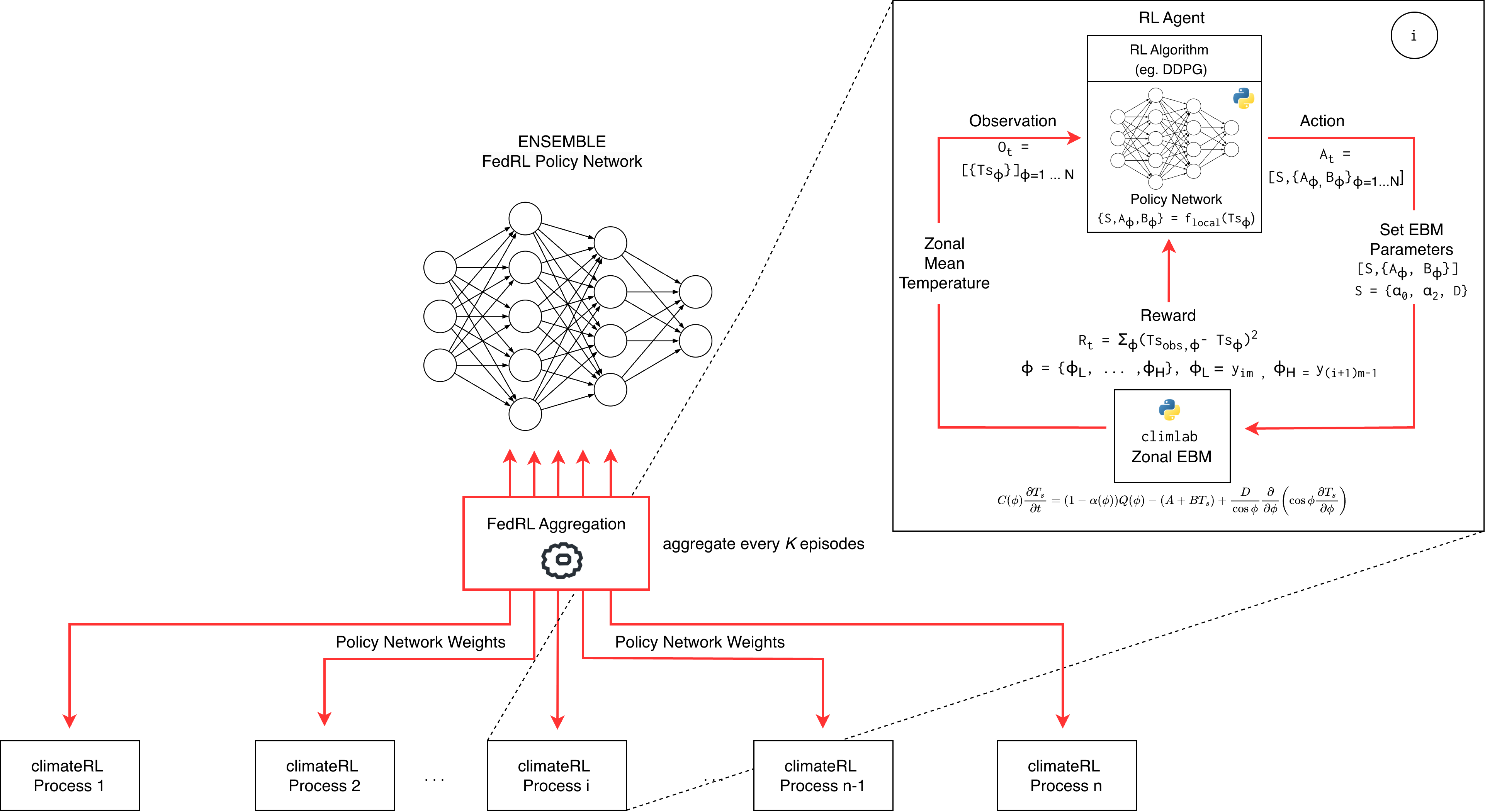}
\caption{\texttt{ebm-v2} multi-agent ensemble with FedRL
agents operate on assigned regions with local rewards while receiving the global profile as input. Periodic aggregation every \(K\) episodes synchronises policy weights across \(n\) agents.}
\label{fig:app-methods-ebm-v2-dataflow}
\end{figure}

\begin{sidewaysfigure}
\centering
\includegraphics[width=0.8\textwidth]{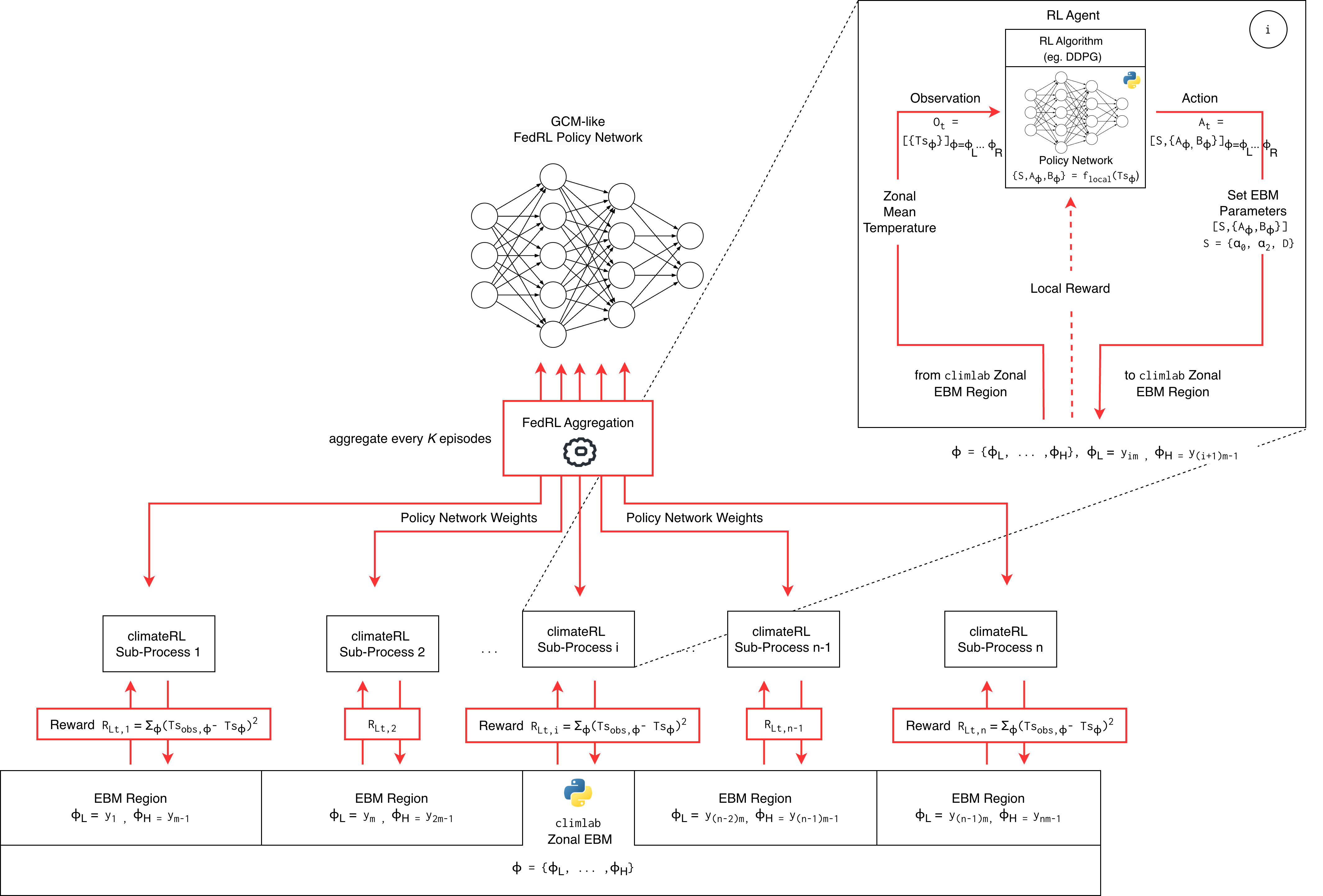}
\caption{Schematics for GCM-like \texttt{ebm-v3}. FedRL agents observe local temperature slices only (unlike \texttt{ebm-v2}) and optimise local rewards. Local rewards are computed over regions for each agent. Policy weights are aggregated every \(K\) episodes via Flower for FedRL and also contribute towards a global non-local policy. \( \{\alpha_0, \alpha_2, D\} \) are averaged over all $n$ sub-processes.}
\label{fig:app-methods-ebm-v3-dataflow}
\end{sidewaysfigure}

\clearpage

\subsection{Coupling Climate Simulations with AI}

Full-fledged climate models are typically implemented in highly optimised Fortran to support efficient numerical integration, whereas most modern AI and RL workflows are developed in Python, leveraging its extensive ML ecosystem~\cite{atkinson_ftorch_2025}. Connecting these two ecosystems without introducing performance bottlenecks or creating significant development overhead requires specialised infrastructure. In this work, {SmartSim}~\cite{partee_using_2022} and its in-memory communication layer {SmartRedis} are used to bridge this gap, enabling direct tensor exchange between Fortran/Python-wrapped solvers and Python-based RL agents. This approach removes the overhead associated with file-based input–output (I/O) and complex foreign-function interfaces, while remaining fully compatible with HPC-scale execution.

SmartSim provides an orchestration layer for hybrid HPC–ML workloads by running simulation workflows alongside in-memory Redis databases~\cite{sanfilippo_redis_2009}, while SmartRedis offers efficient tensor exchange through native Fortran and Python APIs. In the ML context, a tensor is a multi-dimensional numerical array (often with gradient-tracking support for backpropagation) that serves as the fundamental unit of computation. By enabling direct in-memory communication of such tensors, SmartSim and SmartRedis allow numerical solvers and learning algorithms to interact seamlessly, ensuring low-latency coupling and supporting scalable in-situ training of AI models within large-scale climate simulations.

Coupling between the climate model and RL agents is achieved by encoding state and numerical model parameter tensors as named keys within the Redis database. Each MPI rank in the climate model periodically writes its state variables into Redis using the SmartRedis Fortran client (or Python client, if the model is Python-based). These tensors are then retrieved by Python-based RL agents via the SmartRedis Python API, where policy updates are computed and the modified parameters are written back under predefined keys. During the subsequent timestep, the climate model reads these updated parameters and incorporates them into the next-step integration, completing a continuous, closed in-memory loop of data exchange between the numerical model and the RL agents.

Coordination of RL agents across different spatial subdomains is achieved using \texttt{flwr}~(Flower) \cite{beutel_flower_2022}, a widely adopted (mostly Python based) federated learning framework. In this configuration, each MPI rank hosts a local RL agent that interacts with its subdomain, updates its parameters, and periodically synchronises with a central aggregator through a strategy such as FedAvg~\cite{mcmahan_communication-efficient_2023}. After every \(K\) episodes, all agents transmit their policy weights for aggregation, receive the globally averaged weights (in case of FedAvg), and resume local fine-tuning for the next \(K-1\) episodes. SmartRedis manages the in-memory exchange of local climate states and model parameters between the distributed agents and the central \texttt{flwr} process, preserving data locality while minimising communication latency.

\subsection{RL Algorithm Summaries}
\label{app:methods-rl-algorithm-summaries}

\begin{sidewaystable}[!h]

\centering
\small
\caption[]{Four point summaries of different RL algorithms used}
\begin{tabular}{lp{11.5cm}}
\toprule
\textbf{Algorithm} & \textbf{Properties} \\ \midrule

Truncated Quantile Critics (TQC)~\cite{kuznetsov_controlling_2020} & 
1. Off-policy actor-critic algorithm that builds on SAC with a distributional critic using quantile regression. \newline
2. Models the full distribution of returns \( Z(s,a) \) as quantiles \( \{ \tau_i \} \), capturing uncertainty and reducing bias. \newline
3. Discards the top \( k \) quantiles before computing target values to avoid overestimation from outlier returns. \newline
4. Provides a smooth and robust training signal for distributional value estimation by minimising the quantile Huber loss:
\[
\mathcal{L}_\tau = \frac{1}{N} \sum_{i=1}^N \rho_\kappa(\tau_i - y)
\]
where \( \tau_i \) is the predicted \( i \)-th quantile, \( y \) is the target return, \( \rho_\kappa \) denotes the Huber loss with threshold \( \kappa \), and \( N \) is the number of quantile estimates used in training. \\ \midrule

Action Value Gradient (AVG)~\cite{vasan_deep_2024} &
1. On-policy actor–critic algorithm that combines reparameterised stochastic policies with log-probability regularisation to stabilise exploration. \newline
2. The actor minimises an entropy-augmented objective:
\[
\mathcal{L}_{\text{actor}} = \alpha \log \pi(a|s) - Q(s, a)
\quad \text{where} \quad a \sim \tanh(\mathcal{N}(\mu_\theta(s), \sigma_\theta(s)))
\]
This promotes both high-return and high-entropy policies by penalising confident low-reward actions. \newline
3. The critic is updated using TD learning with scaled errors:
\[
\delta = \frac{r + \gamma V(s') - Q(s, a)}{\hat{\sigma}_\delta}
\quad \text{where} \quad \hat{\sigma}_\delta \text{ is a running estimate of } \text{std}(\delta)
\]
Normalising TD errors ensures stable critic gradients even in the presence of large reward magnitudes or early-stage noise. \newline
4. Supports robust learning in continuous control by incorporating \texttt{tanh}-based action squashing, TD error scaling, and optional clipping of actions to respect environment bounds. \\

\bottomrule
\end{tabular}
\end{sidewaystable}

\clearpage

\begin{sidewaystable}[!h]
\ContinuedFloat
\centering
\small
\caption{Four point summaries (contd.)}
\label{tbl:app-methods-exp-setup-rl-algos}
\begin{tabular}{lp{11.5cm}}
\toprule
\textbf{Algorithm} & \textbf{Properties} \\ \midrule
REINFORCE~\cite{williams_simple_1992} & 
1. Off-policy Monte Carlo algorithm that performs updates only at the end of full trajectories \( (\tau) \), and does not include a critic. \newline
2. Uses a stochastic policy \( \pi_\theta(a|s) \) to generate actions and samples episodes from the environment. \newline
3. Optimises the expected return via the score function (log-derivative) estimator: \( \nabla_\theta J(\theta) = \mathbb{E}[\nabla_\theta \log \pi_\theta(\tau) R(\tau)] \). \newline
4. Suffers from high variance and slow convergence, often mitigated by introducing baselines (advantage estimates) or reward normalisation. \\ \midrule

Deterministic Policy Gradient (DPG)~\cite{silver_deterministic_2014} & 
1. On-policy actor-critic algorithm where the actor uses a deterministic policy \( \mu_\theta(s) \), suited to continuous action spaces. \newline
2. Critic is trained using TD learning to estimate the Q-function, while actor update uses gradient chain rule over Q. \newline
3. Gradient of the objective is \( \nabla_\theta J(\theta) = \mathbb{E}_{s \sim \rho^\mu} [ \nabla_\theta \mu_\theta(s) \nabla_a Q^\mu(s,a)|_{a = \mu_\theta(s)} ] \). \newline
4. Lacks target networks, making the learning process less stable and more sensitive to hyperparameters. \\ \midrule

Deep Deterministic Policy Gradient (DDPG)~\cite{lillicrap_continuous_2019} & 
1. Off-policy actor-critic algorithm that extends DPG using deep function approximators and additional stabilisation mechanisms. \newline
2. Introduces experience replay and target networks with soft target updates to decorrelate samples and improve training stability. \newline
3. Actor and critic networks are both updated similar to DPG. \newline
4. Overestimation bias and sensitivity to exploration noise often limit performance unless mitigated by design changes (e.g. TD3). \\ \midrule

Twin Delayed DDPG (TD3)~\cite{fujimoto_addressing_2018} & 
1. Off-policy actor-critic method designed to reduce the overestimation bias observed in DDPG. \newline
2. Uses a double critic architecture where the minimum of two Q-value estimates is used for critic updates. \newline
3. Actor is updated less frequently than the critics, and target networks are softly updated to reduce update variance. \newline
4. Injects temporally correlated Gaussian noise into the target actions to promote exploration in continuous action spaces. \\ 

\bottomrule
\end{tabular}
\end{sidewaystable}

\clearpage

\begin{sidewaystable}[!h]
\ContinuedFloat
\centering
\small
\caption[]{Four point summaries (contd.)}
\begin{tabular}{lp{11.5cm}}
\toprule
\textbf{Algorithm} & \textbf{Properties} \\ \midrule

Trust Region Policy Optimisation (TRPO)~\cite{schulman_trust_2015} & 
1. On-policy stochastic actor-critic algorithm with a theoretical guarantee of monotonic policy improvement. \newline
2. Formulates a constrained optimisation problem using a KL-divergence bound to control step sizes and prevent policy collapse. \newline
3. Uses advantage-weighted surrogate objectives with a linear approximation and solves the constraint via conjugate gradient. \newline
4. Highly stable but computationally expensive due to second-order updates and line search in large parameter spaces. \\ \midrule

Proximal Policy Optimisation (PPO)~\cite{schulman_proximal_2017} & 
1. On-policy stochastic actor-critic algorithm designed as a first-order alternative to TRPO with comparable stability. \newline
2. Uses a clipped surrogate objective to prevent excessively large policy updates: \newline
\[
L^{\text{CLIP}}(\theta) = \mathbb{E} \left[ \min \left( r_t(\theta) \hat{A}_t,\ \text{clip}(r_t(\theta), 1 - \epsilon, 1 + \epsilon) \hat{A}_t \right) \right]
\]
where \( r_t(\theta) = \frac{\pi_\theta(a_t|s_t)}{\pi_{\theta_{\text{old}}}(a_t|s_t)} \) and \( \hat{A}_t \) is an estimator of the advantage function. \newline
3. Leverages Generalised Advantage Estimation (GAE) to reduce variance of the policy gradient with tunable bias–variance trade-off. \newline
4. Efficient, robust and widely used in practice due to its balance of ease of implementation and empirical performance. \\ \midrule

Soft Actor-Critic (SAC)~\cite{haarnoja_soft_2018} & 
1. Off-policy actor-critic method for continuous control that augments the reward with an entropy term: \newline
\[
J(\pi) = \sum_t \mathbb{E}_{(s_t, a_t) \sim \rho_\pi} \left[ r(s_t, a_t) + \alpha \mathcal{H}(\pi(\cdot|s_t)) \right]
\] 
where \( \mathcal{H} \) is the entropy and \( \alpha \) is a temperature parameter. \newline
2. Encourages exploration by optimising for both reward and entropy, using a stochastic policy \( \pi(a|s) \). \newline
3. Maintains two Q-networks and uses the minimum value to reduce overestimation bias, with delayed soft target updates. \newline
4. Features automatic entropy tuning, high sample efficiency, and strong empirical performance on continuous control benchmarks. \\ 

\bottomrule
\end{tabular}
\end{sidewaystable}

\clearpage

\subsection{RL Algorithm Pseudocodes}
\label{app:methods-rl-algorithm-pseudocodes}


\subsubsection{REINFORCE}

\begin{algorithm}[!h]
\caption{REINFORCE}
\label{alg:app-pseudocode-REINFORCE}
\begin{algorithmic}[1]
\State \textbf{Input:} Gym environment, Number of episodes $M$, Steps per episode $N$, Learning rate $\alpha$, Discount factor $\gamma$

\State \textbf{Initialise:} Policy network parameters $\theta$, Actor network $\pi_{\theta}$, Learning rate $\alpha$

\State \textbf{Pre-Setup:} Configure seed and environment variables, prepare environment and logging \\

\For{$episode = 1$ \textbf{to} $M$}
    \State Initialise episode buffer $B \leftarrow \emptyset$
    \State Observe initial state $s_0$
    \For{$t = 0$ \textbf{to} $N-1$}
        \State Select action $a_t \sim \pi_{\theta}(s_t)$
        \State Execute action $a_t$ and observe reward $r_t$ and new state $s_{t+1}$
        \State Store transition $(s_t, a_t, r_t)$ in $B$
        \State $s_t \leftarrow s_{t+1}$
    \EndFor
    \State $G \leftarrow 0$, $\mathcal{L}(\theta) \leftarrow 0$
    \For{$t$ \textbf{in} $B$ \textbf{reversed}}
    \State $G \leftarrow r_t + \gamma G$
    \State $\mathcal{L}(\theta) \leftarrow \mathcal{L}(\theta) - \nabla_{\theta} G \log \pi_{\theta}(a_t|s_t)$
    \EndFor
    \State Update policy parameters $\theta$ using accumulated gradients:
    \State \[
    \theta \leftarrow \theta +  \eta \frac{1}{|B|} \mathcal{L}(\theta)
    \]
\EndFor
\end{algorithmic}
\end{algorithm}

\clearpage

\subsubsection{Deterministic Policy Gradient (DPG)}

\begin{algorithm}[!h]
\caption{Deterministic Policy Gradient (DPG)}
\label{alg:app-pseudocode-DPG}
\begin{algorithmic}[1]
\State \textbf{Input:} Gym environment, Total timesteps $T$, Discount factor $\gamma$, Learning rate for policy $\eta_{\pi}$, Learning rate for Q-network $\eta_Q$, Batch size $B$, Exploration noise $\sigma$

\State \textbf{Initialise:} Policy network parameters $\theta$, Q-function network parameters $\phi$

\State \textbf{Pre-Setup:} Configure seed and environment variables, prepare environment and logging \\

\For{$t = 1$ \textbf{to} $T$}
    \State Observe state $s$ and select action $a = \pi_{\theta}(s) + \epsilon$, where $\epsilon \sim \mathcal{N}(0, \sigma)$
    \State Execute action $a$ and observe next state $s'$, reward $r$, and termination signal $d$
    \State Calculate Q-values:
    \State \[
    y(r, s', d) = r + \gamma (1 - d) Q_{\phi}(s', \pi_{\theta}(s'))
    \]
    \State Update Q-function by minimising the loss:
    \State \[
    \phi \gets \phi - \eta_Q \nabla_{\phi} \left(Q_{\phi}(s, a) - y(r, s', d)\right)^2
    \]
    \State Update policy by one step of gradient ascent:
    \State \[
    \theta \gets \theta + \eta_{\pi} \nabla_{\theta} Q_{\phi}(s, \pi_{\theta}(s))
    \]
\EndFor

\end{algorithmic}
\end{algorithm}

\clearpage

\subsubsection{Deep Deterministic Policy Gradient (DDPG)}

\begin{algorithm}[!h]
\caption{Deep Deterministic Policy Gradient (DDPG)}
\label{alg:app-pseudocode-DDPG}
\begin{algorithmic}[1]

\State \textbf{Input:} Gym environment, Total timesteps $T$, Replay buffer size $N$, Discount factor $\gamma$, Target smoothing coefficient $\tau$, Batch size $B$, Learning rate $\eta$, Exploration noise $\sigma$

\State \textbf{Initialise:} Policy network parameters $\theta$, Q-function network parameters $\phi$, target network parameters $\theta_{\text{targ}}$, $\phi_{\text{targ}}$, empty replay buffer $\mathcal{D}$

\State \textbf{Pre-Setup:} Configure seed and environment variables, prepare environment and logging \\

\For{$t = 1$ \textbf{to} $T$}
    \State Observe state $s$ and select action $a = \pi_{\theta}(s)$ \State Add exploration noise $a \gets a + \epsilon$, where $\epsilon \sim \mathcal{N}(0, \sigma)$ if required
    \State Execute action $a$ and observe next state $s'$, reward $r$, and termination signal $d$
    \State Store transition $(s, a, r, s', d)$ in $\mathcal{D}$
    \If{$t \geq \text{learning\_starts}$}
        \State Sample a minibatch of $B$ transitions $(s, a, r, s', d)$ from $\mathcal{D}$
        \State Compute target for Q-function update:
        \State \[
        y(r, s', d) = r + \gamma (1 - d) Q_{\phi_{\text{targ}}}(s', \pi_{\theta_{\text{targ}}}(s'))
        \]
        \State Update Q-function by minimising the loss:
        \State \[
        \phi \gets \phi - \eta \nabla_{\phi} \frac{1}{|B|} \sum_{(s,a,r,s',d) \in B} \left( Q_{\phi}(s, a) - y(r, s', d) \right)^2
        \]
        \State Update policy by one step of gradient ascent:
        \State \[
        \theta \gets \theta + \eta \nabla_{\theta} \frac{1}{|B|} \sum_{s \in B} Q_{\phi}(s, \pi_{\theta}(s))
        \]
        \State Soft-update target networks:
        \State \[
        \theta_{\text{targ}} \gets \tau \theta + (1 - \tau) \theta_{\text{targ}},  \quad
        \phi_{\text{targ}} \gets \tau \phi + (1 - \tau) \phi_{\text{targ}}
        \]
    \EndIf
\EndFor

\end{algorithmic}
\end{algorithm}

\clearpage

\subsubsection{Twin Delayed DDPG (TD3)}

\begin{algorithm}[!h]
\caption{Twin Delayed DDPG (TD3}
\label{alg:app-pseudocode-TD3}
\begin{algorithmic}[1]
\State \textbf{Input:} Gym environment, Total timesteps $T$, Learning rate $\eta$, Replay buffer size $N$, Discount factor $\gamma$, Target smoothing coefficient $\tau$, Batch size $B$, Policy noise $\sigma_{\pi}$, Noise clip $\sigma_{\text{clip}}$, Exploration noise $\sigma_{\text{exploration}}$, Policy update frequency $f_{\pi}$

\State \textbf{Initialise:} Actor network $\theta$, Critic networks $\phi_1$, $\phi_2$, Target networks $\theta_{targ}$, $\phi_{targ, 1}$, $\phi_{targ, 2}$, Empty replay buffer $\mathcal{D}$

\State \textbf{Pre-Setup:} Configure seed and environment variables, prepare environment and logging \\

\For{$t = 1$ \textbf{to} $T$}
    \State Observe state $s$ and select action $a = \pi_{\theta}(s)$ \State Add exploration noise $a \gets a + \epsilon$, where $\epsilon \sim \mathcal{N}(0, \sigma_{\text{exploration}})$ if required
    \State Execute action $a$ and observe next state $s'$, reward $r$, and done signal $d$
    \State Store transition $(s, a, r, s', d)$ in $\mathcal{D}$
    \If{$t \geq \text{learning\_starts}$}
        \State Sample a minibatch of $B$ transitions $(s, a, r, s', d)$ from $\mathcal{D}$
        \State Compute target actions:
        \State \[
        a' \leftarrow \pi_{\theta_{targ}}(s') + \text{clip}(\mathcal{N}(0, \sigma_{\pi}), -\sigma_{\text{clip}}, \sigma_{\text{clip}})
        \]
        \State Compute target Q-values:
        \State \[
        y(r, s', d) \leftarrow r + \gamma (1 - d) \min_{i=1,2} Q_{\phi_{targ, i}}(s', a')
        \]
        \State Update critic networks by minimising the loss:
        \State \[
        \quad \phi_i \leftarrow \phi_i - \eta \nabla_{\phi_i} \frac{1}{|B|} \sum_{(s,a,r,s',d) \in B} \left( Q_{\phi_i}(s, a) - y(r, s', d) \right)^2, \text{for } i = 1, 2 
        \]
        \If{$t \mod f_{\pi} = 0$}
            \State Update actor network by policy gradient:
            \State \[
            \theta \leftarrow \theta + \eta \nabla_{\theta} \frac{1}{|B|} \sum_{s \in B} Q_{\phi_1}(s, \pi_\theta(s))
            \]
            \State Soft update target networks:
            \State \[
            \theta_{targ} \leftarrow \tau \theta + (1 - \tau) \theta_{targ}, \quad \phi_{targ, i} \leftarrow \tau \phi_i + (1 - \tau) \phi_{targ, i} \text{ for } i=1,2
            \]
        \EndIf
    \EndIf
\EndFor
\end{algorithmic}
\end{algorithm}

\clearpage

\subsubsection{Trust Region Policy Optimization (TRPO)}

\begin{algorithm}[!h]
\caption{Trust Region Policy Optimization (TRPO)}
\label{alg:app-pseudocode-TRPO}
\begin{algorithmic}[1]
\State \textbf{Input:} Gym environment, Total timesteps $T$, Mini-batch size $M$, Number of steps per episode $N$, Discount factor $\gamma$, GAE lambda $\lambda$, KL divergence limit $\delta$, Trust region update size $\beta$

\State \textbf{Initialise:} Policy parameters $\theta$, Value function parameters $\phi$

\State \textbf{Pre-Setup:} Configure seed and environment variables, prepare environment and logging \\

\For{$iteration = 1, 2, \dots, \frac{T}{N}$}
    \State Collect set of trajectories $\mathcal{D} = \{\tau_i\}$ by running policy $\pi_\theta$ in the environment
    \State Compute returns $\{{R_i}\}$ and advantage estimates $\{\hat{A_i}\}$ using GAE

    \For{$epoch = 1, 2, \dots, K$}
        \State Shuffle $D$ to create $M$ mini-batches
        \For{each mini-batch $t$}
            \State Update value function by minimising the MSE loss: 
            \State 
            \[
            L(\phi) = \frac{1}{2M}\sum_{t} \left(V_\phi(s_t) - \hat{R}_t\right)^2
            \]
            \State Compute the surrogate objective (policy loss):
            \[
            L^{\pi}(\theta) = \frac{1}{M} \sum_{t} \frac{\pi_\theta(a_t|s_t)}{\pi_{\theta_{old}}(a_t|s_t)} \hat{A}_t
            \]
            \State Compute policy gradient $\nabla_{\theta} L^{\pi}(\theta)$
            \State Apply conjugate gradient to estimate the natural policy gradient $\hat{g}$
            \[
            \hat{g} \approx (\nabla_{\theta}^2 KL(\pi_{\theta_{old}} \| \pi_\theta))^{-1} \nabla_{\theta} L^{\pi}(\theta)
            \]
            \State Compute step size $\alpha$ using line search:
            \[
            \alpha = \sqrt{\frac{2\delta}{\hat{g}^T H \hat{g}}}, \text{ where $H$ is the Hessian of $KL(\pi_{\theta_{old}} \| \pi_\theta)$}
            \]
            \State Update policy $\theta \leftarrow \theta + \alpha \hat{g}$ using an exponential increment strategy
        \EndFor
        \State \textbf{break} if $KL(\pi_{\theta_{old}} \| \pi_\theta) > \delta$
    \EndFor
\EndFor

\end{algorithmic}
\end{algorithm}

\clearpage

\subsubsection{Proximal Policy Optimization (PPO)}

\begin{algorithm}[!h]
\caption{Proximal Policy Optimization (PPO)}
\label{alg:app-pseudocode-PPO}
\begin{algorithmic}[1]
\State \textbf{Input:} Gym environment, Total timesteps $T$, Number of steps per episode $N$, Mini-batch size $M$, Update epochs $K$, Learning rate $\alpha$, Discount factor $\gamma$, GAE lambda $\lambda$, Clipping parameter $\epsilon$, VF coefficient $c_1$, Entropy coefficient $c_2$, KL divergence limit $\delta$

\State \textbf{Initialise:} Policy parameters $\theta$, Value function parameters $\phi$

\State \textbf{Pre-Setup:} Configure seed and environment variables, prepare environment and logging \\

\For{$iteration = 1, 2, \dots, \frac{T}{N}$}
    \State Collect set of trajectories $D = \{\tau_i\}$ by running policy $\pi_\theta$ in the environment
    \State Compute returns $\{{R_i}\}$ and advantage estimates $\{\hat{A_i}\}$ using GAE
    \For{$epoch = 1, 2, \dots, K$}
        \State Shuffle $D$ to create $M$ mini-batches
        \For{each mini-batch $t$}
            \State Compute ratio $r_t(\theta) = \frac{\pi_\theta(a_t|s_t)}{\pi_{\theta_{old}}(a_t|s_t)}$
            \State Compute clipped surrogate objective (policy loss):
            \[
            L^{CLIP}(\theta) = \hat{\mathbb{E}}_t \left[ \min(r_t(\theta) \hat{A}_t, \text{clip}(r_t(\theta), 1-\epsilon, 1+\epsilon) \hat{A}_t) \right]
            \]
            \State Compute value function loss:
            \[
            L^{VF}(\phi) = \left( V_\phi(s_t) - \hat{R}_t \right)^2
            \]
            \State Compute entropy: $S[\pi_\theta](s_t)$
            \State Compute total loss:
            \[
            L(\theta, \phi) = -L^{CLIP}(\theta) + c_1 L^{VF}(\phi) - c_2 S[\pi_\theta](s_t)
            \]
            \State Update $\theta$ and $\phi$ using stochastic gradient descent
        \EndFor
    \State \textbf{break} if $KL(\pi_{\theta_{old}} \| \pi_\theta) > \delta$
    \EndFor
\EndFor
\end{algorithmic}
\end{algorithm}

\clearpage

\subsubsection{Soft Actor-Critic (SAC)}

\begin{algorithm}[!h]
\caption{Soft Actor-Critic (SAC)}
\label{alg:app-pseudocode-SAC}
\begin{algorithmic}[1]

\State \textbf{Input:} Gym environment, Total timesteps $T$, Replay buffer size $N$, Discount factor $\gamma$, Target smoothing coefficient $\tau$, Batch size $B$, Learning rate for policy $\eta_{\pi}$, Learning rate for Q-network $\eta_Q$

\State \textbf{Initialise:} Policy network parameters $\theta$, Critic network parameters $\phi_1$, $\phi_2$, Target critic parameters $\phi_{\text{targ},1}$, $\phi_{\text{targ},2}$, Empty replay buffer $\mathcal{D}$, actor $\pi_\theta$, Entropy coefficient $\alpha$, Target entropy coefficient $\alpha_{\text{targ}}$

\State \textbf{Pre-Setup:} Configure seed and environment variables, prepare environment and logging \\

\For{$t = 1$ \textbf{to} $T$}
    \State Observe state $s$ and select action $a \sim \pi_\theta(s)$ with exploration strategy if required
    \State Execute action $a$ and observe next state $s'$, reward $r$, and termination signal $d$
    \State Store transition $(s, a, r, s', d)$ in $\mathcal{D}$
    \If{$t \geq \text{learning\_starts}$}
        \State Sample a minibatch of $B$ transitions $(s, a, r, s', d)$ from $\mathcal{D}$
        \State Compute targets for critic updates:
        \State \[
        y(r, s', d) = r + \gamma (1 - d) \left( \min_{i=1,2} Q_{\phi_{\text{targ},i}}(s', \tilde{a}') - \alpha \log \pi_\theta(\tilde{a}' | s') \right) \]
        \State where $\tilde{a}' \sim \pi_\theta(s')$
        \State Update Q-functions by one step of gradient descent:
        \State \[
        \phi_i \gets \phi_i - \eta_Q \nabla_{\phi_i} \frac{1}{|B|} \sum_{(s,a,r,s',d) \in B} \left( Q_{\phi_i}(s, a) - y(r, s', d) \right)^2 \text{ for } i=1,2
        \]
        \State Update policy by one step of gradient ascent:
        \State \[
        \theta \gets \theta + \eta_{\pi} \nabla_\theta \frac{1}{|B|} \sum_{s \in B} \left( \min_{i=1,2} Q_{\phi_i}(s, \pi_\theta(s)) - \alpha \log \pi_\theta(a|s) \right)
        \]
        \State Soft-update target networks:
        \State \[
        \phi_{\text{targ},i} \gets \tau \phi_i + (1 - \tau) \phi_{\text{targ},i} \text{ for } i = 1, 2
        \]
        \State Optionally adjust $\alpha$ based on entropy targets:
        \State \[
        \alpha \gets \alpha + \eta_Q \nabla_{\alpha} \frac{\alpha}{|B|}\sum_{s \in B} \left( \log \pi_\theta(a|s) + \alpha_{\text{targ}} \right)
        \]
    \EndIf
\EndFor

\end{algorithmic}
\end{algorithm}

\clearpage

\subsubsection{Truncated Quantile Critics (TQC)}

\enlargethispage{2\baselineskip}

\begin{algorithm}[!h]
\caption{Truncated Quantile Critics (TQC)}
\label{alg:app-pseudocode-TQC}
\begin{algorithmic}[1]
\State \textbf{Input:} Gym environment, Total timesteps $T$, Replay buffer size $N$, Discount factor $\gamma$, Smoothing coefficient $\tau$, Batch size $B$, Learning rate $\eta$, Number of quantiles $N_q$, Number of critics $N_c$, Drop quantiles $N_{\text{drop}}$, Entropy coefficient $\alpha$, Target entropy coefficient $\alpha_{\text{targ}}$

\State \textbf{Initialise:} Actor network $\theta$, Critic network parameters $\phi_1, \dots, \phi_{N_c}$, Target critic network parameters $\phi_{\text{targ},1}, \dots, \phi_{\text{targ},N_c}$, Replay buffer $\mathcal{D}$

\State \textbf{Pre-Setup:} Configure seed and environment variables, prepare environment and logging \\

\For{$t = 1$ \textbf{to} $T$}
    \State Select action $a \sim \pi_\theta(s)$ based on current policy and exploration strategy
    \State Execute action $a$ and observe next state $s'$, reward $r$, and done signal $d$
    \State Store transition tuple $(s, a, r, s', d)$ in $\mathcal{D}$
    \If{$t \geq \text{learning\_starts}$} 
        \For{$i = 1$ \textbf{to} $N_c$}
            \State Sample a minibatch of $B$ transitions $(s, a, r, s', d)$ from $\mathcal{D}$
            \State Compute target quantile values for critic $\phi_{target, i}$:
            \State \[
        y(r, s', d) = r + \gamma (1 - d) \left( Q_{\phi_{\text{targ},i}}(s', \tilde{a}', N_{\text{drop}}) - \alpha \log \pi_\theta(\tilde{a}' | s') \right) \]
        \State where $\tilde{a}' \sim \pi_\theta(s')$        
            \State Update critic $\phi_i$ by minimising the quantile Huber loss:
            \[
            \text{L}^{\phi_i} = \frac{1}{N_q} \sum_{k=1}^{N_q} \text{HuberLoss}(Q_{\phi_i}(s_j, a_j, \tau_k) - y_j) \]
            \State where $\tau_k$ are the quantile fractions
        \EndFor
        \State Update policy by one step of gradient ascent:
        \State \[
        \theta \gets \theta + \eta \nabla_\theta \frac{1}{|B|} \sum_{s \in B} \left( - \alpha \log \pi_\theta(a|s) + \frac{1}{N_c} \sum_{i=1}^{N_c} Q_{\phi_i}(s, \pi_\theta(s)) \right)
        \]
        \State Soft-update target networks:
        \State \[
        \phi_{\text{targ},i} \gets \tau \phi_i + (1 - \tau) \phi_{\text{targ},i} \text{ for } i = 1, 2, ..., N_c
        \]
        \State Optionally adjust $\alpha$ based on entropy targets:
        \State \[
        \alpha \gets \alpha + \eta \nabla_{\alpha} \frac{\alpha}{|B|}\sum_{s \in B} \left( \log \pi_\theta(a|s) + \alpha_{\text{targ}} \right)
        \]
    \EndIf
\EndFor

\end{algorithmic}
\end{algorithm}

\clearpage

\subsubsection{Action Value Gradient (AVG)}

\begin{algorithm}[!h]
\caption{Action Value Gradient (AVG)}
\label{alg:app-pseudocode-AVG}
\begin{algorithmic}[1]

\State \textbf{Input:} Gym environment, Total timesteps $T$, Discount factor $\gamma$, Entropy coefficient $\alpha$

\State \textbf{Initialise:} Actor network parameters $\theta$, Critic network parameters $\phi$, actor $\pi_\theta(a|s)$, critic $Q_\phi(s, a)$, TD error normaliser $\hat{\sigma}_\delta$, Return $G$

\State \textbf{Pre-Setup:} Configure seed and environment variables, prepare environment and logging \\

\For{$t = 1$ \textbf{to} $T$}
    \State Select action $a \sim \pi_\theta(s)$ using squashed \texttt{tanh}
    \State Execute action $a$ and observe next state $s'$, reward $r$, and done signal $d$
    \State Compute entropy-augmented reward: $r_{\text{ent}} = r - \alpha \log \pi_\theta(a | s)$
    \State Accumulate return: $G \gets G + r_{\text{ent}}$
    \If {episode done}
        \State Update TD error scaler $\hat{\sigma}_\delta$ using $G$ and $r_{\text{ent}}$
        \State $G \gets 0$        
    \Else        
        \State Update TD error scaler $\hat{\sigma}_\delta$ using $\gamma$ and $r_{\text{ent}}$
    \EndIf
    \State Compute target value: $V(s') = Q_\phi(s', \pi_\theta(s')) - \alpha \log \pi_\theta(a'|s')$ where ${a}' \sim \pi_\theta(s')$        
    \State Compute the critic loss as the square of the scaled TD error: 
    \[
    \mathcal{L}^{\phi} = 
    \left[ \frac{r + \gamma (1 - d) V(s') - Q_\phi(s, a)}{\hat{\sigma}_\delta} \right]^2
    \]
    \State Compute the actor loss:
    \[
    \mathcal{L}^{\theta} = \alpha \log \pi_\theta(a|s) - Q_\phi(s, a)
    \]
    
    \State Update both $\theta$ and $\phi$ through one step of gradient descent
\EndFor

\end{algorithmic}
\end{algorithm}


\clearpage

\subsection{Experimental Setup}

\subsubsection{RL Algorithms}

Nine RL algorithms (shown in Figure~\ref{fig:exp-setup-rl-algo-progression}, summaries and pseudocodes in Appendices~\ref{app:methods-rl-algorithm-summaries} and~\ref{app:methods-rl-algorithm-pseudocodes}) are evaluated across the single-agent environments (\texttt{scbc-v0/1/2, rce-v0, rce-17-v0/v1}, and \texttt{ebm-v0/1}). These include five on-policy methods: REINFORCE~\cite{williams_simple_1992}, TRPO~\cite{schulman_trust_2015}, PPO~\cite{schulman_proximal_2017}, DPG~\cite{silver_deterministic_2014}, and AVG~\cite{vasan_deep_2024}, which update policies using freshly sampled trajectories, and five off-policy methods:  DDPG~\cite{lillicrap_continuous_2019}, TD3~\cite{fujimoto_addressing_2018}, SAC~\cite{haarnoja_soft_2018}, and TQC~\cite{kuznetsov_controlling_2020}, which utilise replay buffers and target networks to improve sample efficiency. Except for REINFORCE, a pure policy-gradient algorithm, all methods follow an actor–critic structure. Table~\ref{tbl:app-methods-exp-setup-rl-algos} summarises their architectures and main features. For the multi-agent environments (\texttt{ebm-v2/3}), only the top three performers from the single-agent experiments are carried forward for evaluation.

\begin{figure}[!h]
    \floatbox[{\capbeside\thisfloatsetup{capbesideposition={right,center},capbesidewidth=0.65\textwidth}}]{figure}[\FBwidth]
    {\caption{Developmental progression of RL algorithms, evaluated in this study. The single agent climateRL experiments span classical policy-gradient methods such as REINFORCE, on-policy algorithms --- DPG, TRPO, PPO, AVG, and advanced off-policy actor–critic approaches --- DDPG, TD3, SAC, TQC. Adapted from \protect\url{https://master-dac.isir.upmc.fr/rl/12_sac.pdf}.}
    \label{fig:exp-setup-rl-algo-progression}}
    {\includegraphics[width=0.35\textwidth]{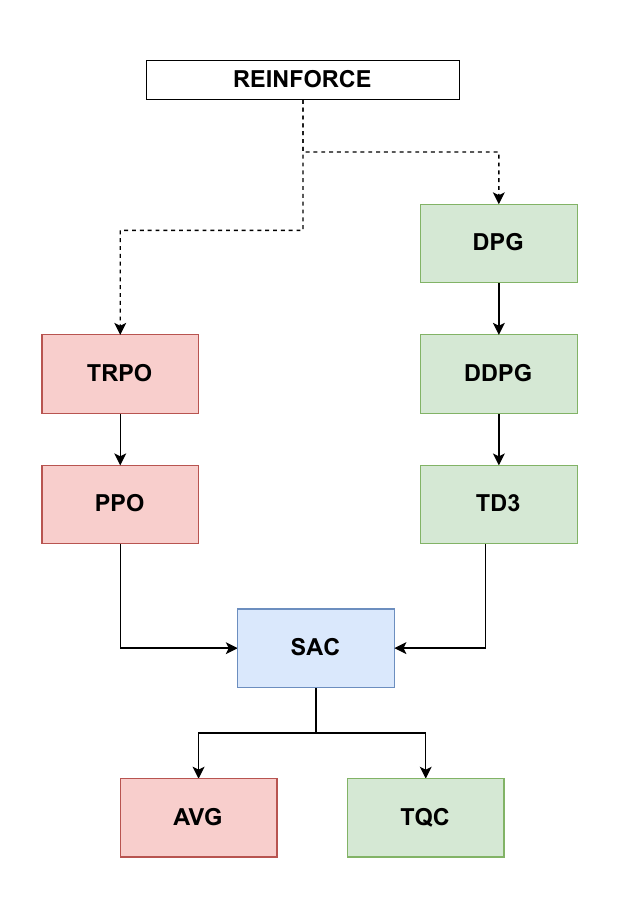}}
\end{figure}

\subsubsection{Hyperparameter Tuning}

Hyperparameter optimisation is performed using {Ray}~\cite{moritz_ray_2018} (a distributed computing framework) on a SLURM-managed cluster, with one head node and three workers (each equipped with 8× AMD EPYC 74F3 cores). For algorithms such as TQC, which require additional hardware acceleration due to multiple critic networks, GPUs are partitioned into four logical devices using Ray to enable efficient parallel trials. Advanced sampling during the search is handled using {Optuna}~\cite{akiba_optuna_2019} (a hyperparameter optimisation framework) within Ray. For each RL algorithm, 32 parallel trials are launched on JASMIN~\cite{lawrence_jasmin_2012}. For multi-agent environments (\texttt{ebm-v2/3}), the best hyperparameters identified for the single-agent baseline (\texttt{ebm-v1}) are reused.

\subsubsection{Evaluation Strategies}

\textbf{Single-agent RL}
\label{sec:app-methods-exp-setup-eval-single-agent-RL}

\noindent Single-agent environments (\texttt{scbc-v0/1/2, rce-v0, rce-17-v0/v1}, and \texttt{ebm-v0/1}) are evaluated under two architectural setups. In the first setup, \texttt{optim-L}, actor–critic network sizes are tuned individually to capture environment-specific complexity. In the second, \texttt{homo-64L}, all networks are fixed to 64 hidden units per layer to enforce uniformity. Each setup is run under two tuning regimes: a constrained regime that mirrors a scenario where full convergence run may not be practical due to compute limitations, and an extended regime with longer tuning horizons (Table~\ref{tbl:app-methods-exp-setup-optuna-timesteps}). Together, these design choices yield 32 distinct experiment configurations, summarised in Appendix~\ref{tbl:app-methods-exp-setup-eval-experiments}.  

\begin{table}[!h]
\centering
\caption{Optimisation timesteps for hyperparameter tuning in each environment. Tuning is performed on seed 1 across specified timesteps/episodes. Episode lengths are in timesteps.}
\label{tbl:app-methods-exp-setup-optuna-timesteps}
\ttfamily
\begin{tabular}{llll}
\toprule
\textbf{\textrm{Environment ID}} & \textbf{\textrm{Tuning ID}} & \textbf{\textrm{Tuning Timesteps}} & \textbf{\textrm{Episode Length}} \\
\cmidrule(lr){1-4}
\multirow{3}{*}{scbc-v0/1/2} 
& \makecell[l]{homo-64L\\optim-L} & 2000 & \multirow{3}{*}{200} \\
\cmidrule(lr){2-3}
& \makecell[l]{homo-64L-60k\\optim-L-60k} & 60000 - TOTAL \\
\cmidrule(lr){1-4}
\multirow{3}{*}{\shortstack[l]{rce-v0\\rce17-v0/1}}
& \makecell[l]{homo-64L\\optim-L} & 5000 & \multirow{3}{*}{500} \\
\cmidrule(lr){2-3}
& \makecell[l]{homo-64L-10k\\optim-L-10k} & 10000 - TOTAL \\
\cmidrule(lr){1-4}
\multirow{3}{*}{ebm-v0/1} 
& \makecell[l]{homo-64L\\optim-L} & 10000 & \multirow{3}{*}{200} \\
\cmidrule(lr){2-3}
& \makecell[l]{homo-64L-20k\\optim-L-20k} & 20000 - TOTAL \\
\bottomrule
\end{tabular}
\end{table}

Since RL lacks a universal train/test split procedure unlike supervised learning~\cite{patterson_empirical_2024}, evaluation is based on three complementary metrics: 

\begin{enumerate}[label=\alph*.]
    \item \textbf{Sample efficiency}: Measured by the number of numerical models steps \(N_{\text{to\_threshold}}\) required to cross an empirically defined return threshold (Table~\ref{tbl:app-methods-exp-setup-eval-thresholds}). {These thresholds are chosen empirically from the episodic return curves, specifically near the onset of the plateau or ankle region, where the agent reaches a consistent "good enough" level of control performance. These therefore serve as heuristic convergence markers for quantifying how quickly an algorithm attains a practically useful policy.}

    \item \textbf{Policy stability}: Defined as the variance \(\sigma^2_{\text{after\_threshold}}\) of episodic returns once the threshold has been crossed, indicating the consistency and reliability of performance over extended training.  

    \item \textbf{Asymptotic performance}: Measured as the difference \(\Delta_{\text{from\_10k/20k/60k}}\) between the final return at the end of the experiment and the return threshold, reflecting the long-horizon convergence behaviour of the algorithm.  
\end{enumerate}

\begin{table}[!h]
\centering
\caption{Empirically determined episodic return thresholds for each RL environment. Error is computed using the formula \( \sqrt{\frac{\text{{$\lvert\mathrm{Threshold} \rvert$}}}{\# \text{Timesteps per episode}}} \). Sparse rewards in \texttt{SimpleClimateBiasCorrection} are implemented in the Gym environment by assigning a constant upper-bound normalised temperature error of 1 at every timestep, except every 5th step (when the reward is activated), yielding a minimum episodic return of \( 200 - \frac{200}{5} = 160 \). Errors shown in brackets for \texttt{RadiativeConvectiveModelEnv} and \texttt{EnergyBalanceModelEnv} are averaged across 17 pressure levels and 96 latitudes respectively.}

\label{tbl:app-methods-exp-setup-eval-thresholds}
\ttfamily
\begin{tabular}{lll}
\toprule
\textbf{\textrm{Environment}} & \textbf{\textrm{Threshold}} & \textbf{\textrm{Error (in K) per Episodic Step}} \\ \midrule
{SimpleClimateBiasCorrection-v0} & -0.25 & $\pm$ 0.035 \\ 
{SimpleClimateBiasCorrection-v1} & -2.718 & $\pm$ 0.116 \\ 
{SimpleClimateBiasCorrection-v2} & $-1 \times (160 + 2.718)$ & $\pm$ 0.116 \\ 
\midrule
{RadiativeConvectiveModel-v0} & -43900 & $\pm$ 9.370 (0.551) \\
{RadiativeConvectiveModel17-v0} & -43700 & $\pm$ 9.348 (0.550) \\
{RadiativeConvectiveModel17-v1} & -43650 & $\pm$ 9.343 (0.549) \\
\midrule
{EnergyBalanceModel-v0} & -10000 & $\pm$ 7.071 (0.074) \\
{EnergyBalanceModel-v1} & -30000 & $\pm$ 12.247 (0.127) \\
\bottomrule
\end{tabular}
\end{table}

This multi-metric evaluation, averaged across 10 random seeds, balances sample efficiency (metrics (a) and (c)) with robustness (metric (b)), providing a holistic view of learning behaviour. Benchmark return thresholds, listed in Table~\ref{tbl:app-methods-exp-setup-eval-thresholds}, are empirically derived from observed learning curves and represent episodic return values that indicate meaningful learning, serving as a consistent reference point for analysis. 

\textbf{1. Rank-based Composite Scoring:}~To compare algorithms across metrics, we adopt a rank-based composite evaluation. Each algorithm is ranked in ascending order for metrics (a)–(c), and an additional penalty is imposed if policy variance exceeds environment-specific thresholds (e.g., \(3 \times 10^{-3}\) for SCBC and \(3 \times 10^{5}\) for RCE and EBM environments). Final scores are obtained by summing ranks across all metrics and sorting in ascending order. This procedure, commonly used in benchmarking studies~\cite{ikhtiarudin_benchrl-qas_2025}, enables fair comparison across heterogeneous regimes while maintaining interpretability.

\textbf{2. Post-Ranking Diagnostics:}~To better understand top-performing algorithms, additional diagnostics are run in inference mode with fixed policy weights. In the RCE setup, mean absolute error (MAE) in temperature is computed against reanalysis data at 100, 200, and 1000~hPa, capturing both upper-tropospheric and near-surface behaviour. For the EBM, the 96 latitudes are aggregated into six 30\textdegree{} zones, where area-weighted RMSE (areaWRMSE), and model bias are evaluated against the reference climatology for both RL-assisted and vanilla \texttt{climlab} run which uses a static fitted solution obtained by calibrating the EBM parameters against the target climatology via linear regression. These diagnostics add to the physical meaning that extend beyond episodic return, quantifying fidelity to observations and highlighting model realism.

\textbf{3. Temporal Evolution:}~In addition to reward- and error-based evaluations, the SCBC, RCE, and EBM environments are examined by visualising the temporal evolution of the state dependent values (i.e., the agent’s actions). Tracking these trajectories shows how policies learn with the evolving physics-based state, providing insight into agent stability and convergence. Such visual diagnostics reveal whether policies converge during learning, oscillate, or follow trends consistent with physical processes, helping to distinguish meaningful learning from artefacts of exploration or instability.

\textbf{Multi-agent RL}
\label{sec:app-methods-exp-setup-eval-multi-agent-RL}

\noindent To assess scalability and coordination in distributed settings, the top three algorithms from the single-agent EBM experiments are re-run in the multi-agent FedRL environments. For consistency, \texttt{ebm-v2} and \texttt{ebm-v3} reuse the best hyperparameters obtained for the single-agent baseline (\texttt{ebm-v1}). Three aggregation regimes are considered: \texttt{fed05}, where parameters are synchronised every five episodes, \texttt{fed10}, where synchronisation occurs every ten episodes, and \texttt{nofed}, where agents train independently for the full 20k timesteps. Each regime is tested under two spatial decompositions: a 2-zone split (northern vs.~southern hemisphere) and a 6-zone split (30\textdegree{} latitude bands spanning polar, mid-latitude, and tropical regions). Together, these choices yield 12 experimental configurations, detailed in Appendix~\ref{tbl:app-methods-exp-setup-eval-fedrl-experiments}.  

Training curves from these multi-agent setups are benchmarked against the single-agent \texttt{ebm-v1} baseline to evaluate convergence dynamics and assess whether spatial decomposition accelerates learning. The areaWRMSE post-ranking diagnostic introduced earlier is applied separately to both \texttt{ebm-v2} and \texttt{ebm-v3} for locally fine-tuned and globally averaged policies. These analyses provide a consistent measure of performance across federated configurations and highlight potential benefits of decentralised, regime-aware learning.

\subsection{climateRL Experiment Codes}

\begin{table}[!h]
\centering
\caption{Experiment codes for each single-agent RL environment (each run across 10 seeds)}
\label{tbl:app-methods-exp-setup-eval-experiments}
\ttfamily
\begin{tabular}{lll}
\toprule
\textbf{\textrm{Group}} & \textbf{\textrm{Environment}} & \textbf{\textrm{Experiment ID}} \\ 
\cmidrule(lr){1-3}
\multirow{12}{*}{\shortstack[l]{\textrm{Simple Climate}\\\textrm{Bias Correction}}}
& \multirow{4}{*}{SimpleClimateBiasCorrection-v0} & scbc-v0-optim-L \\
& & scbc-v0-optim-L-60k \\
& & scbc-v0-homo-64L \\
& & scbc-v0-homo-64L-60k \\
\cmidrule(lr){2-3}
& \multirow{4}{*}{SimpleClimateBiasCorrection-v1} & scbc-v1-optim-L \\
& & scbc-v1-optim-L-60k \\
& & scbc-v1-homo-64L \\
& & scbc-v1-homo-64L-60k \\
\cmidrule(lr){2-3}
& \multirow{4}{*}{SimpleClimateBiasCorrection-v2} & scbc-v2-optim-L \\
& & scbc-v2-optim-L-60k \\
& & scbc-v2-homo-64L \\
& & scbc-v2-homo-64L-60k \\
\cmidrule(lr){1-3}
\multirow{12}{*}{\shortstack[l]{\textrm{Radiative Convective}\\\textrm{Equilibrium (RCE)}}}
& \multirow{4}{*}{RadiativeConvectiveModel-v0} & rce-v0-optim-L \\
& & rce-v0-optim-L-10k \\
& & rce-v0-homo-64L \\
& & rce-v0-homo-64L-10k \\
\cmidrule(lr){2-3}
& \multirow{4}{*}{RadiativeConvectiveModel17-v0} & rce17-v0-optim-L \\
& & rce17-v0-optim-L-10k \\
& & rce17-v0-homo-64L \\
& & rce17-v0-homo-64L-10k \\
\cmidrule(lr){2-3}
& \multirow{4}{*}{RadiativeConvectiveModel17-v1} & rce17-v1-optim-L \\
& & rce17-v1-optim-L-10k \\
& & rce17-v1-homo-64L \\
& & rce17-v1-homo-64L-10k \\
\cmidrule(lr){1-3}
\multirow{8}{*}{\shortstack[l]{\textrm{Energy Balance}\\\textrm{Model (EBM)}}}
& \multirow{4}{*}{EnergyBalanceModel-v0} & ebm-v0-optim-L \\
& & ebm-v0-optim-L-20k \\
& & ebm-v0-homo-64L \\
& & ebm-v0-homo-64L-20k \\
\cmidrule(lr){2-3}
& \multirow{4}{*}{EnergyBalanceModel-v1} & ebm-v1-optim-L \\
& & ebm-v1-optim-L-20k \\
& & ebm-v1-homo-64L \\
& & ebm-v1-homo-64L-20k \\
\bottomrule
\end{tabular}
\end{table}

\begin{table}[!h]
\centering
\caption{Experiment codes for each FedRL environment (each run across 10 seeds)}
\label{tbl:app-methods-exp-setup-eval-fedrl-experiments}
\ttfamily
\begin{tabular}{lll}
\toprule
\textbf{\textrm{Group}} & \textbf{\textrm{FedRL Environment}} & \textbf{\textrm{Experiment ID}} \\ 
\cmidrule(lr){1-3}
\multirow{12}{*}{\shortstack[l]{\textrm{Energy Balance}\\\textrm{Model (EBM)}}}
& \multirow{6}{*}{EnergyBalanceModel-v2} & ebm-v2-optim-L-20k-a6-fed05 \\
& & ebm-v2-optim-L-20k-a6-fed10 \\
& & ebm-v2-optim-L-20k-a6-nofed \\
\cmidrule(lr){3-3}
& & ebm-v2-optim-L-20k-a2-fed10 \\
& & ebm-v2-optim-L-20k-a2-nofed \\
& & ebm-v2-optim-L-20k-a2-fed10 \\
\cmidrule(lr){2-3}
& \multirow{6}{*}{EnergyBalanceModel-v3} & ebm-v3-optim-L-20k-a6-fed05 \\
& & ebm-v3-optim-L-20k-a6-fed10 \\
& & ebm-v3-optim-L-20k-a6-nofed \\
\cmidrule(lr){3-3}
& & ebm-v3-optim-L-20k-a2-fed05 \\
& & ebm-v3-optim-L-20k-a2-fed10 \\
& & ebm-v3-optim-L-20k-a2-nofed \\
\bottomrule
\end{tabular}
\end{table}

\clearpage

\normalsize

\label{app:additional_results}
\section{Additional Results}

\subsection{Single-agent RL}

\subsubsection{SCBC Environment}
\label{app:results-scbc-training}

\begin{figure}[!h]
    \centering
    \makebox[\textwidth][c]{\includegraphics[scale=0.55]{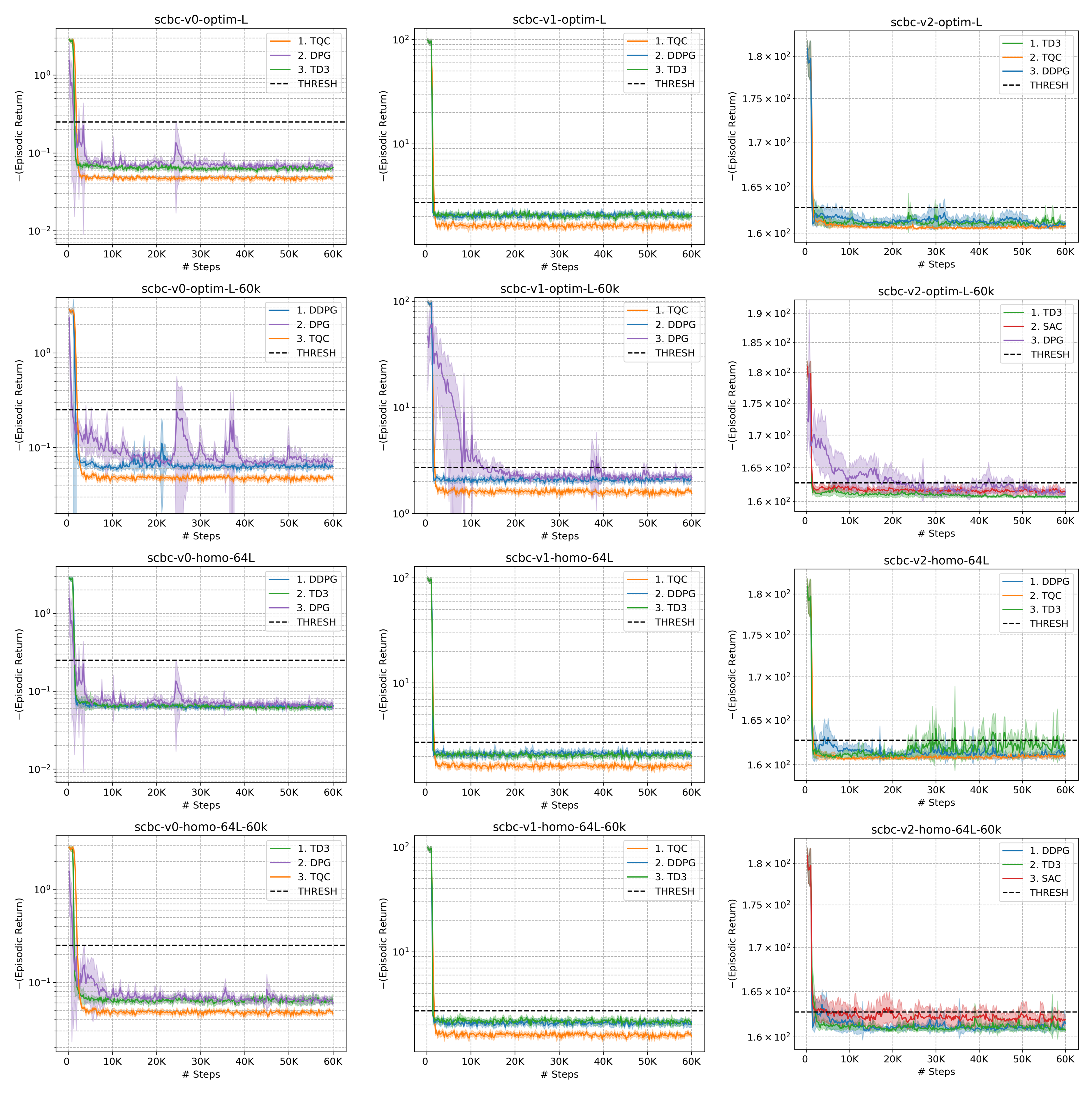}}
    \caption{Training curves across 10 seeds for SCBC environments (v0, v1, v2) across all twelve tuning configurations (\texttt{scbc-v0/1/2-optim-L-60k} reproduced for easy reference). Episodic returns are plotted on a log scale. Shaded regions denote $\pm$1.96 standard deviation (95\% confidence intervals). Threshold values are mentioned in Table~\ref{tbl:app-methods-exp-setup-eval-thresholds}.}
    \label{fig:app-results-scbc-training-curves}
\end{figure}

\clearpage

\subsubsection{RCE Environment}
\label{app:results-rce-training}

\begin{figure}[!h]
    \centering
    \makebox[\textwidth][c]{\includegraphics[scale=0.55]{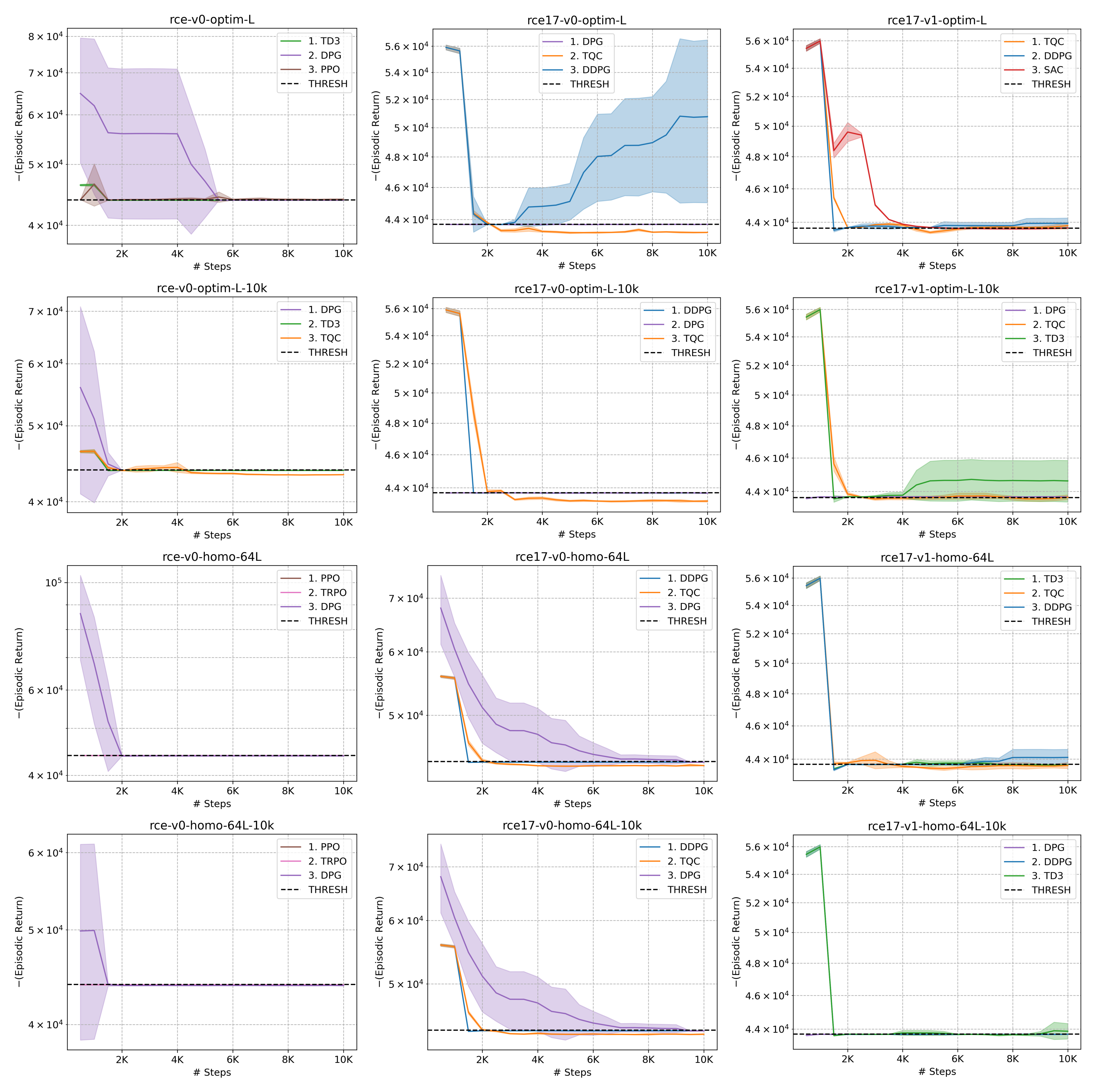}}    
    \caption{Training curves for RCE environments (\texttt{rce-v0}, \texttt{rce17-v0}, \texttt{rce17-v1}) across all twelve tuning configurations (\texttt{rce-v0/rce17-v0/1-optim-L-10k} reproduced for easy reference). Episodic returns are plotted on a log scale. Shaded regions denote $\pm$1.96 standard deviation (95\% confidence intervals). Threshold values are mentioned in Table~\ref{tbl:app-methods-exp-setup-eval-thresholds}.}
    \label{fig:app-results-rce-training-curves}
\end{figure}

\clearpage

\subsubsection{EBM Environment}
\label{app:results-ebm-v01-training}

\begin{figure}[!h]
    \centering
    \makebox[\textwidth][c]{\includegraphics[scale=1]{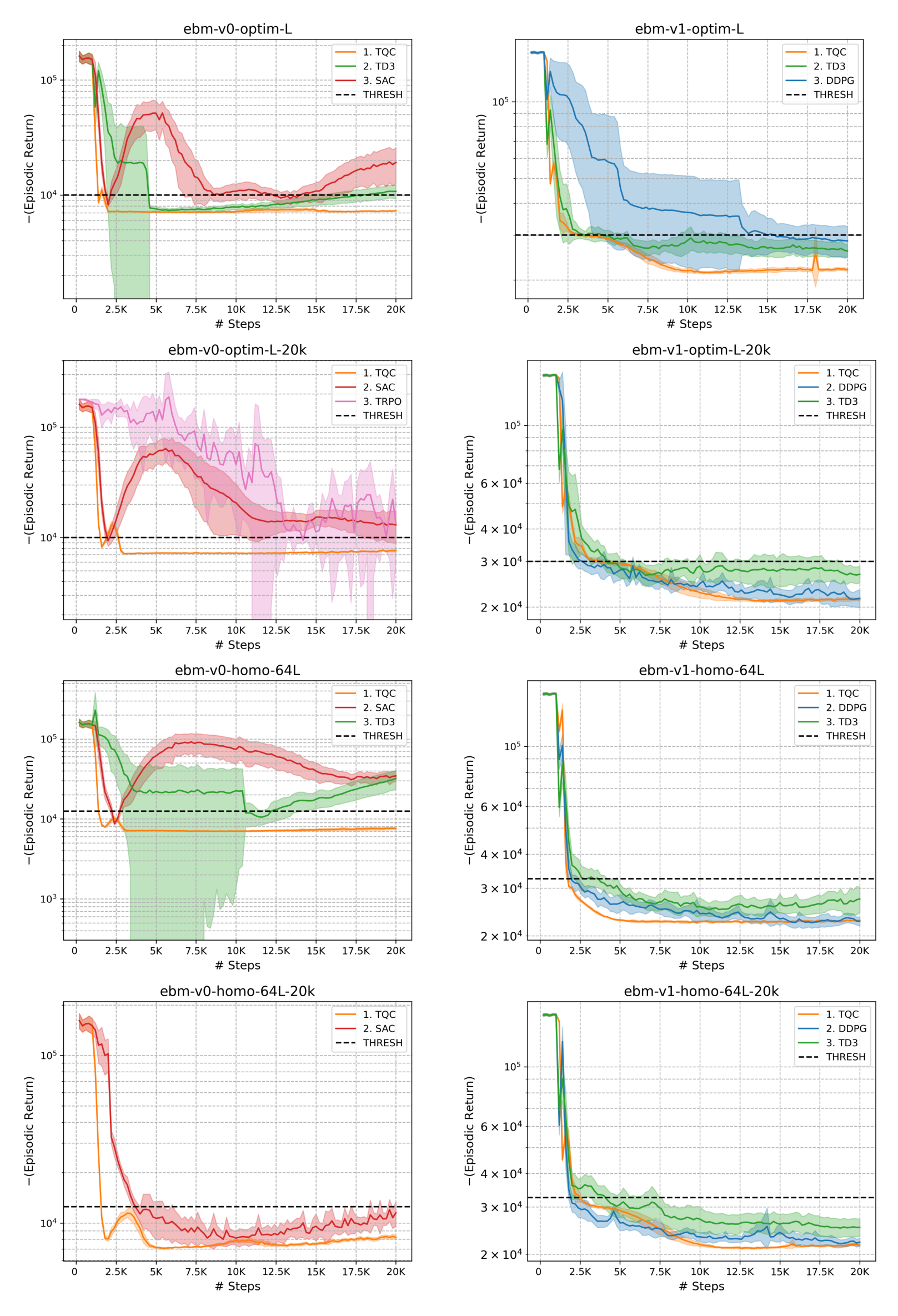}}    
    \caption{Episodic return curves (log-scaled) with 95\% spreads over 10 seeds for the top-3 RL algorithms across eight single-agent EBM configurations (\texttt{ebm-v0/1-optim-L-20k} reproduced for easy reference). Threshold values are mentioned in Table~\ref{tbl:app-methods-exp-setup-eval-thresholds}.}
    \label{fig:app-results-ebm-v01-training-curves}
\end{figure}

\subsection{Multi-agent RL}

\subsubsection{FedRL Skill Metrics}
\label{app:results-ebm-fedrl-skill-metrics}

\vspace{-0.75cm}

\begin{sidewaystable}[H]
\centering
\small
\caption{Zonal‑band errors for \texttt{ebm-v2-optim-L-20k-a2}. Each subtable reports mean ± std and relative gain \% versus \texttt{ebm-v1} for three regimes \texttt{fed05}, \texttt{fed10} and \texttt{nofed}, along with a comparison against the static baseline \texttt{climlab}.}
\setlength{\tabcolsep}{4pt}
\captionsetup[sub]{font=small}
\begin{minipage}[t]{\textwidth}
\centering
\subcaption{DDPG}
\begin{tabular}{l c cc cc cc c}
\toprule
 & \multirow{2}{*}{\texttt{climlab}} & \multicolumn{2}{c}{\texttt{fed05}} & \multicolumn{2}{c}{\texttt{fed10}} & \multicolumn{2}{c}{\texttt{nofed}} & \multirow{2}{*}{\texttt{ebm-v1}} \\
 &  & Mean ± Std & Gain \% & Mean ± Std & Gain \% & Mean ± Std & Gain \% &  \\
\midrule
90°S–60°S & 11.453 & 5.11 ± 0.533 & 37.550 & 6.14 ± 0.964 & 25.040 & 16.17 ± 14.718 & -97.490 & 8.19 ± 3.433 \\
60°S–30°S & 7.768  & 3.08 ± 1.155 & 51.030 & 3.39 ± 0.775 & 46.080 & 10.30 ± 10.647 & -63.620 & 6.30 ± 4.019 \\
30°S–0°   & 2.730  & 3.46 ± 1.984 & 48.890 & 3.92 ± 1.642 & 42.080 & 12.24 ± 13.889 & -81.120 & 6.76 ± 3.013 \\
0°–30°N   & 3.746  & 2.80 ± 2.384 & 39.050 & 1.89 ± 1.325 & 58.760 & 2.23 ± 1.246  & 51.460  & 4.59 ± 2.068 \\
30°N–60°N & 6.398  & 2.35 ± 0.866 & 11.350 & 2.26 ± 1.307 & 14.730 & 2.38 ± 1.253  & 10.360  & 2.65 ± 0.997 \\
60°N–90°N & 5.566  & 1.60 ± 0.682 & 54.090 & 1.95 ± 0.995 & 44.060 & 2.49 ± 0.758  & 28.400  & 3.48 ± 1.790 \\
\bottomrule
\end{tabular}
\vspace{0.1cm}
\end{minipage}\hfill
\begin{minipage}[t]{\textwidth}
\centering
\subcaption{TD3}
\begin{tabular}{l c cc cc cc c}
\toprule
 & \multirow{2}{*}{\texttt{climlab}} & \multicolumn{2}{c}{\texttt{fed05}} & \multicolumn{2}{c}{\texttt{fed10}} & \multicolumn{2}{c}{\texttt{nofed}} & \multirow{2}{*}{\texttt{ebm-v1}} \\
 &  & Mean ± Std & Gain \% & Mean ± Std & Gain \% & Mean ± Std & Gain \% &  \\
\midrule
90°S–60°S & 11.453 & 8.00 ± 1.380 & 17.850  & 6.87 ± 0.825 & 29.520  & 7.72 ± 1.371 & 20.750  & 9.74 ± 3.330 \\
60°S–30°S & 7.768  & 7.32 ± 2.327 & -2.870  & 5.51 ± 2.337 & 22.530  & 4.98 ± 1.696 & 29.980  & 7.12 ± 3.800 \\
30°S–0°   & 2.730  & 4.47 ± 1.889 & -17.750 & 4.77 ± 2.313 & -25.480 & 3.49 ± 1.969 & 8.210   & 3.80 ± 1.713 \\
0°–30°N   & 3.746  & 4.03 ± 2.409 & -42.080 & 5.67 ± 3.152 & -100.100& 4.06 ± 2.277 & -43.360 & 2.83 ± 1.247 \\
30°N–60°N & 6.398  & 4.30 ± 2.556 & 14.860  & 4.79 ± 3.138 & 5.170   & 4.39 ± 1.941 & 13.060  & 5.06 ± 2.602 \\
60°N–90°N & 5.566  & 3.54 ± 1.867 & 34.680  & 3.50 ± 1.518 & 35.360  & 5.63 ± 1.948 & -4.030  & 5.42 ± 2.721 \\
\bottomrule
\end{tabular}
\vspace{0.1cm}
\end{minipage}\hfill
\begin{minipage}[t]{\textwidth}
\centering
\subcaption{TQC}
\begin{tabular}{l c cc cc cc c}
\toprule
 & \multirow{2}{*}{\texttt{climlab}} & \multicolumn{2}{c}{\texttt{fed05}} & \multicolumn{2}{c}{\texttt{\texttt{fed10}}} & \multicolumn{2}{c}{\texttt{nofed}} & \multirow{2}{*}{\texttt{ebm-v1}} \\
 &  & Mean ± Std & Gain \% & Mean ± Std & Gain \% & Mean ± Std & Gain \% &  \\
\midrule
90°S–60°S & 11.453 & 8.28 ± 0.896 & 8.550  & 8.13 ± 0.765 & 10.300 & 8.85 ± 1.124 & 2.320  & 9.06 ± 1.549 \\
60°S–30°S & 7.768  & 7.46 ± 1.618 & 6.440  & 7.08 ± 1.382 & 11.120 & 8.24 ± 1.716 & -3.390 & 7.97 ± 1.459 \\
30°S–0°   & 2.730  & 2.83 ± 1.195 & -26.510& 2.34 ± 1.053 & -4.390 & 3.32 ± 1.269 & -48.170& 2.24 ± 0.868 \\
0°–30°N   & 3.746  & 2.30 ± 0.511 & -25.400& 2.19 ± 0.604 & -19.480& 2.49 ± 1.074 & -35.470& 1.84 ± 0.561 \\
30°N–60°N & 6.398  & 0.93 ± 0.222 & 61.460 & 0.92 ± 0.149 & 61.860 & 1.23 ± 0.423 & 49.160 & 2.42 ± 0.706 \\
60°N–90°N & 5.566  & 1.34 ± 0.219 & 42.680 & 1.33 ± 0.352 & 43.140 & 1.44 ± 0.297 & 38.240 & 2.33 ± 0.767 \\
\bottomrule
\end{tabular}
\end{minipage}
\end{sidewaystable}

\clearpage

\begin{sidewaystable}[!h]
\centering
\small
\caption{Zonal‑band errors for \texttt{ebm-v2-optim-L-20k-a6}. Each subtable reports mean ± std and relative gain \% versus \texttt{ebm-v1} for three regimes \texttt{fed05}, \texttt{fed10} and \texttt{nofed}, along with a comparison against the static baseline \texttt{climlab}.}
\setlength{\tabcolsep}{4pt}
\captionsetup[sub]{font=small}
\begin{minipage}[t]{\textwidth}
\centering
\subcaption{DDPG}
\begin{tabular}{l c cc cc cc c}
\toprule
 & \multirow{2}{*}{\texttt{climlab}} & \multicolumn{2}{c}{\texttt{fed05}} & \multicolumn{2}{c}{\texttt{fed10}} & \multicolumn{2}{c}{\texttt{nofed}} & \multirow{2}{*}{\texttt{ebm-v1}} \\
 &  & Mean ± Std & Gain \% & Mean ± Std & Gain \% & Mean ± Std & Gain \% &  \\
\midrule
90°S–60°S & 11.453 & 7.26 ± 1.845 & 11.340 & 8.17 ± 2.886 & 0.260  & 20.52 ± 11.476 & -150.600 & 8.19 ± 3.433 \\
60°S–30°S & 7.768  & 1.63 ± 1.478 & 74.100 & 1.51 ± 0.324 & 76.020 & 1.62 ± 0.843  & 74.240  & 6.30 ± 4.019 \\
30°S–0°   & 2.730  & 1.92 ± 1.029 & 71.600 & 1.79 ± 0.624 & 73.580 & 2.31 ± 0.776  & 65.870  & 6.76 ± 3.013 \\
0°–30°N   & 3.746  & 1.89 ± 1.032 & 58.750 & 2.49 ± 1.509 & 45.850 & 2.59 ± 1.607  & 43.640  & 4.59 ± 2.068 \\
30°N–60°N & 6.398  & 1.71 ± 0.697 & 35.620 & 2.19 ± 0.643 & 17.530 & 1.81 ± 0.998  & 31.730  & 2.65 ± 0.997 \\
60°N–90°N & 5.566  & 2.43 ± 1.643 & 30.190 & 2.30 ± 1.338 & 34.050 & 2.42 ± 1.224  & 30.670  & 3.48 ± 1.790 \\
\bottomrule
\end{tabular}
\vspace{0.1cm}
\end{minipage}\hfill
\begin{minipage}[t]{\textwidth}
\centering
\subcaption{TD3}
\begin{tabular}{l c cc cc cc c}
\toprule
 & \multirow{2}{*}{\texttt{climlab}} & \multicolumn{2}{c}{\texttt{fed05}} & \multicolumn{2}{c}{\texttt{fed10}} & \multicolumn{2}{c}{\texttt{nofed}} & \multirow{2}{*}{\texttt{ebm-v1}} \\
 &  & Mean ± Std & Gain \% & Mean ± Std & Gain \% & Mean ± Std & Gain \% &  \\
\midrule
90°S–60°S & 11.453 & 7.01 ± 2.553 & 28.000  & 6.04 ± 0.933 & 38.030  & 15.90 ± 9.258 & -63.150  & 9.74 ± 3.330 \\
60°S–30°S & 7.768  & 3.16 ± 1.133 & 55.560  & 3.52 ± 1.276 & 50.510  & 3.52 ± 1.627 & 50.480   & 7.12 ± 3.800 \\
30°S–0°   & 2.730  & 7.68 ± 2.389 & -102.170& 8.16 ± 1.813 & -114.730& 6.10 ± 2.088 & -60.670  & 3.80 ± 1.713 \\
0°–30°N   & 3.746  & 7.27 ± 2.023 & -156.640& 7.63 ± 1.859 & -169.470& 6.05 ± 2.532 & -113.500 & 2.83 ± 1.247 \\
30°N–60°N & 6.398  & 3.92 ± 1.599 & 22.380  & 3.80 ± 0.851 & 24.770  & 3.47 ± 1.360 & 31.460   & 5.06 ± 2.602 \\
60°N–90°N & 5.566  & 2.97 ± 1.614 & 45.090  & 2.48 ± 1.563 & 54.120  & 5.55 ± 5.617 & -2.380   & 5.42 ± 2.721 \\
\bottomrule
\end{tabular}
\vspace{0.1cm}
\end{minipage}\hfill
\begin{minipage}[t]{\textwidth}
\centering
\subcaption{TQC}
\begin{tabular}{l c cc cc cc c}
\toprule
 & \multirow{2}{*}{\texttt{climlab}} & \multicolumn{2}{c}{\texttt{fed05}} & \multicolumn{2}{c}{\texttt{\texttt{fed10}}} & \multicolumn{2}{c}{\texttt{nofed}} & \multirow{2}{*}{\texttt{ebm-v1}} \\
 &  & Mean ± Std & Gain \% & Mean ± Std & Gain \% & Mean ± Std & Gain \% &  \\
\midrule
90°S–60°S & 11.453 & 34.96 ± 7.034 & -285.900 & 28.35 ± 7.956 & -212.890 & 33.90 ± 7.319 & -274.160 & 9.06 ± 1.549 \\
60°S–30°S & 7.768  & 1.68 ± 1.087  & 78.890  & 1.69 ± 1.215  & 78.780  & 1.28 ± 0.393  & 83.920  & 7.97 ± 1.459 \\
30°S–0°   & 2.730  & 1.97 ± 1.814  & 11.850  & 2.25 ± 1.590  & -0.480  & 0.80 ± 0.213  & 64.350  & 2.24 ± 0.868 \\
0°–30°N   & 3.746  & 1.21 ± 0.670  & 34.310  & 1.77 ± 1.363  & 3.430   & 0.75 ± 0.190  & 59.300  & 1.84 ± 0.561 \\
30°N–60°N & 6.398  & 1.97 ± 1.492  & 18.750  & 1.73 ± 0.551  & 28.600  & 1.17 ± 0.330  & 51.920  & 2.42 ± 0.706 \\
60°N–90°N & 5.566  & 30.70 ± 8.534 & -1215.560& 32.88 ± 9.606 & -1308.920& 43.12 ± 14.594& -1747.640& 2.33 ± 0.767 \\
\bottomrule
\end{tabular}
\end{minipage}
\end{sidewaystable}

\begin{sidewaystable}[!h]
\centering
\small
\caption{Zonal‑band errors for \texttt{ebm-v3-optim-L-20k-a2}. Each subtable reports mean ± std and relative gain \% versus \texttt{ebm-v1} for three regimes \texttt{fed05}, \texttt{fed10} and \texttt{nofed}, along with a comparison against the static baseline \texttt{climlab}.}
\setlength{\tabcolsep}{4pt}
\captionsetup[sub]{font=small}
\begin{minipage}[t]{\textwidth}
\centering
\subcaption{DDPG}
\begin{tabular}{l c cc cc cc c}
\toprule
 & \multirow{2}{*}{\texttt{climlab}} & \multicolumn{2}{c}{\texttt{fed05}} & \multicolumn{2}{c}{\texttt{fed10}} & \multicolumn{2}{c}{\texttt{nofed}} & \multirow{2}{*}{\texttt{ebm-v1}} \\
 &  & Mean ± Std & Gain \% & Mean ± Std & Gain \% & Mean ± Std & Gain \% &  \\
\midrule
90°S–60°S & 11.453 & 6.69 ± 1.766 & 18.240 & 7.52 ± 2.718 & 8.180  & 7.76 ± 2.193 & 5.200  & 8.19 ± 3.433 \\
60°S–30°S & 7.768  & 3.30 ± 1.168 & 47.540 & 4.20 ± 1.272 & 33.290 & 3.98 ± 2.135 & 36.810 & 6.30 ± 4.019 \\
30°S–0°   & 2.730  & 4.84 ± 2.151 & 28.340 & 3.77 ± 2.190 & 44.220 & 3.26 ± 1.873 & 51.720 & 6.76 ± 3.013 \\
0°–30°N   & 3.746  & 2.42 ± 1.786 & 47.180 & 2.96 ± 2.276 & 35.560 & 2.49 ± 1.461 & 45.840 & 4.59 ± 2.068 \\
30°N–60°N & 6.398  & 2.00 ± 0.885 & 24.650 & 2.73 ± 1.056 & -2.980 & 2.67 ± 1.400 & -0.700 & 2.65 ± 0.997 \\
60°N–90°N & 5.566  & 1.96 ± 0.681 & 43.670 & 1.69 ± 1.035 & 51.400 & 1.63 ± 1.024 & 53.230 & 3.48 ± 1.790 \\
\bottomrule
\end{tabular}
\vspace{0.1cm}
\end{minipage}\hfill
\begin{minipage}[t]{\textwidth}
\centering
\subcaption{TD3}
\begin{tabular}{l c cc cc cc c}
\toprule
 & \multirow{2}{*}{\texttt{climlab}} & \multicolumn{2}{c}{\texttt{fed05}} & \multicolumn{2}{c}{\texttt{fed10}} & \multicolumn{2}{c}{\texttt{nofed}} & \multirow{2}{*}{\texttt{ebm-v1}} \\
 &  & Mean ± Std & Gain \% & Mean ± Std & Gain \% & Mean ± Std & Gain \% &  \\
\midrule
90°S–60°S & 11.453 & 7.53 ± 1.444 & 22.700  & 7.42 ± 1.159 & 23.880  & 7.54 ± 1.390 & 22.630  & 9.74 ± 3.330 \\
60°S–30°S & 7.768  & 5.99 ± 2.663 & 15.800  & 5.51 ± 2.616 & 22.590  & 5.00 ± 2.586 & 29.790  & 7.12 ± 3.800 \\
30°S–0°   & 2.730  & 3.84 ± 2.378 & -1.030  & 3.65 ± 2.640 & 3.990   & 3.32 ± 1.380 & 12.690  & 3.80 ± 1.713 \\
0°–30°N   & 3.746  & 7.52 ± 2.875 & -165.300& 7.54 ± 3.263 & -166.260& 5.94 ± 2.701 & -109.610& 2.83 ± 1.247 \\
30°N–60°N & 6.398  & 6.55 ± 1.837 & -29.630 & 6.85 ± 2.390 & -35.450 & 7.02 ± 2.581 & -38.880 & 5.06 ± 2.602 \\
60°N–90°N & 5.566  & 3.94 ± 2.302 & 27.250  & 4.00 ± 1.708 & 26.240  & 4.49 ± 1.536 & 17.170  & 5.42 ± 2.721 \\
\bottomrule
\end{tabular}
\vspace{0.1cm}
\end{minipage}\hfill
\begin{minipage}[t]{\textwidth}
\centering
\subcaption{TQC}
\begin{tabular}{l c cc cc cc c}
\toprule
 & \multirow{2}{*}{\texttt{climlab}} & \multicolumn{2}{c}{\texttt{fed05}} & \multicolumn{2}{c}{\texttt{\texttt{fed10}}} & \multicolumn{2}{c}{\texttt{nofed}} & \multirow{2}{*}{\texttt{ebm-v1}} \\
 &  & Mean ± Std & Gain \% & Mean ± Std & Gain \% & Mean ± Std & Gain \% &  \\
\midrule
90°S–60°S & 11.453 & 8.27 ± 0.378 & 8.700   & 10.78 ± 4.522 & -18.950  & 8.26 ± 0.303 & 8.840   & 9.06 ± 1.549 \\
60°S–30°S & 7.768  & 7.51 ± 0.869 & 5.780   & 10.76 ± 4.459 & -35.000  & 7.62 ± 0.729 & 4.450   & 7.97 ± 1.459 \\
30°S–0°   & 2.730  & 2.60 ± 0.718 & -15.960 & 7.27 ± 5.706  & -224.680 & 3.02 ± 0.490 & -34.920 & 2.24 ± 0.868 \\
0°–30°N   & 3.746  & 2.84 ± 0.774 & -54.750 & 9.67 ± 12.637 & -426.670 & 3.90 ± 0.434 & -112.690& 1.84 ± 0.561 \\
30°N–60°N & 6.398  & 3.59 ± 1.179 & -48.070 & 11.05 ± 17.379& -355.740 & 4.13 ± 0.461 & -70.350 & 2.42 ± 0.706 \\
60°N–90°N & 5.566  & 3.35 ± 1.003 & -43.330 & 11.17 ± 18.477& -378.590 & 3.36 ± 0.394 & -43.800 & 2.33 ± 0.767 \\
\bottomrule
\end{tabular}
\end{minipage}
\end{sidewaystable}

\begin{sidewaystable}[!h]
\centering
\small
\caption{Zonal‑band errors for \texttt{ebm-v3-optim-L-20k-a6}. Each subtable reports mean ± std and relative gain \% versus \texttt{ebm-v1} for three regimes \texttt{fed05}, \texttt{fed10} and \texttt{nofed}, along with a comparison against the static baseline \texttt{climlab}.}
\setlength{\tabcolsep}{4pt}
\captionsetup[sub]{font=small}
\begin{minipage}[t]{\textwidth}
\centering
\subcaption{DDPG}
\begin{tabular}{l c cc cc cc c}
\toprule
 & \multirow{2}{*}{\texttt{climlab}} & \multicolumn{2}{c}{\texttt{fed05}} & \multicolumn{2}{c}{\texttt{fed10}} & \multicolumn{2}{c}{\texttt{nofed}} & \multirow{2}{*}{\texttt{ebm-v1}} \\
 &  & Mean ± Std & Gain \% & Mean ± Std & Gain \% & Mean ± Std & Gain \% &  \\
\midrule
90°S–60°S & 11.453 & 7.01 ± 2.782 & 14.340  & 7.22 ± 1.399 & 11.800  & 7.23 ± 1.652 & 11.690  & 8.19 ± 3.433 \\
60°S–30°S & 7.768  & 4.34 ± 3.372 & 31.000  & 3.69 ± 1.102 & 41.450  & 8.08 ± 4.816 & -28.370 & 6.30 ± 4.019 \\
30°S–0°   & 2.730  & 1.25 ± 0.695 & 81.440  & 1.22 ± 0.526 & 81.890  & 19.92 ± 4.845& -194.680& 6.76 ± 3.013 \\
0°–30°N   & 3.746  & 1.48 ± 0.617 & 67.830  & 1.71 ± 1.218 & 62.710  & 17.39 ± 5.346& -278.850& 4.59 ± 2.068 \\
30°N–60°N & 6.398  & 1.51 ± 0.710 & 43.220  & 1.57 ± 0.796 & 40.880  & 5.92 ± 6.272& -123.010& 2.65 ± 0.997 \\
60°N–90°N & 5.566  & 1.17 ± 0.442 & 66.490  & 1.39 ± 0.680 & 60.240  & 1.76 ± 1.176& 49.480 & 3.48 ± 1.790 \\
\bottomrule
\end{tabular}
\vspace{0.1cm}
\end{minipage}\hfill
\begin{minipage}[t]{\textwidth}
\centering
\subcaption{TD3}
\begin{tabular}{l c cc cc cc c}
\toprule
 & \multirow{2}{*}{\texttt{climlab}} & \multicolumn{2}{c}{\texttt{fed05}} & \multicolumn{2}{c}{\texttt{fed10}} & \multicolumn{2}{c}{\texttt{nofed}} & \multirow{2}{*}{\texttt{ebm-v1}} \\
 &  & Mean ± Std & Gain \% & Mean ± Std & Gain \% & Mean ± Std & Gain \% &  \\
\midrule
90°S–60°S & 11.453 & 14.05 ± 2.941 & -44.230 & 15.86 ± 1.632 & -62.780 & 12.47 ± 8.601 & -28.000 & 9.74 ± 3.330 \\
60°S–30°S & 7.768  & 10.90 ± 0.644 & -53.120 & 10.59 ± 0.481 & -48.860 & 9.36 ± 5.302  & -31.570 & 7.12 ± 3.800 \\
30°S–0°   & 2.730  & 21.30 ± 0.639 & -460.730& 21.14 ± 0.473 & -456.380& 17.06 ± 4.570 & -349.090& 3.80 ± 1.713 \\
0°–30°N   & 3.746  & 21.41 ± 0.467 & -655.550& 21.23 ± 0.611 & -649.370& 14.34 ± 2.496 & -405.970& 2.83 ± 1.247 \\
30°N–60°N & 6.398  & 9.47 ± 0.516  & -87.290 & 9.36 ± 0.581  & -85.160 & 5.78 ± 0.977  & -14.340 & 5.06 ± 2.602 \\
60°N–90°N & 5.566  & 4.57 ± 0.484  & 15.600  & 4.48 ± 0.536  & 17.280  & 9.27 ± 6.494  & -71.180 & 5.42 ± 2.721 \\
\bottomrule
\end{tabular}
\vspace{0.1cm}
\end{minipage}\hfill
\begin{minipage}[t]{\textwidth}
\centering
\subcaption{TQC}
\begin{tabular}{l c cc cc cc c}
\toprule
 & \multirow{2}{*}{\texttt{climlab}} & \multicolumn{2}{c}{\texttt{fed05}} & \multicolumn{2}{c}{\texttt{\texttt{fed10}}} & \multicolumn{2}{c}{\texttt{nofed}} & \multirow{2}{*}{\texttt{ebm-v1}} \\
 &  & Mean ± Std & Gain \% & Mean ± Std & Gain \% & Mean ± Std & Gain \% &  \\
\midrule
90°S–60°S & 11.453 & 21.12 ± 1.703 & -133.090 & 23.23 ± 8.282 & -156.370 & 21.01 ± 1.702 & -131.930 & 9.06 ± 1.549 \\
60°S–30°S & 7.768  & 18.84 ± 0.971 & -136.340 & 22.08 ± 8.752 & -177.000 & 18.43 ± 1.047 & -131.180 & 7.97 ± 1.459 \\
30°S–0°   & 2.730  & 9.12 ± 0.779  & -307.220 & 10.84 ± 4.123 & -383.720 & 8.54 ± 0.511  & -281.410 & 2.24 ± 0.868 \\
0°–30°N   & 3.746  & 9.41 ± 0.789  & -412.530 & 10.82 ± 2.018 & -489.610 & 8.62 ± 0.503  & -369.770 & 1.84 ± 0.561 \\
30°N–60°N & 6.398  & 11.87 ± 2.754 & -389.580 & 15.20 ± 5.150 & -527.000 & 9.85 ± 0.519  & -306.400 & 2.42 ± 0.706 \\
60°N–90°N & 5.566  & 14.21 ± 7.349 & -509.050 & 20.12 ± 8.341 & -762.090 & 9.07 ± 1.682  & -288.430 & 2.33 ± 0.767 \\
\bottomrule
\end{tabular}
\end{minipage}
\end{sidewaystable}

\clearpage

\subsubsection{Local Skill Plots}
\label{app:results-ebm-v23-ddpg-local-skill}

\begin{figure}[!h]
    \centering
    \makebox[\textwidth][c]{\includegraphics[scale=1]{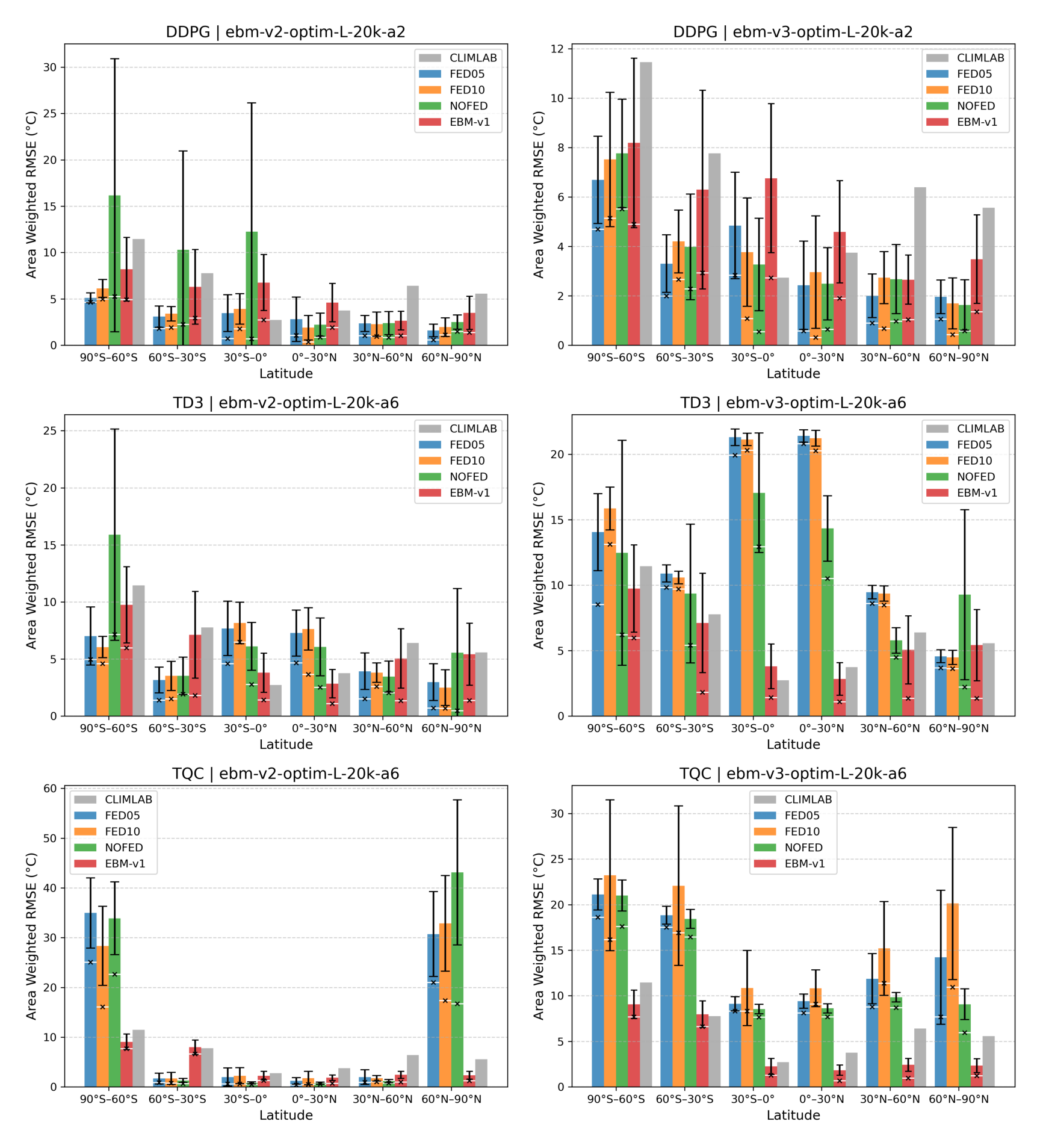}}  
    \caption{Comparison of zonal skill achieved by DDPG, TD3 and TQC under FedRL coordination in \texttt{ebm-v2} and \texttt{ebm-v3}, using the 2-agent spatial decomposition (DDPG) and 6-agent spatial decomposition (\texttt{a6} - TD3 and TQC). Skill is evaluated using areaWRMSE between predicted and reference temperature profiles, averaged with 95\% CI spreads over 10 seeds. Each subplot reports results for three FedRL schemes: \texttt{fed05}, \texttt{fed10}, \texttt{nofed}, along with single-agent \texttt{ebm-v1} and the static \texttt{climlab} baseline. White horizontal bars with a cross indicate the best-performing seed for each scheme. Both setups adopt the same policy network architecture and hyperparameters as \texttt{ebm-v1}. Tabulated results in Appendix~\ref{app:results-ebm-fedrl-skill-metrics}.}
    \label{fig:app-results-ebm-v23-ddpg-local-skill}
\end{figure}

For TD3 (second row in Figure~\ref{fig:app-results-ebm-v23-ddpg-local-skill}), \texttt{ebm-v2} results show competitive skill in tropical and mid-latitude zones under \texttt{fed05}, often matching or surpassing the single-agent \texttt{ebm-v1}. However, variance rises sharply in the polar bands, particularly in the Southern Hemisphere, where strong gradients prove harder to capture. In \texttt{ebm-v3}, TD3 exhibits more pronounced instability: although \texttt{fed05} remains the most stable regime, episodic collapses at high latitudes drive RMSE higher than in \texttt{ebm-v2}. This behaviour suggests that the reduced, region-specific inputs of \texttt{ebm-v3} may conflict with hyperparameters tuned for global inputs in \texttt{ebm-v1}, causing mismatches in critic ensemble updates and amplifying instability.

TQC (third row in Figure~\ref{fig:app-results-ebm-v23-ddpg-local-skill}) performs strongly in the tropical bands of \texttt{ebm-v2} under \texttt{fed05}, achieving clear gains over the single-agent \texttt{ebm-v1}. Yet the method shows instability at high latitudes and wider error spreads under \texttt{fed10}, highlighting its dependence on frequent synchronisation for stability. In \texttt{ebm-v3}, performance degrades further, with mid-latitude instability and polar areaWRMSE in several cases exceeding that of \texttt{ebm-v1}. The large critic ensemble that benefits global contexts may be less effective when learning from regional input profiles and localised rewards, leading to overfitting or noisy updates. Overall, these results suggest that while both TD3 and TQC can deliver strong performance in favourable regimes, DDPG’s simpler architecture is more resilient to the structural shifts between \texttt{ebm-v2} and \texttt{ebm-v3}.

\end{appendix}

\clearpage


\end{document}